\documentclass[pdflatex,sn-nature]{sn-jnl}

\usepackage{graphicx}%
\usepackage{amsmath,amssymb,amsfonts}%
\usepackage{textcomp}%
\usepackage{url}%
\usepackage{booktabs}%
\usepackage{subcaption}%
\usepackage{hyperref}%

\usepackage{tabularx}%
\usepackage[table]{xcolor}%
\usepackage[most]{tcolorbox}%
\usepackage{float}%
\usepackage{longtable}%
\usepackage{array}%
\usepackage{enumitem}%
\usepackage{pifont}%
\usepackage[section]{placeins}%
\usepackage{makecell}%

\newcommand{\checked}{\ding{110}}
\newcommand{\unchecked}{\ding{111}}

\definecolor{coachcolor}{HTML}{E59740}
\definecolor{didacticcolor}{HTML}{42988E}
\definecolor{controlcolor}{HTML}{999999}

\begin{document}

\title[Practicing with language models cultivates empathic communication]{Practicing with language models cultivates human empathic communication}

\author*[1,2,3]{\fnm{Aakriti} \sur{Kumar}}\email{aakriti.kumar@kellogg.northwestern.edu}
\author[1,2]{\fnm{Nalin} \sur{Poungpeth}}
\author[4]{\fnm{Diyi} \sur{Yang}}
\author[5]{\fnm{Bruce} \sur{Lambert}}
\author*[1,2,3,6]{\fnm{Matthew} \sur{Groh}}\email{matthew.groh@kellogg.northwestern.edu}

\affil*[1]{\orgname{Kellogg School of Management, Northwestern University}}
\affil[2]{\orgname{Northwestern Institute on Complex Systems, Northwestern University}}
\affil[3]{\orgname{Ryan Institute on Complexity, Northwestern University}}
\affil[4]{\orgdiv{Department of Computer Science}, \orgname{Stanford University}}
\affil[5]{\orgdiv{Department of Communication Studies}, \orgname{Northwestern University}}
\affil[6]{\orgdiv{Department of Computer Science}, \orgname{Northwestern University}}

\abstract{Empathy is central to human connection, yet people often struggle to express it effectively. In blinded evaluations, large language models (LLMs) generate responses that are often judged more empathic than human-written ones. Yet when a response is attributed to AI, recipients feel less heard than when comparable responses are attributed to a human. We built a conversation platform in which participants are asked to offer empathic support to an LLM expressing realistic troubles and conducted a randomized experiment collecting 33,938 messages spanning 2,904 text-based conversations between 968 participants and their LLM conversational partners. We find participants report feeling empathy but systematically fail to express it, but an LLM coaching intervention offering personalized feedback on effective empathic communication significantly boosts it without homogenizing participants' responses. Moreover, we derive a data-driven taxonomy of idiomatic empathic expressions in naturalistic dialogues across personal and workplace trouble scenarios. These results advance the scientific understanding of how empathy is expressed and demonstrate a scalable, AI-based intervention for scaffolding and cultivating it.}


\maketitle

\section*{Introduction}\label{sec:introduction}
Empathy underpins human social life, shaping relationships, cooperation, and well-being. Yet communicating in a manner that makes another person feel heard can be challenging in practice~\cite{goldsmith2004communicating, zaki2015addressing, sharma2023human, yin2024ai}. Empathy relies on observational learning~\cite{zhou2024social} and empathic communication is a learnable skill~\cite{teding2016efficacy, gryglewicz2020examining, riess2012empathy, covey20207, suchman1997model, bylund2005examining, schumann2014addressing} that shows substantial individual variability~\cite{drollinger2006development, sharma2020computational, kumar2026large}. In blinded evaluations, large language model (LLM) responses to people expressing troubles are judged as more empathetic than the average human-written ones~\cite{ayers2023comparing, sharma2023human, sorin2024large, inzlicht2024praise, li2024skill, ovsyannikova2025third}. Nevertheless, most people report feeling significantly less heard and supported after they learn that an empathic message comes from an AI~\cite{yin2024ai, rubin2025comparing}. In light of the importance of human presence in empathic communication \cite{perry2023ai, li2026random}, LLMs' superior skill relative to the average person \cite{inzlicht2024praise, yin2024ai}, and growing evidence of LLMs' effectiveness as coaches and tutors \cite{sharma2023human, kestin2025ai, khasentino2025personal}, a natural question emerges: How can people learn from LLMs to respond to other's troubles in a way that makes them feel heard and understood?

The costs of failed empathic communication are profound. In society, a lack of human connection is linked to rising loneliness, polarization, and decline in well-being \cite{bruce2019loneliness, surkalim2022prevalence, pei2025bridging}. At work, breakdowns in empathy undermine collaboration, leadership, and trust \cite{lloyd2017listening, li2024ethical}. As remote work and hybrid human-AI teams reduce the informal interactions that once built mutual understanding \cite{yang2022effects, emanuel2026home}, the capacity to make others feel heard becomes harder to practice and more essential to sustain. AI chatbots do not match human connection when it comes to fulfilling psychological and social needs \cite{machia2024social, zimmerman2024human, wenger2026people, li2026random}, and heavy reliance on them for social support may foster dependence and worsen well-being \cite{phang2025investigating}. In contrast, everyday empathic exchanges with other people are associated with increased well-being \cite{depow2021experience, moore2025expressing}. Likewise, learning about the personal narratives of others can increase connection with highly stigmatized groups~\cite{reddan2024film}. For these reasons,  empathic communication as a skill is worthy of cultivation in humans and should not be outsourced to AI.

\begin{figure}[p]
    \centering
    \begin{subfigure}{.94\linewidth}
        \subcaption{}    
        \includegraphics[width=\linewidth, trim=0 0 0 .5cm, clip]{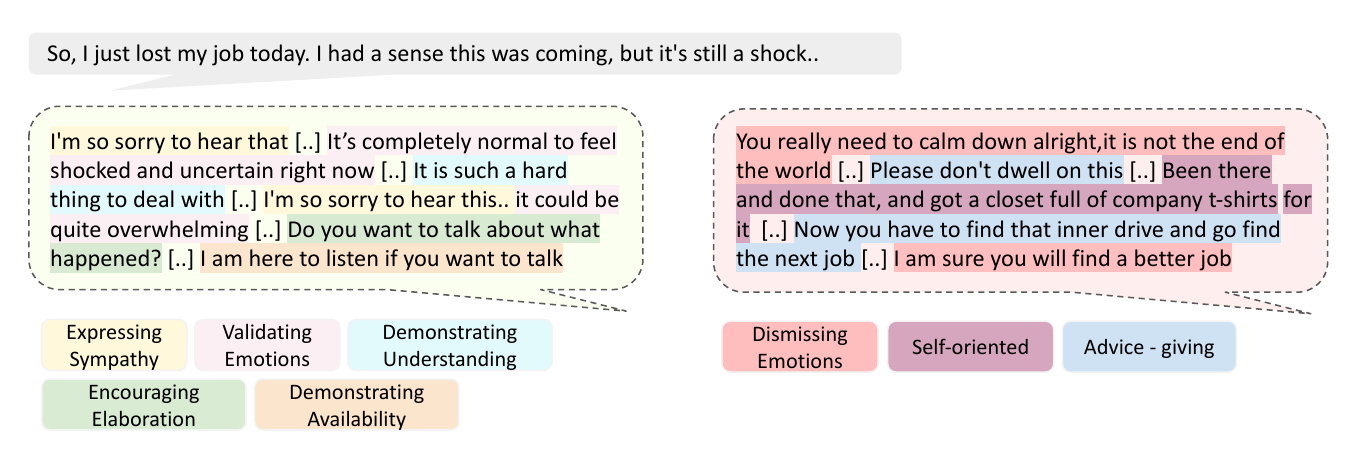}
    \end{subfigure}
        \begin{subfigure}{.94\linewidth}
        \subcaption{}
        \includegraphics[width=\linewidth]{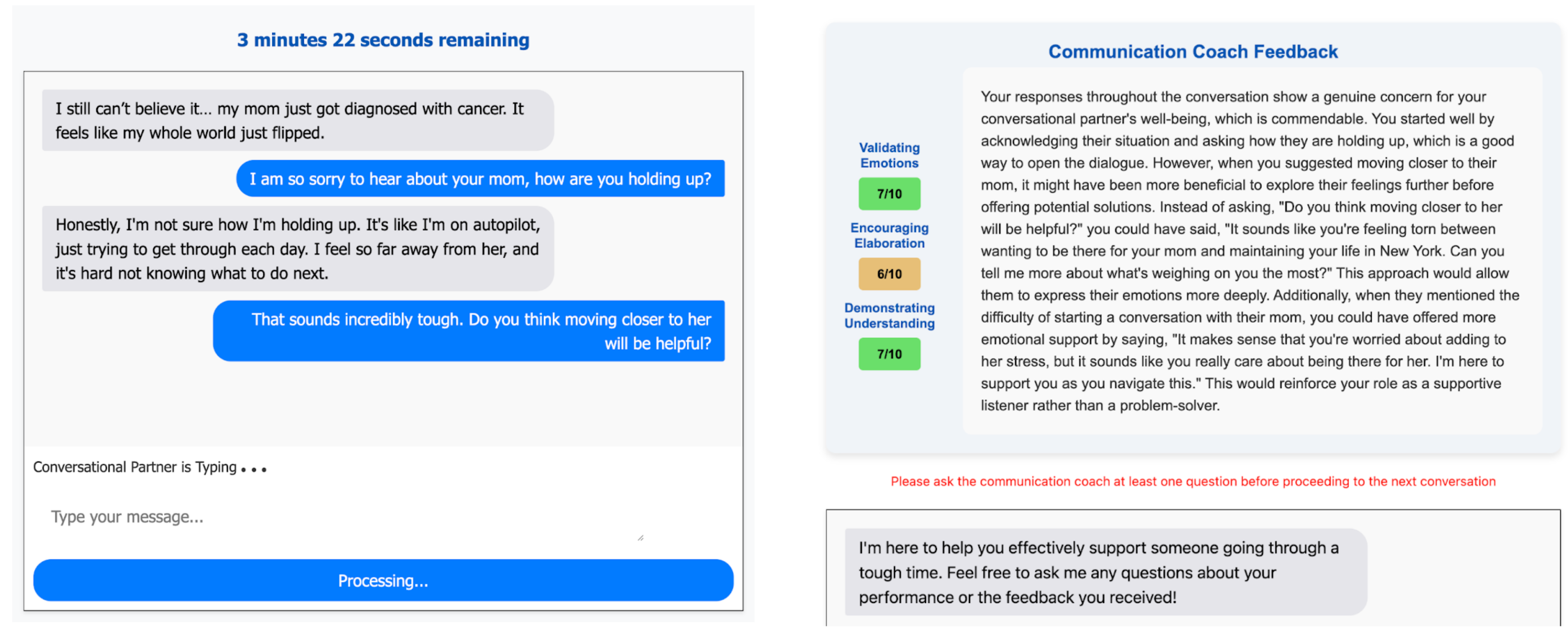}
    \end{subfigure}
    \begin{subfigure}{.94\linewidth}
        \subcaption{}
        \includegraphics[width=\linewidth]{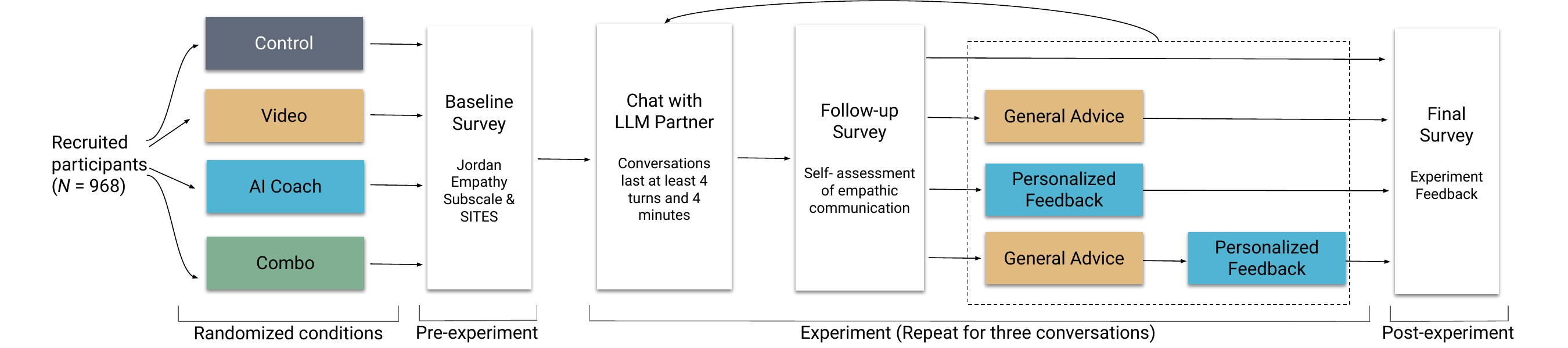}
    \end{subfigure}
    \caption{\textbf{Overview of the \textit{Lend an Ear} experiment.} A. Examples of empathic responses by participants to a conversational partner's disclosure of job loss on the Lend an Ear platform that tend to make people feel heard (left panel) and that tend to not be effective at making people feel heard (right panel). B. User interface of the chat window with an LLM conversational partner (left) and personalized feedback from the AI coach (right). C. Experimental design flowchart illustrating participant recruitment, random assignment to four conditions (Control, Video, AI Coach, Combined), and the sequence of surveys, conversations, and feedback.}
    \label{exp-platform}
\end{figure}

Empathic communication is not a single behavior but a constellation of communicative components. Research on empathic communication shows that responses which make people feel heard encourage elaboration~\cite{eyal2018perspective, moyers2014motivational, rodriguez2019measuring}, validate emotions \cite{rizvi2016dialectical}, and demonstrate understanding~\cite{rogers1961becoming, bodie2011active, mercer2004consultation, kim2018social}; counterproductive ones offer unsolicited advice \cite{Feeney_Collins_2018, goldsmith1997normative}, shift focus away from the speaker's experience~\cite{goldsmith1997normative, weger2014relative}, or dismiss emotions~\cite{jones2004personcentered, hacker2008relational} (See Fig.~\ref{exp-platform}A for examples of these responses). Empathic communication is highly context-dependent, often defying simplistic rubrics and making structured training a challenge. For instance, subtle phrasing can signal validation in one context but come across as patronizing in another, complicating efforts to codify it. This complexity makes empathic communication difficult to measure and teach. While traditional interventions for empathic communication training have been shown to be effective \cite{paulus2022effectiveness, king2013best, kahriman2016effect}, they are resource-intensive and hence limited in reach. Brief, scalable empathic-mindset interventions have shown to shift outcomes in field settings \cite{okonofua2022scalable}, but these target empathic disposition rather than the communicative idiom through which empathy is expressed.

LLMs, with their demonstrated capabilities to generate \cite{lee2024large, yin2024ai, rubin2025comparing} and evaluate empathic communication in text \cite{kumar2026large}, offer a way forward. LLM-powered systems can simulate realistic practice partners, deliver personalized feedback, serve as reliable evaluators, and scale to reach learners who would otherwise lack access to structured training \cite{yang2024social}. This approach has started to show promise in coaching people across a range of interpersonal skills including conflict resolution in personal relationships \cite{chun2025conflictlens}, professional communication \cite{li2024exploring}, negotiation \cite{dinnar2025negotiation, duddu2025does}, democratic deliberation \cite{tessler2024ai}, and counseling \cite{louie2026can}. These findings offer a blueprint for a scalable way to both measure and cultivate empathic communication using AI tools. Whether brief LLM-based interventions can cultivate empathic communication remains an open empirical question.

In the \textit{Lend an Ear} experiment, we ask whether LLMs can be used to help people practice and improve their ability to communicate empathically. \textit{Lend an Ear} is designed as an interactive role-playing game where people practice offering empathic support to an AI role-playing partner. In each conversation, an AI partner simulates someone experiencing either a personal trouble (a family member diagnosed with cancer in one and passing away in another) or a workplace trouble (losing a job, getting passed over for a promotion, and feeling undervalued at work). Participants role-play as supporters in three conversations, offering responses across multiple conversational turns, and receive personalized feedback from an AI coach or through short videos. In a preregistered randomized experiment with 968 participants, producing 2904 conversations with an average of 11 turns per conversation, and a total of 16,975 human messages and 16,963 LLM-generated messages, we evaluate the impact of personalized feedback from an LLM coach on participants' empathic communication performance. 

Our dataset of 16,975 human messages enables quantitative and qualitative analysis of how people express empathy in naturalistic conversation and when their attempts at offering empathic support align (or not) with established frameworks for empathic communication.  Surprisingly, our results reveal a disconnect between self-reported empathy, felt empathy, and expressed empathy, suggesting that feeling empathy and communicating empathically are distinct. We find that a brief coaching intervention, powered by LLM feedback, significantly improves expressed empathy across six preregistered dimensions of prescriptive (encouraging elaboration, validating emotions, demonstrating understanding) and proscriptive (giving unsolicited advice, reorienting the conversation to oneself, dismissing emotions) communication behaviors. Finally, in a follow-up human preference experiment, we find that these behavioral shifts correspond to what independent raters perceive as more empathic. 

These findings advance the science of empathic communication by demonstrating that targeted practice with LLM partners can improve performance in controlled settings, by providing a data-driven taxonomy of empathic message contents grounded in naturalistic dialogue, and by showcasing the silent empathy effect that trait empathy is unrelated to empathic communication skill. Our approach demonstrates a scalable method for strengthening empathic skills at a time when human connection is both deeply needed and increasingly fragile. More broadly, this work offers guidance for reframing empathic communication from an intangible ``soft'' skill into a ``hard'' skill that can be quantified, trained, and strengthened.

\subsection*{Lend an Ear Platform}

We designed a custom, interactive web platform called \textit{Lend an Ear} to evaluate whether practicing empathic communication with AI conversational partners and receiving personalized LLM-generated feedback can improve participants' empathic communication skills. This role-playing setup simulates realistic interpersonal scenarios in which a conversational partner seeks support, allowing participants to practice offering empathic responses across multiple conversational turns. LLM conversational partners simulated five distinct scenarios spanning workplace troubles (losing a job, getting passed over for a promotion, and feeling undervalued at work) and personal troubles (a family member diagnosed with cancer in one and passing away in another). For each scenario, the LLM role-playing agent was provided with a detailed background story establishing their identity and the specific trouble they were experiencing. These LLM conversational partners were instructed to behave as individuals seeking to feel heard and understood, expressing their concerns and emotions in two to three sentences per turn across multi-turn conversations. At the end of the experiment, we asked participants to rate their agreement with the statement, ``The troubles that my conversational partners described seemed realistic'', on a five-point Likert scale (1 = not at all, 2 = slightly, 3 = somewhat, 4 = quite a bit, 5 = very much). 91\% of participants rated the scenarios as ``quite a bit'' or ``very much'' realistic. An LLM communication coach prompted using a comprehensive framework of empathic communication principles provided automated, personalized feedback on empathic communication skills. See Methods for details. Fig.~\ref{exp-platform}B shows screenshots of the chat interface and the coach feedback window.

We conducted a preregistered experiment where participants were randomly assigned to one of four conditions: (1) a control condition with no feedback, (2) two short instructive videos (57 and 35 seconds) featuring a human communication coach, (3) an interactive AI coaching system with access to participants' conversations and availability for follow-up questions, and (4) a combination of the AI coaching system and the human coach videos (see Preregistration in Materials and Methods for more details). Fig.~\ref{exp-platform}C illustrates the experimental flow. We recruited 968 participants via Prolific, targeting a demographically representative U.S. sample, resulting in 2,904 conversations and 33,938 messages. Participants were randomly assigned to one of the four conditions. They first reviewed instructions outlining the procedure and their role in the conversations, and then completed a baseline survey, including the Jordan empathy subscale~\cite{jordan2016empathy} and the SITES measure~\cite{konrath2018development} to capture self-reported trait empathy. Participants engaged in a four-minute text-based conversation initiated by the AI partner. After each conversation, participants completed a brief four-item self-assessment of their empathic responses. Depending on their assigned condition, they either proceeded directly to the next conversation (control) or received feedback based on their assigned coaching intervention before continuing. This cycle repeated until each participant completed three conversations with different LLM role-playing partners.

\section*{Results}
In the results presented here, we evaluate how participants communicate empathic support, how coaching interventions affect participant performance, and how self-reported trait empathy relates to expressed empathic communication. We use an LLM-as-judge paradigm to score participants' responses on six preregistered dimensions of empathic communication including encouraging elaboration (asking questions to prompt the partner to share more about their experiences and emotions) \cite{bodie2011active}, validating emotions (acknowledging and affirming the partner's feelings) \cite{kim2018social}, demonstrating understanding (paraphrasing the partner's experiences to show comprehension) \cite{gerdes2011teaching}, providing unsolicited advice (offering guidance without first asking if it is wanted) \cite{fitzsimons2004reactance}, self-oriented responding (shifting focus away from the partner's experience) \cite{burleson2008counts}, and dismissing emotions (minimizing or invalidating the partner's feelings) \cite{yao2023person}. These dimensions can be reliably annotated by LLMs~\cite{kumar2026large} and serve as the primary dependent variables for our analysis. Fig.~\ref{exp-platform}A shows examples of normative prescriptive and proscriptive empathic responses to a support seeker's disclosure of a job loss. Finally, we also present results from a follow-up human preference experiment where independent raters choose what they believe to be the more empathic conversation from pairs of Lend an Ear participant conversations, allowing us to evaluate whether higher-scoring conversations also align with people’s preferences for empathic communication. 

\subsection*{Mapping Empathic Communication with k-Sparse Autoencoders}
We find high variability in participants' responses with respect to the wording of how they respond, their alignment with empathic communication norms, and the conceptual message with which they respond. We find 97.5\% of 16,975 messages written by participants are unique with only 421 exact duplicated messages (e.g. 13 messages saying “I am so sorry to hear that”, 13 messages saying “I’m sorry to hear that”). Prior to any intervention,  participants' responses' alignment with empathic communication norms varied widely spanning nearly the entire possible range of scores from -11 to 12 with a standard deviation of 4.1 points. This diversity reflects  variation in how people communicate empathy. See Supplementary Information for baseline differences in empathic communication scores across workplace and personal troubles in the first conversation across conditions.

\begin{figure}[h]
    \centering

    \begin{subfigure}{.95\textwidth}
        \subcaption{}
        \includegraphics[width=\linewidth]{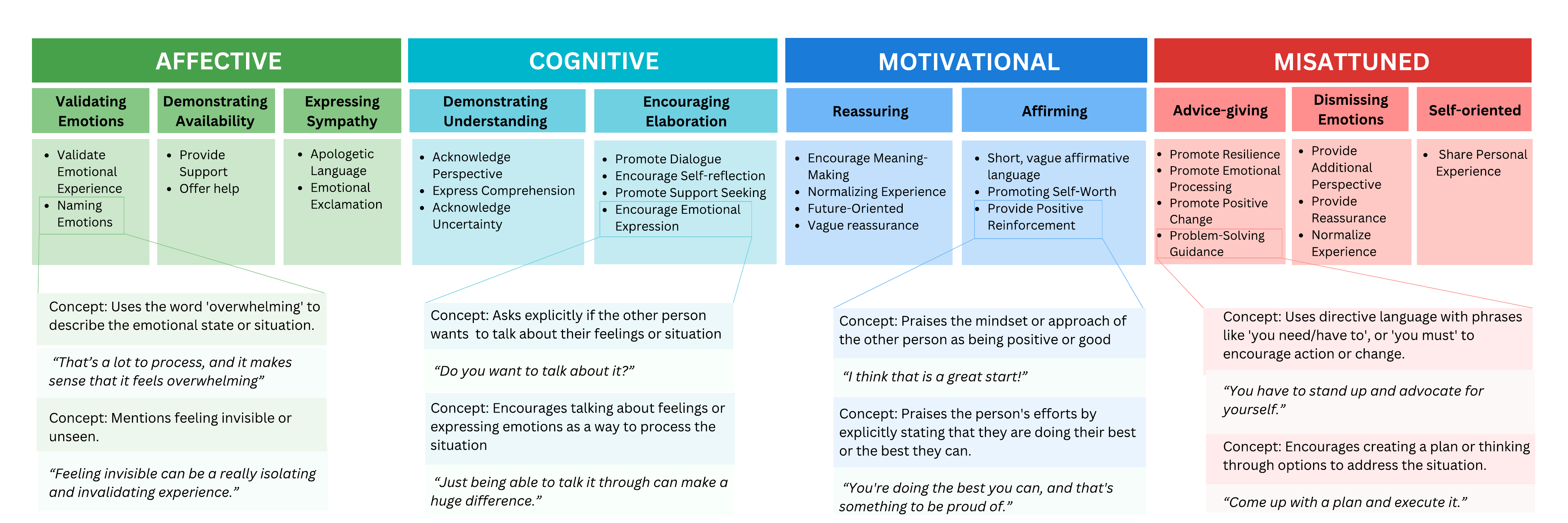}
    \end{subfigure}

    \begin{subfigure}{.95\textwidth}
        \subcaption{}
        \includegraphics[width=\linewidth]{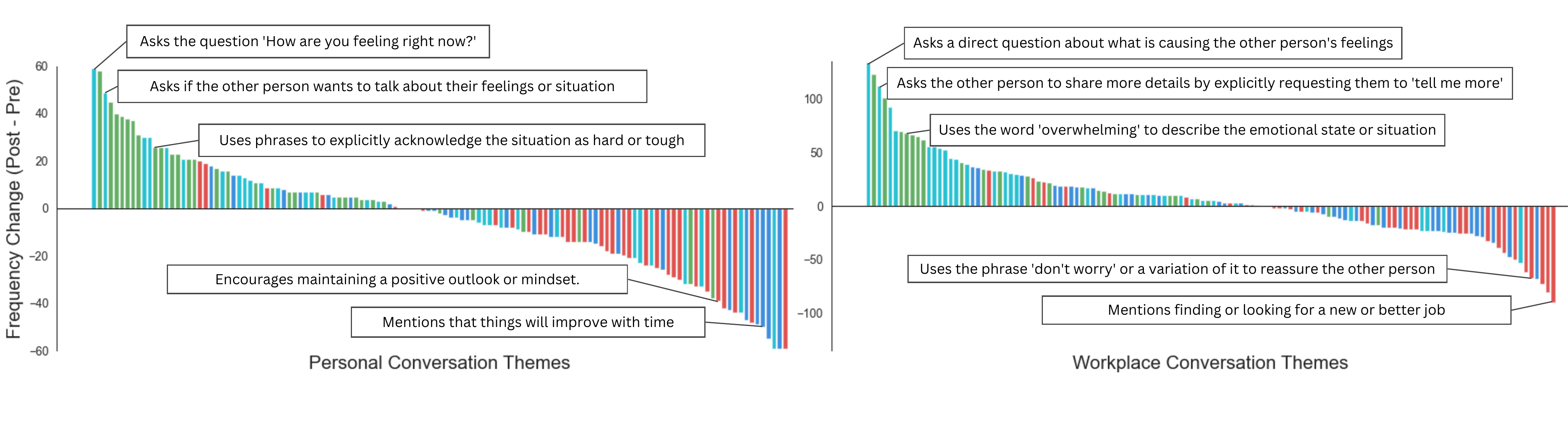}
    \end{subfigure}

    \caption{\textbf{Communication patterns in conversations.} \textbf{A.} Hierarchical taxonomy of empathic communication in naturalistic dialogues in Lend an Ear. This four-level structure integrates bottom-up discovery of 128 themes via k-sparse autoencoders with qualitatively coded top-down theoretical categories (Affective, Cognitive, Motivational, and Misattuned). \textbf{B.} Frequency changes (post minus pre) across all 128 themes identified by the k-sparse autoencoder for the AI coach and Combined conditions, ranked by magnitude of change for personal and workplace scenarios. Categories are color-coded as Affective (green), Cognitive (cyan), Motivational (blue), and Misattuned (red).}
    \label{fig:hierarchy}
\end{figure}

We map communicative diversity by empirically identifying the phrasal lexicon~\cite{becker1975phrasal, o1995managing, lambert2001semi} (the communicative moves participants used) using a k-sparse autoencoder (kSAE) on text embeddings of 29,520 sentence-level units extracted from 16,975 messages in 2,904 conversations. The kSAE learns a compressed, interpretable representation of the embeddings by reconstructing the input while activating only the top-k features per input and enforcing sparsity~\cite{peng2025use, singhdiscovering}. This sparsity constraint helps distill recurring linguistic patterns in a data-driven way, surfacing latent concepts that capture thematically coherent expressions across our data. In our analysis, each sentence was assigned to its top two activating features to account for polysemous sentences that could align with multiple thematic concepts. We identified 128 latent concepts as optimal through a grid search over the number of latent features ranging from $2^4$ (16) to $2^8$ (256), balancing clustering quality (silhouette score of 0.42 for 128, compared to 0.35 for 64 features and 0.38 for 256) with interpretability and thematic distinctiveness (see Methods). To interpret the resulting latent features, we used an LLM to generate human-readable descriptions of each feature based on high-activating examples, allowing us to scale analysis across the large dataset.

We developed a four-level hierarchical taxonomy combining the bottom-up data-driven approach with a top-down theory-driven mapping. The bottom-up approach leveraged the kSAE to identify meaningful themes directly from the data. From a top-down perspective, empathy is well-studied in psychology and includes three dimensions of empathic engagement~\cite{zaki2012neuroscience, perry2023ai}: affective empathy (sharing others' emotions while maintaining a self–other distinction), cognitive empathy (recognizing and understanding others' emotional states), and motivational empathy (empathic concern reflected in care for the other and willingness to invest effort in their well-being). Our data confirms that participants frequently produced messages aligning with these component dimensions. Affective empathy accounted for 25\%, cognitive empathy for 27\%, and motivational empathy for 26\% of messages. The other 22\% of messages were categorized as misattuned behaviors that normative models of empathy recommend avoiding such as giving unsolicited advice (e.g., \emph{``You just need to move on''}), dismissing emotions, and redirecting focus to oneself~\cite{goldsmith1997normative, weger2014relative, jones2004personcentered}. 

Integrating these approaches, we imposed the top-level theoretical categories (Affective, Cognitive, Motivational, and Misattuned) onto the 128 kSAE identified themes. Three human annotators then performed qualitative coding to organize the themes into two intermediate hierarchical layers, creating a tree structure with meaningful subcategories (e.g., under Affective: ``Validating Emotions'' as a mid-level node grouping clusters such as ``Naming emotions'' and ``Validating emotional experience''). The resulting taxonomy is illustrated in Fig.~\ref{fig:hierarchy}, demonstrating how bottom-up discovery of linguistic concepts through SAEs align with top-down theoretical constructs, offering a data-driven foundation for understanding the idiomatic and thematic structure of empathic messages in digital contexts while extending existing theory. Supplementary Information presents all kSAE-identified themes and corresponding theoretical categories. This taxonomy provides a lens for examining training effects by revealing which specific communicative moves participants adopted or reduced after coaching.

\subsection*{Personalized Feedback Boosts Empathic Communication}

\begin{figure}[h]
    \centering

    \begin{subfigure}{.98\textwidth}
        \subcaption{}
        \includegraphics[width=\linewidth]{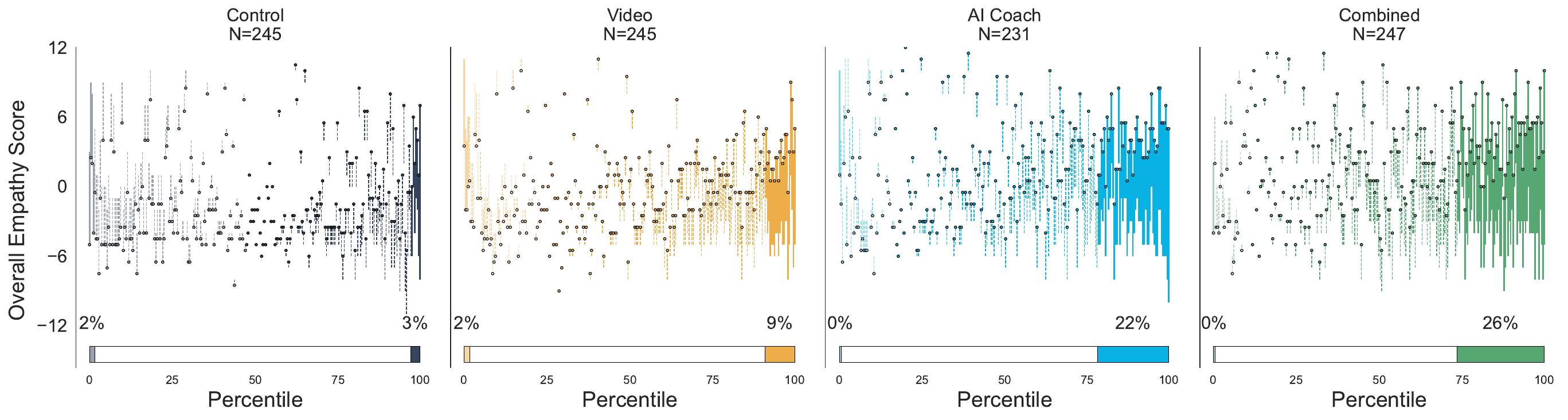}
    \end{subfigure}

    \begin{subfigure}{.98\textwidth}
        \subcaption{}
        \includegraphics[width=\linewidth]{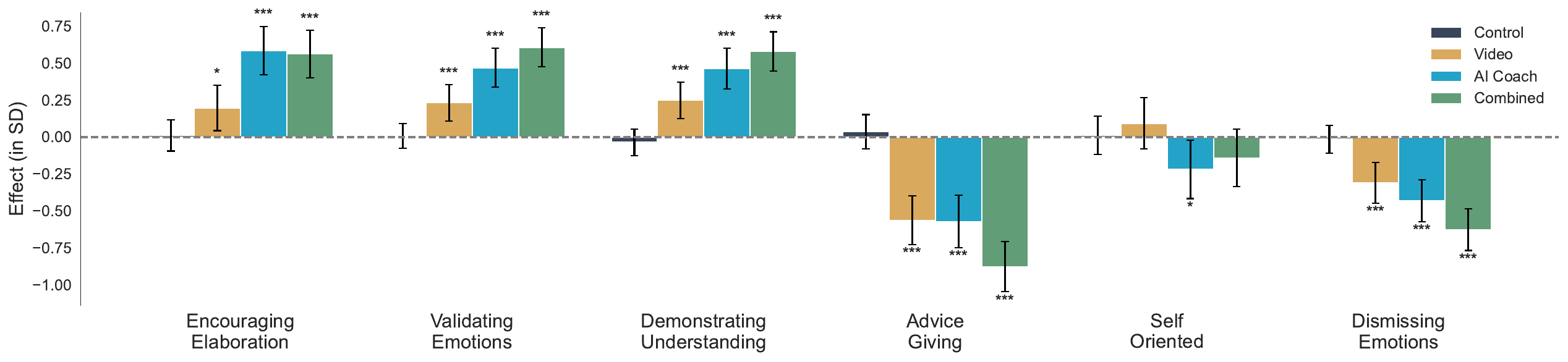}
    \end{subfigure}

    \caption{\textbf{Change in empathic communication performance.} \textbf{A.} Each vertical line represents one participant, connecting their baseline score (conversation 1) to their post-intervention score (mean of conversations 2-3), represented by a dot. Dotted lines indicate change; darker lines indicate reliable change. The y-axis is the overall empathy score calculated as the sum of prescriptive behavior ratings minus proscriptive behavior ratings. The x-axis ranks participants by percentile, sorted by magnitude of change within each condition. Black circles mark post-intervention scores. The horizontal bar indicates the RCI threshold for reliable change. \textbf{B.} Standardized intervention effects (in SD units) on six preregistered dimensions of empathic communication. Bars show OLS regression coefficients comparing each condition to baseline, with 95\% confidence intervals. Asterisks indicate statistical significance (* $p < 0.05$, ** $p < 0.01$, *** $p < 0.001$). All analyses follow the preregistered analysis plan.}
    \label{fig:pre-post}
\end{figure}

Personalized feedback from the AI coach produced reliable individual-level improvement in performance that exceeded what would be expected from measurement error alone. We computed the Reliable Change Index (RCI) for each participant, allowing us to classify individual change as reliable improvement, reliable decline, or measurement noise. In the control condition, only 4.5\% of participants exceeded the RCI threshold in either direction (2.9\% improved, 1.6\% declined), indicating that most observed variation reflected noise rather than true change in empathic communication performance. Video instruction produced marginal improvement (9.0\% improved, 1.6\% declined). In contrast, personalized feedback and combined training produced higher rates of reliable improvement: 21.6\% and 26.3\% respectively with near-zero decline (0.4\% each). Fig.~\ref{fig:pre-post}A shows the change in overall empathic performance score for all participants in each experimental condition. Supplementary Information shows individual trajectories (light gray) from pre- to post-intervention across conditions.

In addition to reliable individual improvement, personalized feedback from the AI coach produced significant gains across all six preregistered dimensions of empathic communication. Fig.~\ref{fig:pre-post}B shows intervention effects in standard deviation units for all conditions and empathic behaviors. Personalized feedback from the AI coach improved all three prescriptive behaviors relative to control (Encouraging Elaboration $\beta = 0.59$, Validating Emotions $\beta = 0.47$, Demonstrating Understanding $\beta = 0.46$; all $p < 0.001$). Combined training produced
comparable gains (Encouraging Elaboration $\beta = 0.56$, $p < 0.001$; Validating Emotions $\beta = 0.61$,  $p < 0.001$; Demonstrating Understanding $\beta = 0.58$,  $p < 0.001$).  
Video instruction also produced significant improvements relative to control, though smaller in magnitude, across Encouraging Elaboration ($\beta = 0.20$, $p < 0.05$), Validating Emotions ($\beta = 0.23$, $p < 0.001$), and Demonstrating Understanding ($\beta = 0.25$, $p < 0.001$). Pairwise comparisons confirmed that AI coach and combined training significantly outperformed video instruction on all three prescriptive behaviors (see Supplementary Information for detailed pairwise statistics). AI coach and combined training did not significantly differ from each other on any prescriptive behavior. 

Personalized feedback from the AI coach significantly reduced all three proscriptive behaviors, including Advice Giving ($\beta=-0.57$, $p<0.001$), Dismissing Emotions ($\beta=-0.43$, $p<0.001$), and Self-Oriented responses ($\beta=-0.22$, $p<0.05$). Combined training significantly reduced Advice Giving ($\beta=-0.88$, $p<0.001$) and Dismissing Emotions ($\beta=-0.62$, $p<0.001$), but did not significantly affect Self-Oriented responses. Video instruction also significantly reduced Advice Giving ($\beta=-0.56$, $p<0.001$) and Dismissing Emotions ($\beta=-0.31$, $p<0.001$), but did not significantly affect Self-Oriented responses. Notably, in a 2 by 2 factorial analysis, Advice Giving was the only outcome to show a significant AI-by-video interaction ($\beta=0.26$, $p=0.039$), indicating that the combined condition reduced advice giving less than would be expected if the separate AI and video effects were additive.

We find significant post-baseline main effects of AI feedback on all six dimensions and of video instruction on four of six dimensions (Supplementary Information, Table 1). Personalized AI feedback produces significantly larger gains than video instruction on four of six dimensions (all three prescriptive dimensions and one of three proscriptive dimensions). Pairwise comparisons further show that combined training outperformed video instruction on all six dimensions and outperformed the AI coach on two dimensions (Supplementary Information, Table 2). See Supplementary Information for additional analyses of empathic communication performance across workplace and personal scenarios.

By combining all 6 dimensions into a single metric, we can get overall effects of each of the interventions. The video instruction produced a 0.55 SD increase, the AI coach produced a 0.98 SD increase, the combined intervention produced a 1.26 SD increase. For perspective, 1 SD increase is equivalent to a 2.9 point gain on the overall empathy score. Extended Data Figure 2 shows the distribution of change in overall empathy score for each of the four conditions.

\subsection*{Coached Participants Adopted More Empathic Strategies}

\begin{figure}[h]
    \centering
   \includegraphics[width=0.95\textwidth]{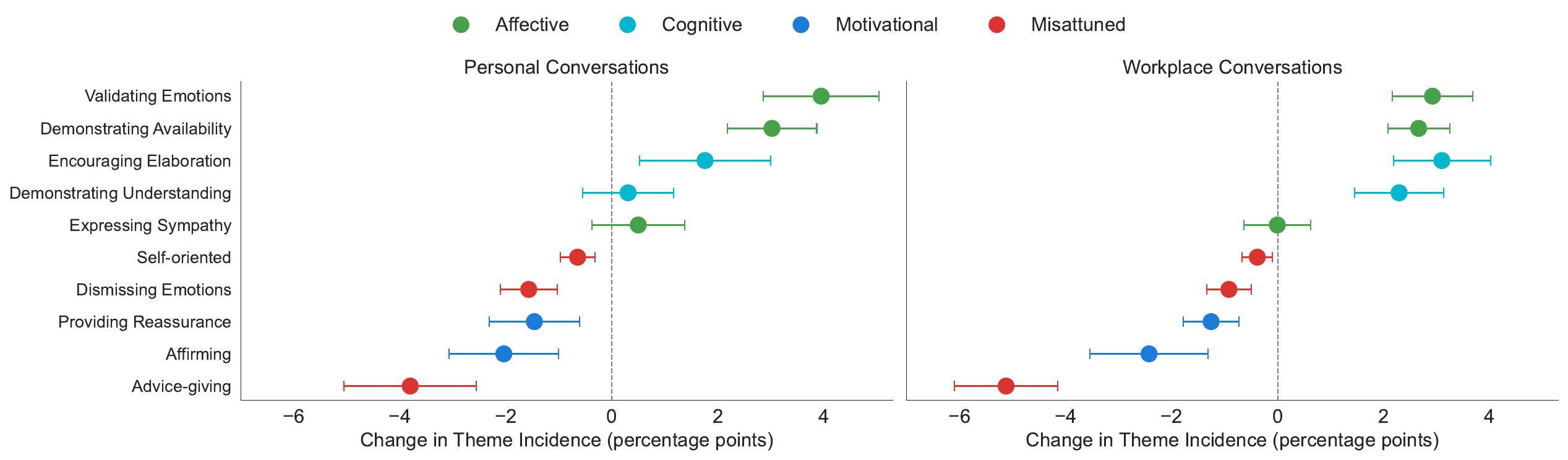}
    \caption{\textbf{Change in empathic communication behaviors.} Change in incidence of empathic communication strategies after feedback from AI Coach in Lend an Ear, shown separately for personal (left) and workplace (right) conversation contexts. Dots represent estimated change in percentage points shown with 95\% confidence intervals. Categories are color-coded by type: Affective (green), Cognitive (cyan), Motivational (blue), and Misattuned (red). }
    \label{fig:ai-coach-strategy-change}
\end{figure}

AI coaching led participants to adopt communicative strategies that aligned with normative models of empathic communication. Fig.~\ref{fig:ai-coach-strategy-change} shows that relative to first conversations (all conditions), post-training conversations (conversations 2 and 3) among AI-coached participants showed higher incidence of empathic strategies: validating emotions increased by 3.9 percentage points (personal, $p < 0.001$) and 2.9 percentage points (workplace, $p < 0.001$), demonstrating availability increased by 3.0 and 2.7 percentage points (both $p < 0.001$), and encouraging elaboration increased by 1.8 and 3.1 percentage points (both $p < 0.001$). In contrast, misattuned behaviors declined, including advice-giving (-3.8 and -5.1 percentage points, both $p < 0.001$) and dismissing emotions (-1.6 and -0.9 percentage points, both $p < 0.001$). 

These shifts were not explained by participants producing longer conversations or engaging in more turns. The total length of conversations and overall engagement levels (as measured by turn counts) remained comparable between pre- and post-training, suggesting that quality rather than quantity of support was the primary change. We find no significant difference in turn counts (personal: 11.82 to 11.57, p=.268; workplace: 11.46 to 11.17, p=.086), total words per conversation (personal: 202.37 to 207.22, p=.644; workplace: 195.47 to 203.08, p=.345), or mean response times (personal: 65.88 to 65.22 seconds, p=.807; workplace: 67.37 to 65.85 seconds, p=.499). This suggests that training changed what participants said, shifting from misattuned to helpful empathy behaviors, without altering their overall level of engagement. As an example, Extended Data Figure 1 illustrates the first (pre-training) and third (post-training) conversations of a participant in the AI coach condition.

The AI coach's feedback focused on the same communicative dimensions on which participants later improved. We coded 4,864 coach-feedback sentences from 956 coach-feedback sessions (2 each for 231 AI Coach and 247 Combined condition participants) using GPT-4o according to the six empathic communication dimensions: validating emotions, encouraging elaboration, demonstrating understanding, avoiding unsolicited advice, avoiding self-orientation, and avoiding dismissive responses. The most common suggestions were on validating emotions (32.7\%), discouraging advice-giving (24.2\%), and encouraging elaboration (21.2\%), followed by demonstrating understanding (11.2\%), discouraging dismissiveness (5.3\%), and discouraging self-orientation (4.3\%). 28.9\% of all sentences also included other content such as praise or general evaluation. The AI coach's initial feedback was similar in length across conversations and conditions (mean = 121.7 words, SD = 13.8).

Participants' follow-up questions for the AI coach were mostly help-seeking questions about how to improve their responses. 54.9\% were help-seeking, including questions like ``How do I encourage elaboration?", ``How do I validate emotions?", and ``What can I do better?". Other common themes were gratitude or acknowledgment of the coach's feedback (18.8\%), including responses like ``Thank you" and ``This is helpful," and requests for further evaluation (16.0\%), such as ``How was my performance?", ``How did I perform?", and ``What did I do wrong?".

\subsection*{AI Coaching Did Not Homogenize Human Responses}

We find evidence that some participants incorporated short fragments of AI coach-suggested wording but rarely copied the coach’s recommended phrases verbatim. In a comparison between the coach's feedback and participants' responses in a following conversation, we find participants adopted full recommended phrases (e.g. ``How are you coping with everything right now?" or ``What has been the hardest part for you so far?'') in only 1.3\% AI Coach conversations and only one instance in Combined conversations, respectively. However, shorter overlap was more common. In 26\% and 23\% of conversations in the AI Coach and Combined conditions, we find participants reused at least one exact trigram from a recommended example phrase, such as ``it sounds like" or ``tell me more". Exact four-gram overlap appeared in 12\% and 10\% of AI Coach and Combined condition conversations, such as ``what do you think" and ``can you tell me". 

\begin{figure}
    \centering
    \begin{subfigure}[t]{.65\linewidth}
        \centering
        \subcaption{}
        \includegraphics[width=\linewidth]{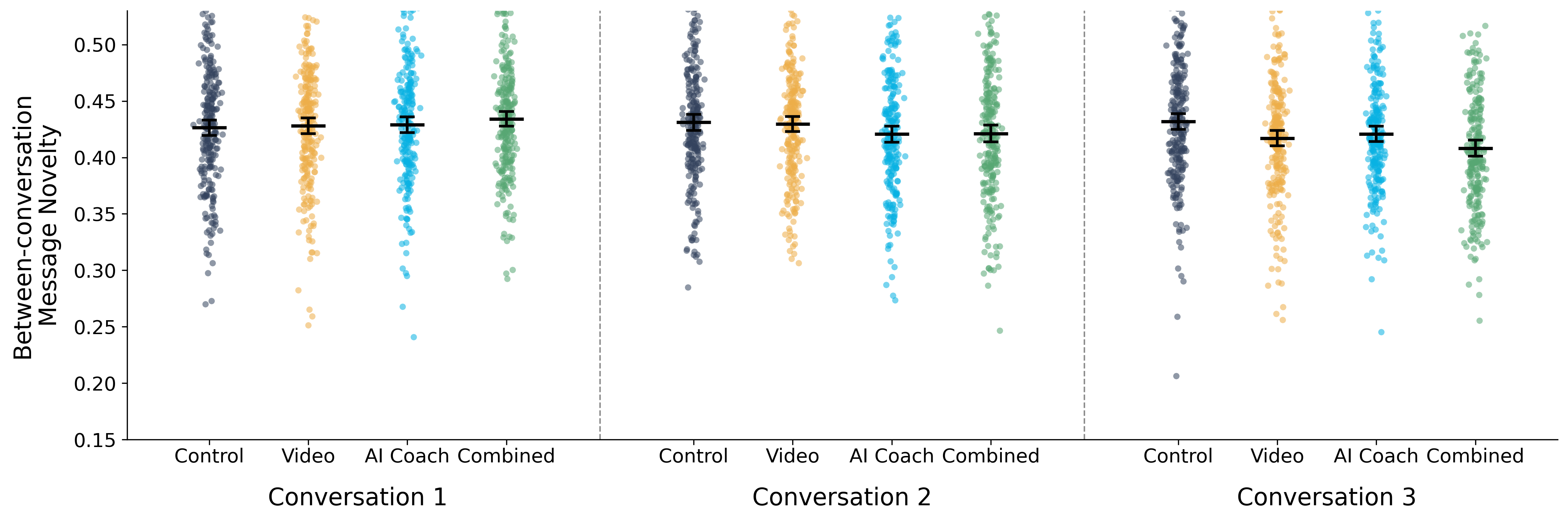}
    \end{subfigure}
    \begin{subfigure}[t]{.34\linewidth}
        \centering
        \subcaption{}
        \includegraphics[width=\linewidth]{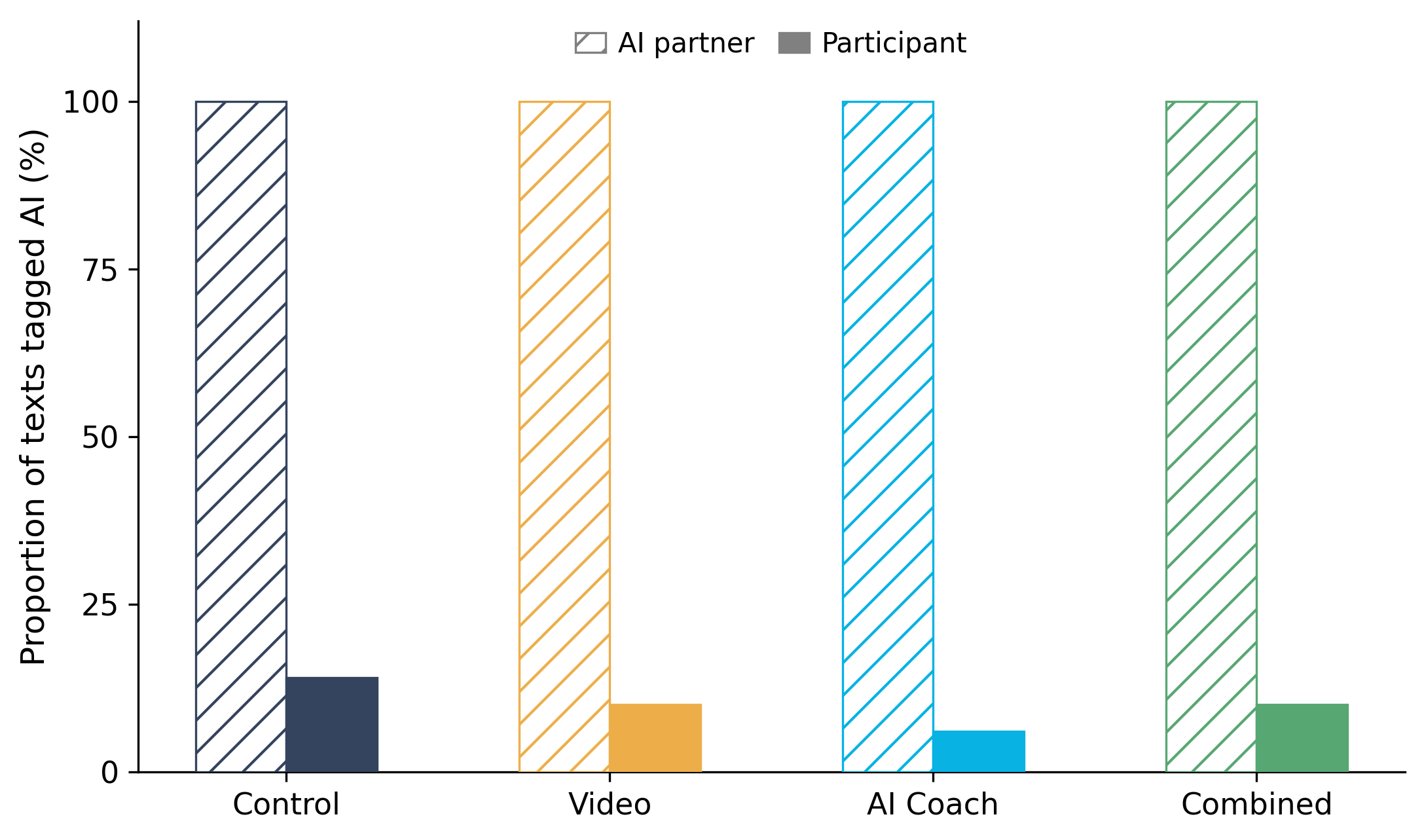}
    \end{subfigure}
    \begin{subfigure}[t]{\linewidth}
        \centering
        \subcaption{}
        \includegraphics[width=\linewidth]{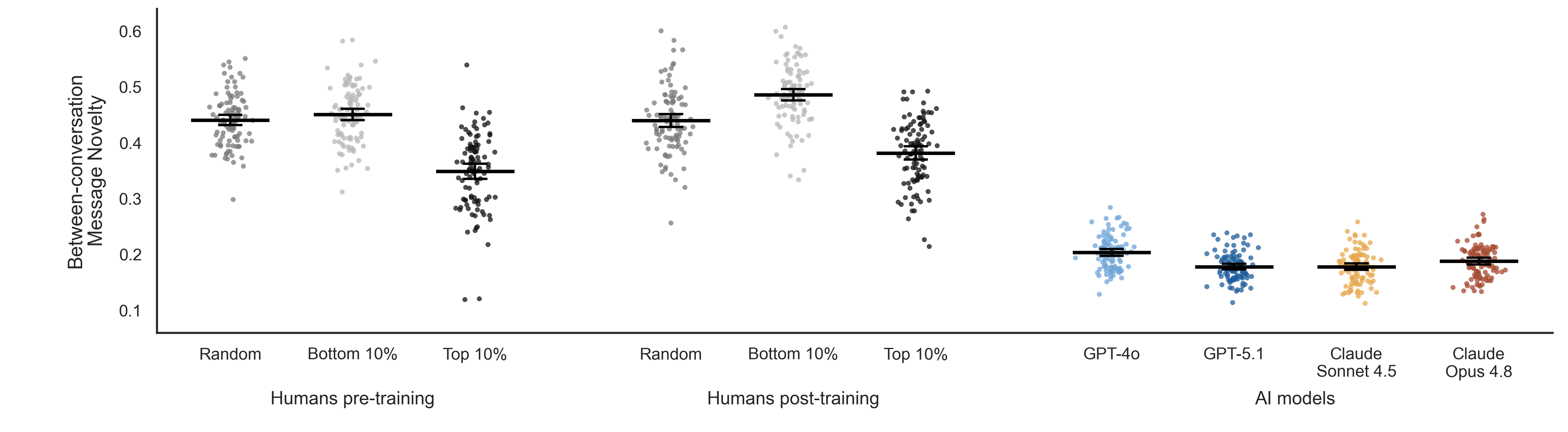}
    \end{subfigure}

    \caption{\textbf{AI coaching did not make participants' responses homogeneous or AI-like.} A. Between-participant semantic novelty by condition, conversation number, and scenario. For each supporter message, we identified the most semantically similar message written by another participant in the same condition, conversation number and scenario and defined novelty as one minus the maximum cosine similarity. Black horizontal lines show means and error bars show 95\% confidence intervals. B. Pangram AI-detection classifications for sharer (AI conversational partner) and supporter (human participant) text by condition. Bars show the proportion of texts tagged as AI defined as Pangram AI-like score $\geq$ 50\%. C. Between-participant semantic novelty for six human comparison populations and four simulated AI-supporter populations. Black horizontal lines show means and error bars show 95\% confidence intervals.}
    \label{fig:templatic}
\end{figure}

We find limited evidence of participants' responses converging after training. A semantic novelty analysis showed that responses remained distant from their nearest neighbor in embedding space across all conditions and conversations (Fig.~\ref{fig:templatic}A). For each supporter message, we identified the most semantically similar message written by another participant in the same condition, conversation number, and scenario, and defined novelty as one minus this maximum cosine similarity. We then averaged message-level novelty within each conversation. Median novelty scores were similar across conditions and conversations, ranging from 0.402 to 0.435. In Conversation 1, novelty did not differ by condition. In Conversation 2, novelty was not significantly lower in the AI Coach and Combined conditions than Control (AI Coach: $\Delta = -0.010$, FDR-adjusted $p = .093$; Combined: $\Delta = -0.010$, FDR-adjusted $p = .106$). In Conversation 3, novelty was significantly lower than Control in the Video condition ($\Delta = -0.015$, FDR-adjusted $p = .015$) and the Combined condition ($\Delta = -0.024$, FDR-adjusted $p < .001$), but not in the AI Coach condition ($\Delta = -0.011$, FDR-adjusted $p = .085$). Participants in training conditions therefore showed no significant convergence in later conversations except Conversation 3 for the Video and Combined conditions. However, the spread of novelty was preserved, with interquartile ranges spanning 0.063 to 0.077 novelty units and no significant differences in variance across condition-by-conversation cells (Brown-Forsythe test, $p = .796$). This pattern is consistent with participants adopting shared response strategies after training, and the small effect sizes and comparable variance suggest that responses did not become homogenized.

We benchmarked what homogenization would look like if supporters began to sound like an LLM by simulating AI supporters in the same role-playing task, generating 100 conversations per model across five trouble scenarios with 20 repetitions each, using GPT-4o, GPT-5.1, Claude Sonnet 4.5, and Claude Opus 4.8 (see Supplementary Information for details). These simulated conversations were less novel than human samples (Fig.~\ref{fig:templatic}C; Welch's $t$-tests, all human-group versus AI-model comparisons, all Bonferroni-corrected $p < 0.001$). Top-decile human conversations were also less novel than random and bottom-decile conversations in both pre-training and post-training samples (two-sided Welch's $t$-tests, Bonferroni-corrected $p < 0.001$). Mean between-participant novelty for the AI models ranged from 0.178 to 0.203, compared to 0.349-0.382 for top-decile human conversations, 0.440-0.441 for random human conversations, and 0.450-0.486 for bottom-decile human conversations across pre-training and post-training samples. The AI models produced high-scoring responses, but they converged to similar phrasing and tactics (see Supplementary Information). Human participants, including those trained by the AI coach, did not move toward this AI-like homogenization pattern.

AI coaching also did not lead participants to adopt LLM-like empathic templates documented in prior work. We tagged participants' messages using five templatic response styles that characterize LLM-generated empathic responses \cite{gueorguieva2026ai}, mapping each conversation to an ordered sequence of empathic tactic codes. The template with an opening move of sympathy or paraphrasing/validation followed by advice, information, or further paraphrasing appeared in 12.5\% of conversations across conditions. However, its prevalence did not rise across successive conversations in the AI Coach condition (13.0\%, 13.9\%, 12.6\%) or the Combined condition (17.8\%, 11.3\%, 16.6\%).  The other four templates were present in less than 5\% of the conversations.

We find no evidence that AI Coach or Combined participants produced more AI-like responses than Control participants. We ran Pangram, a leading AI-writing detection classifier~\cite{jabarian2025artificial}, on a random subsample of 50 conversations from each condition. We classified a text as AI-generated if its Pangram AI-like score was at least 50\%. Pangram classified 100\% of the AI-generated sharer texts as AI across all four conditions (See Figure \ref{fig:templatic}B). In contrast, 90\% of participant supporter texts were classified as human-written.The share of participant supporter texts tagged as AI was low in all conditions (Control = 14.0\%, Video = 10.0\%, AI Coach = 6.0\%, Combined = 10.0\%).

\subsection*{Felt Empathy Does Not Predict Expressed Empathy}
\begin{figure*}
    \begin{subfigure}{\linewidth}
        \subcaption{}
        \includegraphics[width=\linewidth]{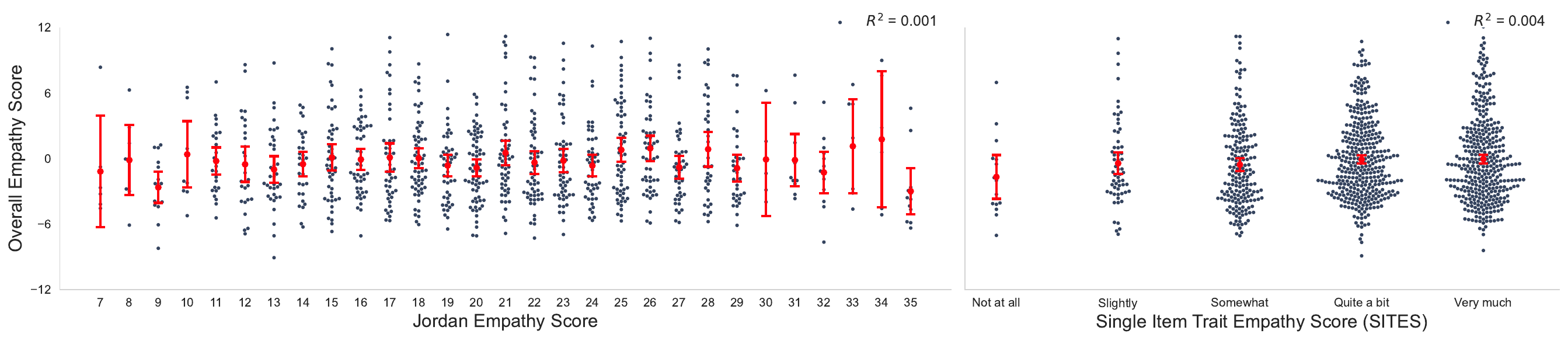}
    \end{subfigure}
    \begin{subfigure}{\linewidth}
        \subcaption{}
        \includegraphics[width=\linewidth]{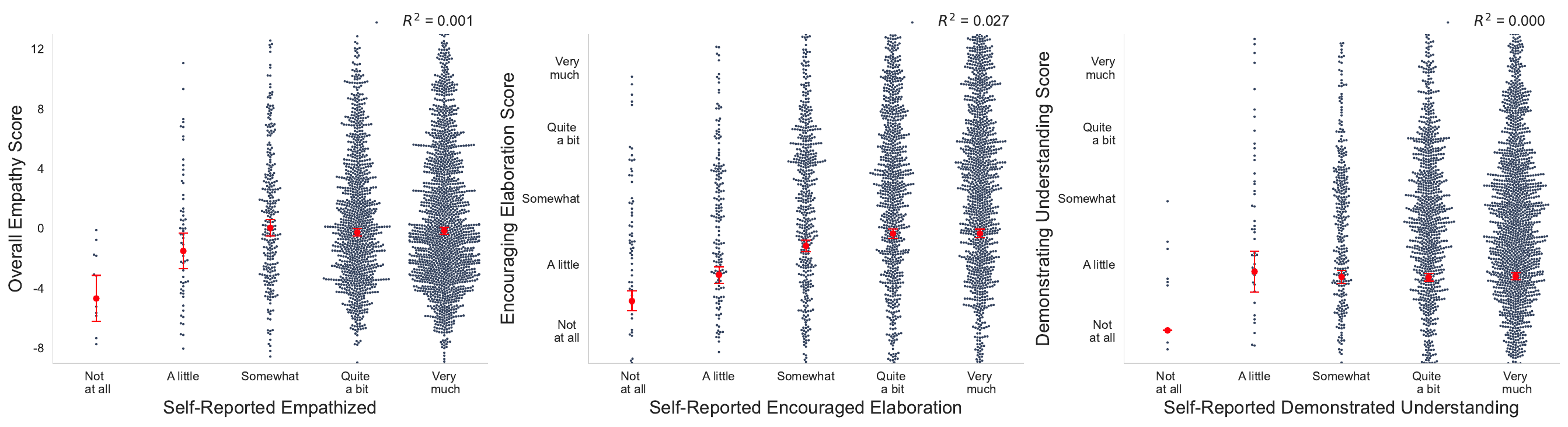}
    \end{subfigure}

    \caption{\textbf{Disconnect between expressed empathy and felt empathy} A) Relationship between trait empathy scores (Jordan Empathy Scale \cite{jordan2016empathy} and SITES \cite{konrath2018development}) and LLM-evaluated empathic communication performance across six behavioral dimensions. Each gray point represents an individual participant, with red dots indicating mean trait empathy scores and error bars showing 95\% confidence intervals. B) Relationship between LLM-evaluated performance and participant self-reported empathic communication performance across three sub-components. Each point represents an individual participant's response and the corresponding LLM evaluation. }
    \label{fig:traitvsexpressed}   
\end{figure*}

We find strong evidence of a lack of a relationship between self-reported trait empathy and empathic communication performance. We measured trait empathy using the Jordan empathy subscale \cite{jordan2016empathy} and the single-item trait empathy scale \cite{konrath2018development}, and evaluated empathic communication performance using LLM raters, which prior work has shown to approach expert-level evaluation reliability \cite{kumar2026large}. Across both trait empathy measures, we find near-zero correlations with overall empathic communication performance for 968 participants, with $R^2$ values ranging from 0.000 to 0.004 (Figure~\ref{fig:traitvsexpressed}A). Trait empathy may reflect an individual's capacity for emotional resonance \cite{jordan2016empathy}, but it does not reliably translate to skilled empathic communication in conversation.

This disconnect extends beyond trait measures to participants' reflections on their own communicative performance. Participants consistently overestimated their empathic communication abilities, rating themselves more favorably than LLM evaluators. Figure~\ref{fig:traitvsexpressed}B illustrates the relationship between LLM-evaluated scores and participant self-reports on three post-conversation reflections on empathy (``I empathized with my conversational partner's experiences and feelings''), demonstrating understanding (``I showed that I understood my conversational partner's pain and emotions''), and encouraging elaboration (``I encouraged my conversational partner to tell me more about their situation''). We find that 74\% and 87\% of participants reported encouraging elaboration and demonstrating understanding ``quite a bit'' or ``very much,'' while LLM evaluators rated only 18\% and 9\% as doing so effectively. Notably, participants' self-assessments were disconnected from their actual performance rather than merely inflated. Across 2,904 post-conversation reflections from 968 participants, we do not find self-reports on any of the three dimensions to be associated with LLM evaluated performance ($R^2=0.001$, $R^2<0.001$, and $R^2=0.027$ for overall empathy, demonstrating understanding, and encouraging elaboration respectively). This result remains robust to alternative specification such as when we restrict the analysis to self-reflections after the first conversation. People believed they communicated empathically because they felt empathy, unaware that feeling and expressing empathy represent distinct competencies. 

These results reveal a fundamental disconnect between the ability to feel empathy and the ability to express it effectively. This challenges the assumption that empathic behavior flows naturally from empathic disposition~\cite{depow2025individual}. Our findings instead point to empathic communication as a performative competence requiring mastery of a specific communicative idiom that involves strategies such as validating emotions, encouraging elaboration, and demonstrating understanding. Individuals may feel others' pain yet lack fluency in this idiom. Our coaching intervention does not attempt to teach people to feel others' feelings, but rather to develop competence in the communicative practices that convey empathy effectively.

\subsection*{People Prefer Conversations that Follow Established Frameworks in Empathic Communication}
\begin{figure}[h]
    \centering

    \begin{subfigure}{.95\linewidth}
        \subcaption{}
        \includegraphics[width=\linewidth]{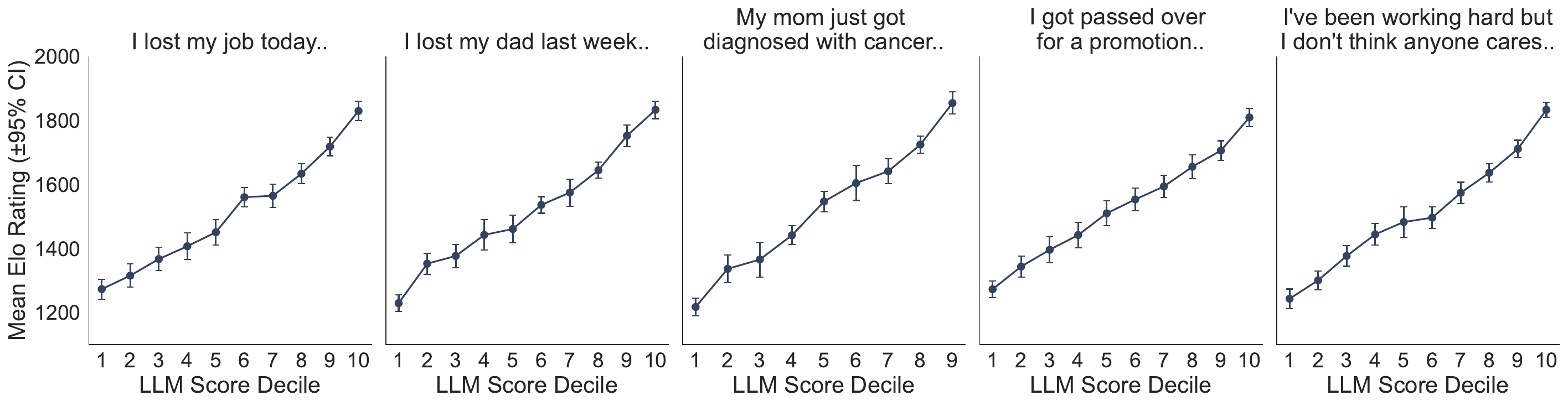}
    \end{subfigure}

    \begin{subfigure}{.475\linewidth}
        \subcaption{}
        \includegraphics[width=\linewidth]{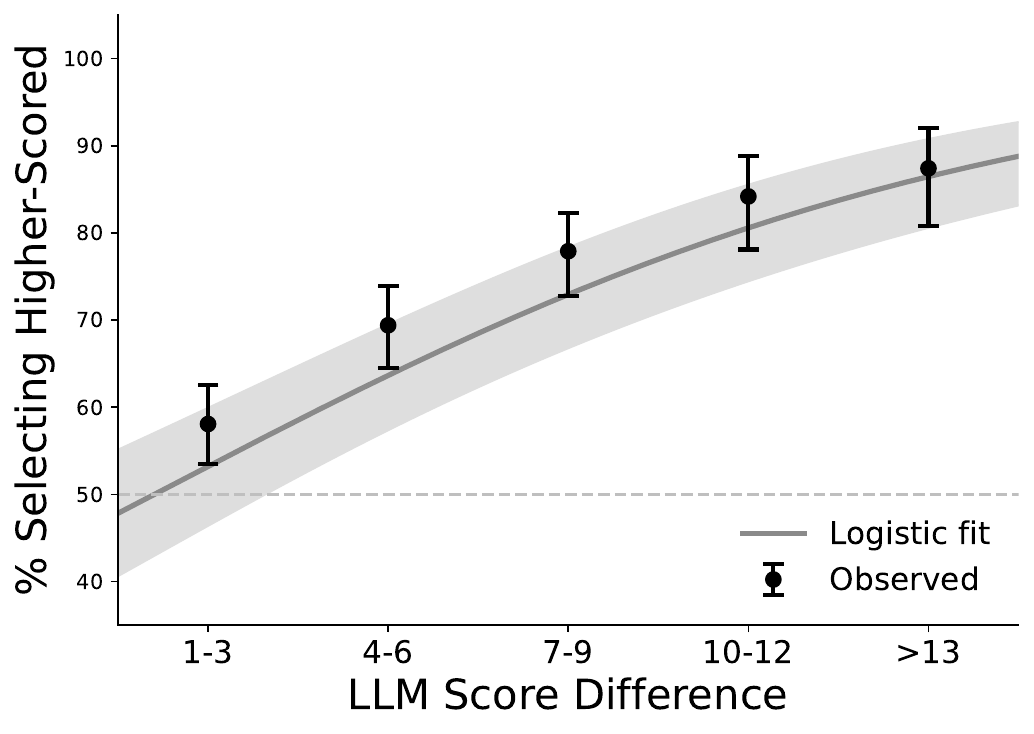}
    \end{subfigure}
    \begin{subfigure}{.475\linewidth}
        \subcaption{}
        \includegraphics[width=\linewidth]{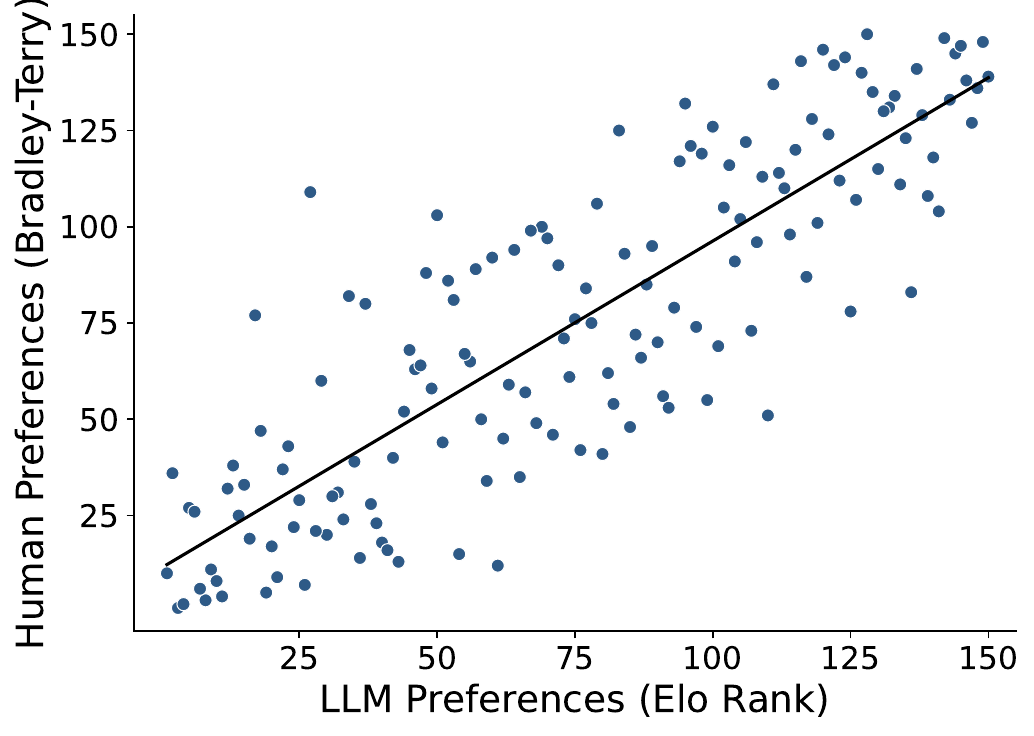}
    \end{subfigure}

    \caption{\textbf{Human preferences align with LLM evaluations of empathic communication quality.} \textbf{A.} Mean Elo ratings (derived from LLM pairwise judgments) increase monotonically with LLM empathy score deciles across all five conversation scenarios. \textbf{B.} Participants increasingly selected conversations with higher LLM empathy scores as the empathic quality gap widened. Points show observed selection rates (with 95\% Wilson confidence intervals) binned by LLM score difference; gray line shows logistic regression fit (OR = 1.15 per point, 95\% CI [1.11, 1.20]). \textbf{C.} Human preference rankings derived from a Bradley-Terry model strongly correlate with LLM Elo rankings (Spearman $\rho$ = 0.85, $p < .001$, $N$ = 150 conversations).}
    \label{fig:preference_elo}
\end{figure}

A follow-up preregistered experiment confirmed that independent human raters prefer the same conversations that LLM evaluators score higher. We recruited 150 participants via Prolific to perform two-alternative forced-choice comparisons for 150 conversations sampled from the Lend an Ear dataset. This sample was constructed by selecting 3 conversations from each decile of Elo ratings derived from pairwise LLM judgments (resulting in 30 conversations per scenario across five scenarios; see Supplementary Information for Elo rankings per decile and Methods for details). On each trial, participants viewed a pair of conversations from the same trouble scenario and selected which demonstrated better empathic communication. All conversations had been previously assessed by LLM judges across six sub-components on a 5-point Likert scale and ranked using Elo ratings from pairwise judgments elicited from an LLM evaluator.

Participants preferred conversations which LLMs scored higher, and this preference increased with the empathic quality gap. When two conversations differed by just 1 point on the LLM evaluation scale, participants were at chance, selecting the higher-scored conversation only 51\% of the time. A 5 point difference increased the observed selection rate to 73\%, and a 10 point difference to 93\%. A logistic regression predicting selection of the higher-scored conversation from LLM score differences (higher minus lower) showed a positive association ($\beta=0.144$, $p<.001$; $\mathrm{OR}=1.155$, 95\% CI [1.11, 1.20]; See Supplementary Information). Model-predicted probabilities were 51\%, 65\%, and 79\% at 1, 5, and 10 point differences, respectively. Preregistered supplementary analyses confirmed that larger LLM score gaps predict greater human–LLM agreement, and no individual sub-component score difference significantly predicted  agreement (see Supplementary Information for details).

We also estimated human-preference rankings from participants’ pairwise choices using a Bradley-Terry (BT) model for each scenario, and compared these to LLM-derived rankings. BT rankings closely matched LLM assessments (Spearman $\rho=0.75$ with LLM empathy scores; $\rho=0.85$ with Elo ratings; $p < 0.001$ for both; See Supplementary Information), indicating  agreement in which conversations are judged better. These results suggest that the LLM-judged dimensions grounded in normative models of empathic communication competence capture qualities that humans prefer and value, supporting the validity of LLMs as scalable evaluators of empathic communication in conversations.

\section*{Discussion}
Our study demonstrates that empathic communication operates as a learnable idiom, a set of conversational moves that constitute empathic response, and that brief LLM-powered interventions can teach people this idiom at scale. Based on 2,904 conversations between 968 participants and their LLM role-playing partners, we find that participants who received personalized LLM feedback quickly learned the empathic communication moves and as a result outperformed those who received no coaching, across six preregistered dimensions of empathic communication. Personalized feedback from an AI coach produced larger gains than control across all six dimensions and larger gains than video instruction on four of the six dimensions. These results provide empirical evidence that AI systems can serve as coaches for developing empathic communication skills, by making explicit what effective response patterns look like and offering structured opportunities to rehearse it. 

Building on prior work using LLMs for evaluating empathic communication \cite{kumar2026large}, we compared participants' self-reported empathic disposition with LLM evaluations of their responses. The results show a deep disconnect between participants self-reported trait empathy, their self-reported empathic performance, and their observed performance. In self-reported assessments, participants consistently overestimated their empathic abilities. Likewise, trait empathy scores as measured by Jordan empathy subscale \cite{jordan2016empathy} and SITES \cite{konrath2018development}, showed only weak correlations with performance. However, participants improved with coaching, suggesting that while people may feel empathy and wish to comfort others, they lack the communicative tools to translate these intentions into empathic responses. This challenges the assumption that empathic behavior naturally follows from empathic disposition and points to a silent empathy effect, where individuals experience empathy but struggle to express it effectively. Encouragingly, recipients tend to apply a generous threshold when evaluating responses and the detection of the intent to comfort may be sufficient to produce the experience of feeling heard. This suggests that brief instruction in basic empathic communication practices may suffice for everyday empathic communication, reserving expert-level training for high-stakes clinical or therapeutic settings.

Independent human raters validated participants' improvements in empathic communication. When asked to choose the more empathic conversation in paired comparisons, raters preferred conversations that LLMs scored higher, with preferences strengthening as the quality gap increased. At a 5-point score difference, raters selected the higher-scored conversation 73\% of the time, rising to 93\% at a 10-point difference. The average effect of the AI coach (2.9 points increase in overall empathy) corresponds to an independent observer, blind to condition, preferring the post-training response approximately two-thirds of the time. Notably, this effect was produced by a single, brief training session.  Our results also show that human-preference rankings closely matched LLM assessments (Spearman $\rho$ = 0.75 with LLM scores). These results indicate that our interventions improve performance not only on theory-driven metrics, but also on dimensions that align with what people actually prefer and value in empathic conversations. This convergence between expert-derived frameworks, LLM evaluations, and lay human preferences suggests that the improvements we observe reflect gains in empathic communication skills rather than artifacts of our measurement approach, and repeated or extended engagement with an AI coach could yield larger and more durable gains.  

Beyond demonstrating training effectiveness, the Lend an Ear platform enabled generating rich, structured conversational data that reveals the fine-grained linguistic idioms of empathic communication. Using k-sparse autoencoders, we mapped 29,520 sentences to 128 categories of empathic responses each for personal and workplace troubles contexts, and organized these within an established framework of cognitive, affective, and motivational empathy, along with misattuned behaviors that participants exhibited (e.g., dismissing feelings, giving unsolicited advice). This offers a look into the natural diversity of empathic expressions in digital text based communication, with affective empathy comprising approximately 25\% of messages, cognitive empathy 27\%, motivational empathy 26\%, and misattuned behaviors 22\% of messages. 

These findings are consistent with a performative account of empathic communication in which responding to another's distress requires learning culturally patterned responses that signal understanding and care. The 128 categories mapped using our k-sparse autoencoder analysis describe such communicative moves rather than inner states. Misattuned behaviors such as advice-giving and self-oriented responding are not failures of empathy as they often arise from a sincere desire to help. However, they represent less effective variants of the empathic idiom. What our intervention taught was the functional idiom: participants learned which moves constitute empathic responses that make others feel heard and practiced producing them. The remaining ingredient is sincere intent. It improved how participants expressed empathy without homogenizing how they expressed it. A speaker who wants to comfort, and who reproduces the functional idiom with that intent, will likely be heard as empathic. Over time the idiom may become automatic, but as we have shown, even brief exposure to the functional model can shift behavior in measurable ways.

Learning to communicate empathically from an LLM carries the risk that people may start to sound like LLMs \cite{yakura2024empirical, kobak2025delving}. LLM-generated empathy tends to converge to templatic forms \cite{gueorguieva2026ai} and generative AI assistance has been shown to homogenize outputs when used in writing \cite{padmakumar2024does} and ideation tasks \cite{doshi2024generative, anderson2024homogenization}. A useful intervention must therefore teach the functional idiom of empathy without flattening the variety of ways in which people convey it. Our findings suggest that participants in the AI coaching condition of Lend an Ear learned the idiom of empathic communication without converging to similar responses or copying the templatic style of LLMs.

The LLM-powered role-playing approach addresses limitations of conventional empathy training programs. Traditional methods require trained experts, substantial time commitments, and often operate in group-based settings that make personalized feedback difficult to deliver at scale. In contrast, our system provides immediate, tailored feedback  based on individual communication patterns in a low-cost, on-demand format. This makes it possible to offer practice and coaching opportunities to a wider audience who might otherwise never access expert-guided empathic communication training.

Our investigation focuses specifically on empathic communication in low-familiarity contexts, including interactions between strangers, acquaintances or workplace colleagues, across five specific trouble scenarios (losing a job, getting passed over for a promotion, feeling undervalued at work, supporting a family member diagnosed with cancer, and grieving the death of a family member). This context constitutes interactions characterized by limited shared history and more formal communication boundaries. However, we do not examine the high-relational context of ``thick empathy''~\cite{crockett2025empathy} that includes romantic partnerships, family, or close friends, which involve shared experience, long term relationships, potential power structures,  communication norms, and other unique dynamics. For instance, empathic communication between spouses and friends may be fundamentally different (e.g. drawing on references to shared experiences and future planning) than support between colleagues or strangers that require different navigation of professional and personal boundaries. 

Crucially, skill at empathic communication is not fundamentally different across these contexts. It always involves learning conventional patterns for conveying concern. The idiom for empathy in intimate conversations differs from the idiom for coworker conversations, but the underlying skill set is the same. Extending our training approach to high-relational contexts would thus require identifying which empathic communication components or which idioms are appropriate within specific relationships, as well as how factors like relationship history, emotional intimacy, power dynamics, and cultural expectations shape its contours. Our low-relational model provides a valuable baseline, demonstrating that the fundamental skill of learning and reproducing an empathic idiom can be trained, with the specific idiom adapted to context. 

In this research, we focused on the US context and future work could explore empathic communication across cultural contexts. Felt empathy for another can vary with perceived group boundaries and racial identification~\cite{mei2025reducing} and an open question remains on whether the experience of feeling heard and supported varies across cultures and social groups.

Our results may raise questions about whether empathic skills acquired through AI training can be authentic. Can trained empathic responses foster genuine connection? However, this concern overlooks important realities about empathic communication in practice. First, our approach does not replace human empathy with artificial empathy. It provides structured practice opportunities to help people better express the empathy they already feel. Second, empathic communication, like any interpersonal skill, exists on a spectrum of natural ability and can be meaningfully improved through training. Healthcare professionals, therapists, and other empathy-dependent practitioners routinely receive structured training to develop more effective empathic responses. This training does not imply their caring is inauthentic. Instead, it helps them learn appropriate vocabulary, timing, and techniques, enhancing their ability to connect with and help others. AI-mediated empathy training serves the same function as human-delivered training. While expert human trainers providing personalized coaching to everyone who could benefit from it would be ideal, it is infeasible because of resource constraints. LLM-powered role-playing games with coaching offer a scalable alternative for developing these crucial interpersonal skills, making empathic communication training accessible to anyone with internet access.

\subsection*{Materials and Methods}
\subsection*{Lend an Ear}
We designed a web-based experimental platform, \textit{Lend an Ear}, using Python, Flask, Javascript, and HTML to facilitate the conversational interactions and data collection. On clicking the link to the platform, participants were directed to the landing page, where we provide informed consent. Next, they saw instructions explaining the experimental procedure and their role in the conversations. After reading instructions, participants responded to a baseline survey, which consisted of questions from the Jordan empathy subscale~\cite{jordan2016empathy} and SITES~\cite{konrath2018development} (See Supplementary Information for exact questions). Once these assessments were completed, participants were directed to their first conversation with an LLM conversational partner.

The conversation begins with an initial message from the conversation partner. Given the coordination problem where conversations rarely end when either conversant wants them to~\cite{mastroianni2021conversations}, we designed the interaction such that once a participant responds to this initial message, a four-minute timer begins to count down and the conversational partner replies.  To encourage active engagement, the timer paused if participants switched tabs and resumed when they returned. Additionally, if participants had not sent a message for more than a minute, the timer would pause and participants would receive a notification that they needed to respond to their conversational partner to continue the experiment and the timer resumed after participants sent a message. 

After each conversation, participants completed a brief self-assessment consisting of four questions evaluating their empathic responses during the interaction. Participants' treatment assignment determined their next step. Participants in the control condition proceeded directly to their next conversation, while those in treatment conditions received feedback before continuing. This process continued until all participants had completed three conversations with different LLM partners. We had three treatment conditions: 1) Video Instruction 2) LLM Coach, 3) Combined Training. 

At the end of the experiment, participants answered questions about their overall experience interacting with the conversational partners. Those assigned to treatment conditions also provided feedback about their experience with the intervention they received during the study.

\subsubsection*{Communication Coach}
We developed an LLM-powered communication coach to provide automated feedback on empathic communication skills. The coach was built using a comprehensive framework of empathic communication that we distilled from the literature on empathic communication in collaboration with a co-author who has over 20 years of professional experience training healthcare professionals in empathic communication techniques. 

The empathic communication framework incorporated six key empathic techniques including validating emotions~\cite{rizvi2016dialectical}, demonstrating understanding by paraphrasing~\cite{rogers1961becoming}, encouraging elaboration and asking open-ended rather than closed-ended questions~\cite{moyers2014motivational}, avoiding unsolicited advice~\cite{goldsmith1997normative}, not being self-oriented, and not being dismissive of the conversational partner's emotions.

We used this framework to provide real-time feedback to participants and to score their conversational performance post-hoc for analysis. The LLM-based scoring system demonstrated high inter-rater reliability with expert annotations of empathic communication across multiple evaluative frameworks, with reliability approaching that of trained human experts~\cite{kumar2026large}. The coach and LLM evaluation were both implemented using GPT-4o. The full prompt used to create the communication coach agent is available in Supplementary Information.

\subsubsection*{Training Videos}
For video instruction, we used two videos, one 35 seconds long and the other 57 seconds long featuring one of the authors who is an expert in empathic communication. At the time of writing, this author has 339,500 followers on Tiktok, and these videos have over 18,100 views and 19,400  views respectively. These videos served as engaging, accessible didactic instruction on general empathy techniques presented in a popular social media format. Links to both videos and the complete transcripts of the videos are available in Supplementary Information.

\subsection*{Troubles Scenarios and LLM Conversational Partner}
We developed five trouble talk scenarios for our LLM conversational partners, comprising three workplace troubles and two personal troubles. The workplace scenarios included: (1) a job loss scenario, where the conversational partner had recently been terminated from their position; (2) a promotion rejection scenario, where the partner had been passed over for an expected promotion; and (3) a workplace recognition scenario, where the partner felt their hard work was going unnoticed and unappreciated by colleagues and supervisors. The personal trouble scenarios consisted of: (4) a parental cancer diagnosis scenario, where the conversational partner was coping with a parent's recent cancer diagnosis; and (5) a parental loss scenario, where the conversational partner was grieving the recent death of a parent. See Supplementary Information for exact scenarios.

For each of the five troubles scenarios, the LLM role-playing agent was provided with a detailed background story establishing their identity and the specific trouble they were experiencing. The LLM partners were instructed to behave as individuals seeking to feel heard and understood, expressing their concerns and emotions in two to three sentences in multi-turn conversations. The full prompts and background narratives used to create these role-playing conversational agents are available in Supplementary Information. The role-playing agents were implemented using GPT-4o.

\subsection*{Participants}
We recruited a demographically representative sample of the U.S. population with respect to age, sex, and ethnicity through the Prolific platform. A total of 1,045 participants were initially recruited for the study. Following preregistered data cleaning procedures, we excluded participants with incomplete data who did not finish all three conversations and associated assessments (n = 51 excluded) or self-reported using AI assistance to complete the experiment (n = 26 excluded). Our final analytic sample consisted of 968 participants, yielding 2,904 total conversations.  Participants had a mean age of 45.62 years (SD = 15.67; median = 46; range = 18–86), and the sample was 51.8\% female (n = 501) and 48.2\% male (n = 467). Ethnicity was 65.3\% White (n = 632), 11.4\% Black (n = 110), 10.3\% Mixed (n = 100), 6.8\% Other (n = 66), and 6.2\% Asian (n = 60). Supplementary Information reports analyses examining associations between demographic characteristics, baseline empathic communication, and improvement over time.

Participants were randomly assigned to one of four experimental conditions: (1) control, (2) video instruction, (3) personalized feedback from an AI coach, and (4) combined training with video and personalized feedback (combining both treatment approaches). Additionally, the order in which participants encountered the trouble scenarios was randomized to control for potential ordering effects.
Participants were compensated at a rate of \$12 per hour for their participation. The experiment took approximately 20 minutes to complete on average, resulting in an average payment of \$4 per participant.

\subsubsection*{Conversation Text Analysis}
To understand the content and patterns of empathic communication in participant responses, we conducted a comprehensive analysis of the conversation data. We used k-sparse autoencoders \cite{peng2025use} to identify themes in participants' messages, analyzing personal and workplace trouble scenarios separately to capture context-specific communication patterns.

We first preprocessed the conversation data by extracting individual turns from each conversation and splitting them into sentence-level units. We filtered for supporter messages (excluding seeker turns) and removed messages shorter than 2 characters to focus on substantive responses. For each trouble type (workplace and personal), we embedded the supporter messages using OpenAI's \texttt{text-embedding-3-large} model to capture semantic content.

We trained k-sparse autoencoders on these embeddings to discover interpretable latent features. We conducted a grid search over the number of latent features, ranging from $2^4$ (16) to $2^8$ (256), to identify the optimal level of granularity for the number of neurons in the sparse hidden layer. The autoencoder architecture used $M = 128$ neurons to examine different granularities of theme extraction with a sparsity parameter of $K=2$, so each input activated at most 2 neurons. Each sentence was assigned to its top two activating features to account for polysemous sentences that could align with multiple thematic concepts. We computed silhouette scores \cite{rousseeuw1987silhouettes} on the original embeddings using top two assignments as labels, to assess whether high-activating sentences for each feature formed cohesive groups. Additionally, we manually reviewed LLM-generated descriptions for interpretability and thematic distinctiveness. We identified 128 latent features as the optimal resolution because it maximized quantitative metrics (silhouette score of 0.42 for 128, compared to 0.35 for 64 features and 0.38 for 256) while yielding interpretable, non-redundant themes that captured the diversity of empathic expressions in our dataset.

For each trained model, we identified the texts that most strongly activated each neuron and used these examples to generate human-interpretable theme labels. We provided task-specific instructions to guide the interpretation process (e.g., ``You are an empathic communication expert. These are messages that a person sends to comfort someone who has shared a workplace-related problem. Describe the broad theme of the message.''), along with representative examples of each neuron's top-activating texts.

Following the automated theme extraction, three authors collaboratively annotated the identified themes to create macro-level categories. Through an iterative coding process, we developed a hierarchical mapping of the types of supportive communication strategies participants employed across different conversational contexts as shown in Fig.~\ref{fig:hierarchy}.

Theme-level percentage-point differences were tested using two-proportion z-tests (post-training vs. pre-training incidence), with Benjamini-Hochberg false-discovery-rate correction applied across themes separately for personal and workplace analyses. 

\subsubsection*{Effect of Interventions}
Following our preregistered analysis plan, we estimated treatment effects using OLS regression with standard errors clustered at the participant level. The model included indicator variables for the three treatment conditions (video instruction, AI coach, and combined training), an indicator variable for \textit{Round} (coded as 1 for Rounds 2/3 and 0 for Round 1 to assess learning effects over time), and interaction terms between each condition and Round to examine differential effects across rounds. Round 1 served as the baseline conversation before any intervention exposure. This model was run separately for each of the six sub-components for empathic communication as the dependent variable. This analysis plan was preregistered prior to data collection. We then conducted pairwise Wald tests to compare intervention effects against each other.

We also analysed the study as a 2 (AI feedback: absent/present) x 2 (video instruction: absent/present) model and estimated post-baseline main effects, their interaction, and the direct AI-versus-video contrast (Supplementary Information, Table 3).

\subsubsection*{Reliable Change Index Analysis}

We used the Reliable Change Index (RCI) \cite{jacobson1992clinical} to determine whether individual participants showed meaningful improvement, decline, or no change. We created an overall empathy score by subtracting proscriptive behavior ratings from prescriptive behavior ratings.
To distinguish real change from measurement noise, we first estimated measurement error using the control group. We fit a random-intercept model to separate true individual differences from random fluctuations within the same person across conversations. This allowed us to compute single-measurement reliability as the ratio of true score variance to total variance.
For each participant, we calculated a change score as the difference between their baseline empathy (conversation 1) and post-intervention empathy (average of conversations 2 and 3). Because the baseline used one conversation while the post-intervention score averaged two, we adjusted the standard error of change accordingly to account for reduced error variance in the averaged score.
We classified participants using standardized change scores (z-scores) as showing reliable improvement ($z \geq 1.96$), no reliable change ($-1.96 < z < 1.96$), or reliable decline ($z \leq -1.96$). These thresholds correspond to change that exceeds what would be expected from measurement error alone at the $p \leq .05$ level.

\subsubsection*{Novelty Analysis}
We conducted a novelty analysis to test whether AI coaching made participants converge on semantically similar responses. Each supporter message was converted to an embedding vector using OpenAI's \texttt{text-embedding-3-small} model. For each message, we identified the most semantically similar message written by another participant in the same condition, conversation number, and scenario using maximum cosine similarity. We masked messages written by the same participant, ensuring that a message could not match itself or another turn from the same participant. Message-level novelty was calculated as one minus this maximum similarity. Higher values indicate that a message was farther from its nearest eligible neighbor. We aggregated message-level novelty by averaging within each participant and conversation, yielding one novelty score per participant per conversation, and then summarized these scores by condition and conversation number.

\subsubsection*{Templatic Text Analysis}
We tested whether AI coaching made participants' responses more formulaic over time by adapting the analysis from prior work on templatic empathic responses in LLMs \cite{gueorguieva2026ai}. We annotated supporter responses at the phrase-span level. Supporter turns were split into sentence- and clause-level spans; preserving question marks for detecting questions. Each span could receive one or more tactic labels. 
We mapped spans onto the tactic categories including emotional expression, paraphrasing or demonstrated understanding, validation, questioning, self-disclosure or self-oriented responding, assistance, empowerment, reappraisal or reassurance, information, and advice. Span labels were assigned using rule-based phrase patterns and existing RPG theme annotations, with Level 1 and Level 2 theme codes mapped onto the templatic-response tactic taxonomy. For each conversation, we constructed an ordered tactic sequence by sorting labels by supporter turn, span position, and within-span tactic order.  Repeated tactic labels were collapsed before template matching. We then tested whether each conversation matched any of the five templatic response patterns identified in prior work.

\subsubsection*{Automated AI-detection Analysis}
As a robustness check, we assessed whether participants' post-intervention responses became more AI-like by using Pangram Labs' AI-detection API. We randomly sampled 50 conversations per experimental condition yielding 200 conversations across the four conditions. For each sampled conversation, we submitted the AI messages and participant responses separately to the Pangram API.

\subsection*{Human Preferences Study}
In a follow-up preregistered study, we tested whether independent human raters prefer the same conversations that our LLM judges evaluate as more empathic. We sampled 150 conversations from the Lend an Ear experiment, with 30 conversations sampled from each of the five trouble scenarios (job loss, passed up for promotion, feeling undervalued at work, parent's cancer diagnosis, or loss of a parent). Each conversation had been previously scored by an LLM along six dimensions of empathic communication and summarized into an overall empathy score, as well as ranked using Elo ratings derived from 10,000 adaptively sampled pairwise LLM forced-choice judgments per scenario (50,000 total; Elo ranks initialized at 1,000; $K = 32$, see Supplementary Information for details). We recruited 183 participants via Prolific. Each participant was assigned a pre-generated sequence of 10 conversation pairs drawn from the pool of 150 conversations. On each of 10 trials, participants viewed a pair of conversations from the same scenario, presented side-by-side in randomized left–right order, and indicated, “Which conversation would make someone sharing a trouble feel more heard?” Participants received no training, examples, or feedback. From these choices we constructed two preregistered binary outcomes: (i) whether the participant selected the conversation with the higher LLM overall empathy score (“Select the Higher Annotation”) and (ii) whether the participant’s choice matched the LLM’s own forced-choice selection for that pair (“Select the Match”). The experiment took approximately 15 minutes to complete on average, resulting in an average payment of \$3 per participant. 

We preregistered a design with 150 participants, each completing 10 trials, for 1,500 trial-level observations. During data collection, one trial sequence was assigned to 34 participants instead of 1, so we recruited 33 additional participants bringing total recruitment to 183. To maintain the intended independence structure and adhere to our preregistered design, we randomly selected one participant from those assigned to the duplicated trial structure (seed = 42 for reproducibility) and excluded the remaining 33 duplicate assignments. This yielded a final sample of 150 participants with 1,500 total pairwise comparisons.

To evaluate whether LLM annotations of empathic communication align with human judgments, we conducted a logistic regression with standard errors clustered at the participant level for two preregistered dependent variables including whether participants selected the conversation with the higher LLM overall score, and whether participants' choice matched the LLM's own forced-choice selection of the more empathic conversation. Each model included the difference in LLM overall scores between the two conversations (ranging from -24 to 24) and indicator variables for conversation topics. To examine heterogeneous effects, we ran a second model replacing the overall score difference with differences in each of the six component scores. Additionally, we fit a Bradley-Terry model to rank conversations based on participants' pairwise choices and correlated these rankings with LLM scores. This analysis plan was preregistered prior to data collection.

We conducted a supplementary sensitivity analysis including all 183 participants, using clustered robust standard errors at the trial structure level to account for non-independence. The sensitivity analysis yielded results substantively identical to the preregistered analysis ($\beta = 0.146$, $p < .001$; OR = 1.157, 95\% CI [1.14, 1.18] vs $\beta = 0.144$, OR = 1.155, 95\% CI [1.11, 1.20]), confirming that participants preferred conversations rated higher by LLMs.

\subsection*{Preregistration}
The Lend an ear experiment and the follow-up human preference experiment recruited participants from Prolific and were preregistered on aspredicted.org at the following URLs: Lend an ear experiment (https://aspredicted.org/hdjx-tdsr.pdf), and human preference experiment \\ (https://aspredicted.org/t8au77.pdf). 

For the Lend an ear experiment, the preregistered primary analyses used linear regressions with clustered standard errors at the participant level to examine the effect of training condition on each of the six empathic communication criteria (validating emotions, encouraging elaboration, demonstrating understanding, unsolicited advice, self-oriented, and dismissing emotions). The model included indicator variables for each of the three treatment conditions (video instruction, AI coach, and combined training), a round indicator (Round 1 = 0, Rounds 2/3 = 1), and condition-by-round interaction terms to assess whether improvements over time differed across conditions. Fig.~\ref{fig:pre-post} presents the results of these preregistered analyses. Fig.~\ref{fig:traitvsexpressed} presents secondary preregistered analyses examining the relationship between LLM-evaluated empathic communication and self-reported empathy. Fig.~\ref{fig:hierarchy} presents exploratory analyses of communication patterns in participant conversations not specified in the preregistration.

For the human preference experiment, the preregistered analyses used logistic regressions with clustered standard errors at the participant level to examine whether human participants' forced-choice selections of the more empathic conversation aligned with LLM evaluations. Two preregistered dependent variables were examined including whether participants selected the conversation with the higher LLM overall score (Select the Higher Annotation), and whether participants' selections matched the LLM's own forced-choice response to the same question (Select the Match). Both models included the difference in overall LLM scores between the two conversations and indicator variables for conversation topic. A Bradley-Terry model was also fit to rank conversations based on participant choices and correlate those rankings with LLM scores and Elo rankings. Fig. \ref{fig:preference_elo} presents these preregistered analyses. A second preregistered regression decomposed the overall score difference into its six component scores (see Supplementary Information for details).

\subsection*{Ethics Approval}
This research complied with all relevant ethical regulations and obtained informed consent from all participants for data we collected. The Northwestern University Institutional Review Board (IRB) determined that the research met the criteria for exemption from further review. The Lend an Ear study's IRB identification number is STU00222032 and human-preference experiment's IRB identification number is STU00223043.

\backmatter

\bmhead{Supplementary information}
Supplementary Information include Materials and Methods, Supplementary Text, Extended Data Figures 1 to 5, Supplementary Figures 1 to  3, Supplementary Tables 1 to 17, and the GUIDE-LLM Checklist \cite{feuerriegelreporting}.

\bmhead{Acknowledgements}
We gratefully acknowledge feedback and comments from participants at the Kellogg MORS Brown Bag Seminar, CODE@MIT conference, University of Chicago's Communication \& Intelligence Symposium, Wharton's AI and the Future of Work conference, and the Penn State AI and Social Research: Empathic AI, Metascience, and Methodology conference. We also acknowledge funding from the Kellogg School of Management, the Ryan Institute on Complexity, and John Chiminski.

\bmhead{Competing interests}
The authors declare no competing interests.


\bmhead{Consent for publication}
Not applicable.

\bmhead{Data availability}
The data used during the current study are available in Zenodo at \href{https://doi.org/10.5281/zenodo.20703371}{https://doi.org/10.5281/zenodo.20703371}.

\bmhead{Code availability}
The code used during the current study is available in Zenodo at \href{https://doi.org/10.5281/zenodo.20703371}{https://doi.org/10.5281/zenodo.20703371}.

\bmhead{Author contributions}
A.K., B.L., and M.G. conceived the investigation; A.K., N.P., and M.G. analyzed the data; A.K. and M.G. wrote the initial manuscript; A.K., N.P., D.Y., B.L., and M.G. reviewed and edited the manuscript.

\bibliography{references}

\begin{thebibliography}{10}
\expandafter\ifx\csname url\endcsname\relax
  \def\url#1{\burl{#1}}\fi
\expandafter\ifx\csname urlprefix\endcsname\relax\def\urlprefix{URL }\fi
\providecommand{\bibinfo}[2]{#2}
\providecommand{\eprint}[2][]{\url{#2}}
\providecommand{\doi}[1]{\url{https://doi.org/#1}}
\bibcommenthead

\bibitem{goldsmith2004communicating}
\bibinfo{author}{Goldsmith, D.~J.}
\newblock \emph{\bibinfo{title}{Communicating Social Support}} Advances in Personal Relationships (\bibinfo{publisher}{Cambridge University Press}, \bibinfo{address}{Cambridge}, \bibinfo{year}{2004}).

\bibitem{zaki2015addressing}
\bibinfo{author}{Zaki, J.} \& \bibinfo{author}{Cikara, M.}
\newblock \bibinfo{title}{Addressing empathic failures}.
\newblock \emph{\bibinfo{journal}{Current Directions in Psychological Science}} \textbf{\bibinfo{volume}{24}}, \bibinfo{pages}{471--476} (\bibinfo{year}{2015}).

\bibitem{sharma2023human}
\bibinfo{author}{Sharma, A.}, \bibinfo{author}{Lin, I.~W.}, \bibinfo{author}{Miner, A.~S.}, \bibinfo{author}{Atkins, D.~C.} \& \bibinfo{author}{Althoff, T.}
\newblock \bibinfo{title}{Human--ai collaboration enables more empathic conversations in text-based peer-to-peer mental health support}.
\newblock \emph{\bibinfo{journal}{Nature Machine Intelligence}} \textbf{\bibinfo{volume}{5}}, \bibinfo{pages}{46--57} (\bibinfo{year}{2023}).

\bibitem{yin2024ai}
\bibinfo{author}{Yin, Y.}, \bibinfo{author}{Jia, N.} \& \bibinfo{author}{Wakslak, C.~J.}
\newblock \bibinfo{title}{Ai can help people feel heard, but an ai label diminishes this impact}.
\newblock \emph{\bibinfo{journal}{Proceedings of the National Academy of Sciences}} \textbf{\bibinfo{volume}{121}}, \bibinfo{pages}{e2319112121} (\bibinfo{year}{2024}).

\bibitem{zhou2024social}
\bibinfo{author}{Zhou, Y.}, \bibinfo{author}{Han, S.}, \bibinfo{author}{Kang, P.}, \bibinfo{author}{Tobler, P.~N.} \& \bibinfo{author}{Hein, G.}
\newblock \bibinfo{title}{The social transmission of empathy relies on observational reinforcement learning}.
\newblock \emph{\bibinfo{journal}{Proceedings of the National Academy of Sciences}} \textbf{\bibinfo{volume}{121}}, \bibinfo{pages}{e2313073121} (\bibinfo{year}{2024}).

\bibitem{teding2016efficacy}
\bibinfo{author}{Teding~van Berkhout, E.} \& \bibinfo{author}{Malouff, J.~M.}
\newblock \bibinfo{title}{The efficacy of empathy training: A meta-analysis of randomized controlled trials.}
\newblock \emph{\bibinfo{journal}{Journal of counseling psychology}} \textbf{\bibinfo{volume}{63}}, \bibinfo{pages}{32} (\bibinfo{year}{2016}).

\bibitem{gryglewicz2020examining}
\bibinfo{author}{Gryglewicz, K.} \emph{et~al.}
\newblock \bibinfo{title}{Examining the effects of role play practice in enhancing clinical skills to assess and manage suicide risk}.
\newblock \emph{\bibinfo{journal}{Journal of Mental Health}}  (\bibinfo{year}{2020}).

\bibitem{riess2012empathy}
\bibinfo{author}{Riess, H.}, \bibinfo{author}{Kelley, J.~M.}, \bibinfo{author}{Bailey, R.~W.}, \bibinfo{author}{Dunn, E.~J.} \& \bibinfo{author}{Phillips, M.}
\newblock \bibinfo{title}{Empathy training for resident physicians: a randomized controlled trial of a neuroscience-informed curriculum}.
\newblock \emph{\bibinfo{journal}{Journal of general internal medicine}} \textbf{\bibinfo{volume}{27}}, \bibinfo{pages}{1280--1286} (\bibinfo{year}{2012}).

\bibitem{covey20207}
\bibinfo{author}{Covey, S.~R.}
\newblock \emph{\bibinfo{title}{The 7 habits of highly effective people}}  (\bibinfo{publisher}{Simon \& Schuster}, \bibinfo{year}{1989}).

\bibitem{suchman1997model}
\bibinfo{author}{Suchman, A.~L.}, \bibinfo{author}{Markakis, K.}, \bibinfo{author}{Beckman, H.~B.} \& \bibinfo{author}{Frankel, R.}
\newblock \bibinfo{title}{A model of empathic communication in the medical interview}.
\newblock \emph{\bibinfo{journal}{Jama}} \textbf{\bibinfo{volume}{277}}, \bibinfo{pages}{678--682} (\bibinfo{year}{1997}).

\bibitem{bylund2005examining}
\bibinfo{author}{Bylund, C.~L.} \& \bibinfo{author}{Makoul, G.}
\newblock \bibinfo{title}{Examining empathy in medical encounters: an observational study using the empathic communication coding system}.
\newblock \emph{\bibinfo{journal}{Health communication}} \textbf{\bibinfo{volume}{18}}, \bibinfo{pages}{123--140} (\bibinfo{year}{2005}).

\bibitem{schumann2014addressing}
\bibinfo{author}{Schumann, K.}, \bibinfo{author}{Zaki, J.} \& \bibinfo{author}{Dweck, C.~S.}
\newblock \bibinfo{title}{Addressing the empathy deficit: beliefs about the malleability of empathy predict effortful responses when empathy is challenging.}
\newblock \emph{\bibinfo{journal}{Journal of personality and social psychology}} \textbf{\bibinfo{volume}{107}}, \bibinfo{pages}{475} (\bibinfo{year}{2014}).

\bibitem{drollinger2006development}
\bibinfo{author}{Drollinger, T.}, \bibinfo{author}{Comer, L.~B.} \& \bibinfo{author}{Warrington, P.~T.}
\newblock \bibinfo{title}{Development and validation of the active empathetic listening scale}.
\newblock \emph{\bibinfo{journal}{Psychology \& Marketing}} \textbf{\bibinfo{volume}{23}}, \bibinfo{pages}{161--180} (\bibinfo{year}{2006}).

\bibitem{sharma2020computational}
\bibinfo{author}{Sharma, A.}, \bibinfo{author}{Miner, A.}, \bibinfo{author}{Atkins, D.} \& \bibinfo{author}{Althoff, T.}
\newblock \bibinfo{title}{A computational approach to understanding empathy expressed in text-based mental health support}.
\newblock \emph{\bibinfo{journal}{Proceedings of the 2020 Conference on Empirical Methods in Natural Language Processing (EMNLP)}} \bibinfo{pages}{5263--5276} (\bibinfo{year}{2020}).

\bibitem{kumar2026large}
\bibinfo{author}{Kumar, A.} \emph{et~al.}
\newblock \bibinfo{title}{When large language models are reliable for judging empathic communication}.
\newblock \emph{\bibinfo{journal}{Nature Machine Intelligence}} \bibinfo{pages}{1--13} (\bibinfo{year}{2026}).

\bibitem{ayers2023comparing}
\bibinfo{author}{Ayers, J.~W.} \emph{et~al.}
\newblock \bibinfo{title}{Comparing physician and artificial intelligence chatbot responses to patient questions posted to a public social media forum}.
\newblock \emph{\bibinfo{journal}{JAMA internal medicine}} \textbf{\bibinfo{volume}{183}}, \bibinfo{pages}{589--596} (\bibinfo{year}{2023}).

\bibitem{sorin2024large}
\bibinfo{author}{Sorin, V.} \emph{et~al.}
\newblock \bibinfo{title}{Large language models and empathy: Systematic review}.
\newblock \emph{\bibinfo{journal}{Journal of Medical Internet Research}} \textbf{\bibinfo{volume}{26}}, \bibinfo{pages}{e52597} (\bibinfo{year}{2024}).

\bibitem{inzlicht2024praise}
\bibinfo{author}{Inzlicht, M.}, \bibinfo{author}{Cameron, C.~D.}, \bibinfo{author}{D’Cruz, J.} \& \bibinfo{author}{Bloom, P.}
\newblock \bibinfo{title}{In praise of empathic ai}.
\newblock \emph{\bibinfo{journal}{Trends in Cognitive Sciences}} \textbf{\bibinfo{volume}{28}}, \bibinfo{pages}{89--91} (\bibinfo{year}{2024}).

\bibitem{li2024skill}
\bibinfo{author}{Herderich, A.} \& \bibinfo{author}{Goldenberg, A.}
\newblock \bibinfo{title}{Skill but not effort drive gpt overperformance over humans in cognitive reframing of negative scenarios} .

\bibitem{ovsyannikova2025third}
\bibinfo{author}{Ovsyannikova, D.}, \bibinfo{author}{de~Mello, V.~O.} \& \bibinfo{author}{Inzlicht, M.}
\newblock \bibinfo{title}{Third-party evaluators perceive ai as more compassionate than expert humans}.
\newblock \emph{\bibinfo{journal}{Communications Psychology}} \textbf{\bibinfo{volume}{3}}, \bibinfo{pages}{4} (\bibinfo{year}{2025}).

\bibitem{rubin2025comparing}
\bibinfo{author}{Rubin, M.} \emph{et~al.}
\newblock \bibinfo{title}{Comparing the value of perceived human versus ai-generated empathy}.
\newblock \emph{\bibinfo{journal}{Nature Human Behaviour}} \bibinfo{pages}{1--15} (\bibinfo{year}{2025}).

\bibitem{perry2023ai}
\bibinfo{author}{Perry, A.}
\newblock \bibinfo{title}{Ai will never convey the essence of human empathy}.
\newblock \emph{\bibinfo{journal}{Nature Human Behaviour}} \textbf{\bibinfo{volume}{7}}, \bibinfo{pages}{1808--1809} (\bibinfo{year}{2023}).

\bibitem{li2026random}
\bibinfo{author}{Li, R.}, \bibinfo{author}{Folk, D.}, \bibinfo{author}{Singh, A.}, \bibinfo{author}{Ungar, L.} \& \bibinfo{author}{Dunn, E.}
\newblock \bibinfo{title}{Is a random human peer better than a highly supportive chatbot in reducing loneliness over time?}
\newblock \emph{\bibinfo{journal}{Journal of Experimental Social Psychology}} \textbf{\bibinfo{volume}{125}}, \bibinfo{pages}{104911} (\bibinfo{year}{2026}).

\bibitem{kestin2025ai}
\bibinfo{author}{Kestin, G.}, \bibinfo{author}{Miller, K.}, \bibinfo{author}{Klales, A.}, \bibinfo{author}{Milbourne, T.} \& \bibinfo{author}{Ponti, G.}
\newblock \bibinfo{title}{Ai tutoring outperforms in-class active learning: an rct introducing a novel research-based design in an authentic educational setting}.
\newblock \emph{\bibinfo{journal}{Scientific Reports}} \textbf{\bibinfo{volume}{15}}, \bibinfo{pages}{17458} (\bibinfo{year}{2025}).

\bibitem{khasentino2025personal}
\bibinfo{author}{Khasentino, J.} \emph{et~al.}
\newblock \bibinfo{title}{A personal health large language model for sleep and fitness coaching}.
\newblock \emph{\bibinfo{journal}{Nature Medicine}} \textbf{\bibinfo{volume}{31}}, \bibinfo{pages}{3394--3403} (\bibinfo{year}{2025}).

\bibitem{bruce2019loneliness}
\bibinfo{author}{Bruce, L.~D.}, \bibinfo{author}{Wu, J.~S.}, \bibinfo{author}{Lustig, S.~L.}, \bibinfo{author}{Russell, D.~W.} \& \bibinfo{author}{Nemecek, D.~A.}
\newblock \bibinfo{title}{Loneliness in the united states: A 2018 national panel survey of demographic, structural, cognitive, and behavioral characteristics}.
\newblock \emph{\bibinfo{journal}{American Journal of Health Promotion}} \textbf{\bibinfo{volume}{33}}, \bibinfo{pages}{1123--1133} (\bibinfo{year}{2019}).

\bibitem{surkalim2022prevalence}
\bibinfo{author}{Surkalim, D.~L.} \emph{et~al.}
\newblock \bibinfo{title}{The prevalence of loneliness across 113 countries: systematic review and meta-analysis}.
\newblock \emph{\bibinfo{journal}{bmj}} \textbf{\bibinfo{volume}{376}} (\bibinfo{year}{2022}).

\bibitem{pei2025bridging}
\bibinfo{author}{Pei, R.} \emph{et~al.}
\newblock \bibinfo{title}{Bridging the empathy perception gap fosters social connection}.
\newblock \emph{\bibinfo{journal}{Nature Human Behaviour}} \bibinfo{pages}{1--14} (\bibinfo{year}{2025}).

\bibitem{lloyd2017listening}
\bibinfo{author}{Lloyd, K.~J.}, \bibinfo{author}{Boer, D.} \& \bibinfo{author}{Voelpel, S.~C.}
\newblock \bibinfo{title}{From listening to leading: Toward an understanding of supervisor listening within the framework of leader-member exchange theory}.
\newblock \emph{\bibinfo{journal}{International Journal of Business Communication}} \textbf{\bibinfo{volume}{54}}, \bibinfo{pages}{431--451} (\bibinfo{year}{2017}).

\bibitem{li2024ethical}
\bibinfo{author}{Li, Q.}
\newblock \bibinfo{title}{Ethical leadership, internal job satisfaction and ocb: the moderating role of leader empathy in emerging industries}.
\newblock \emph{\bibinfo{journal}{Humanities and Social Sciences Communications}} \textbf{\bibinfo{volume}{11}}, \bibinfo{pages}{1--9} (\bibinfo{year}{2024}).

\bibitem{yang2022effects}
\bibinfo{author}{Yang, L.} \emph{et~al.}
\newblock \bibinfo{title}{The effects of remote work on collaboration among information workers}.
\newblock \emph{\bibinfo{journal}{Nature human behaviour}} \textbf{\bibinfo{volume}{6}}, \bibinfo{pages}{43--54} (\bibinfo{year}{2022}).

\bibitem{emanuel2026home}
\bibinfo{author}{Emanuel, N.}, \bibinfo{author}{Harrington, E.} \& \bibinfo{author}{Pallais, A.}
\newblock \bibinfo{title}{Home alone: Remote work, isolation, and mental health}.
\newblock \emph{\bibinfo{journal}{Science}} \textbf{\bibinfo{volume}{392}}, \bibinfo{pages}{eaec7671} (\bibinfo{year}{2026}).

\bibitem{machia2024social}
\bibinfo{author}{Machia, L.~V.}, \bibinfo{author}{Corral, D.} \& \bibinfo{author}{Jakubiak, B.~K.}
\newblock \bibinfo{title}{Social need fulfillment model for human--ai relationships}  (\bibinfo{year}{2024}).

\bibitem{zimmerman2024human}
\bibinfo{author}{Zimmerman, A.}, \bibinfo{author}{Janhonen, J.} \& \bibinfo{author}{Beer, E.}
\newblock \bibinfo{title}{Human/ai relationships: challenges, downsides, and impacts on human/human relationships}.
\newblock \emph{\bibinfo{journal}{AI and Ethics}} \textbf{\bibinfo{volume}{4}}, \bibinfo{pages}{1555--1567} (\bibinfo{year}{2024}).

\bibitem{wenger2026people}
\bibinfo{author}{Wenger, J.~D.}, \bibinfo{author}{Cameron, C.~D.} \& \bibinfo{author}{Inzlicht, M.}
\newblock \bibinfo{title}{People choose to receive human empathy despite rating ai empathy higher}.
\newblock \emph{\bibinfo{journal}{Communications Psychology}}  (\bibinfo{year}{2026}).

\bibitem{phang2025investigating}
\bibinfo{author}{Phang, J.} \emph{et~al.}
\newblock \bibinfo{title}{Investigating affective use and emotional well-being on chatgpt}.
\newblock \emph{\bibinfo{journal}{arXiv preprint arXiv:2504.03888}}  (\bibinfo{year}{2025}).

\bibitem{depow2021experience}
\bibinfo{author}{Depow, G.~J.}, \bibinfo{author}{Francis, Z.} \& \bibinfo{author}{Inzlicht, M.}
\newblock \bibinfo{title}{The experience of empathy in everyday life}.
\newblock \emph{\bibinfo{journal}{Psychological Science}} \textbf{\bibinfo{volume}{32}}, \bibinfo{pages}{1198--1213} (\bibinfo{year}{2021}).

\bibitem{moore2025expressing}
\bibinfo{author}{Moore, J.} \emph{et~al.}
\newblock \bibinfo{title}{Expressing stigma and inappropriate responses prevents llms from safely replacing mental health providers.}
\newblock \emph{\bibinfo{journal}{Proceedings of the 2025 ACM Conference on Fairness, Accountability, and Transparency}} \bibinfo{pages}{599--627} (\bibinfo{year}{2025}).

\bibitem{reddan2024film}
\bibinfo{author}{Reddan, M.~C.}, \bibinfo{author}{Garcia, S.~B.}, \bibinfo{author}{Golarai, G.}, \bibinfo{author}{Eberhardt, J.~L.} \& \bibinfo{author}{Zaki, J.}
\newblock \bibinfo{title}{Film intervention increases empathic understanding of formerly incarcerated people and support for criminal justice reform}.
\newblock \emph{\bibinfo{journal}{Proceedings of the National Academy of Sciences}} \textbf{\bibinfo{volume}{121}}, \bibinfo{pages}{e2322819121} (\bibinfo{year}{2024}).

\bibitem{eyal2018perspective}
\bibinfo{author}{Eyal, T.}, \bibinfo{author}{Steffel, M.~L.} \& \bibinfo{author}{Epley, N.}
\newblock \bibinfo{title}{Perspective mistaking: Accurately understanding the mind of another requires getting perspective, not taking perspective}.
\newblock \emph{\bibinfo{journal}{Journal of Personality and Social Psychology}} \textbf{\bibinfo{volume}{114}}, \bibinfo{pages}{547--571} (\bibinfo{year}{2018}).

\bibitem{moyers2014motivational}
\bibinfo{author}{Moyers, T.} \emph{et~al.}
\newblock \bibinfo{title}{Motivational interviewing treatment integrity coding manual 4.1 (miti 4.1)}.
\newblock \emph{\bibinfo{journal}{Unpublished manual}}  (\bibinfo{year}{2014}).

\bibitem{rodriguez2019measuring}
\bibinfo{author}{Rodriguez, A.~M.} \& \bibinfo{author}{Lown, B.~A.}
\newblock \bibinfo{title}{Measuring compassionate healthcare with the 12-item schwartz center compassionate care scale}.
\newblock \emph{\bibinfo{journal}{PloS one}} \textbf{\bibinfo{volume}{14}}, \bibinfo{pages}{e0220911} (\bibinfo{year}{2019}).

\bibitem{rizvi2016dialectical}
\bibinfo{author}{Rizvi, S.} \& \bibinfo{author}{Thomas, M.}
\newblock \bibinfo{title}{Dialectical behavior therapy} \bibinfo{pages}{125--130} (\bibinfo{year}{2016}).

\bibitem{rogers1961becoming}
\bibinfo{author}{Rogers, C.~R.}
\newblock \emph{\bibinfo{title}{On Becoming a Person: A Therapist’s View of Psychotherapy}}  (\bibinfo{publisher}{Houghton Mifflin}, \bibinfo{address}{Boston}, \bibinfo{year}{1961}).

\bibitem{bodie2011active}
\bibinfo{author}{Bodie, G.~D.}
\newblock \bibinfo{title}{The active-empathic listening scale (aels): Conceptualization and evidence of validity within the interpersonal domain}.
\newblock \emph{\bibinfo{journal}{Communication Quarterly}} \textbf{\bibinfo{volume}{59}}, \bibinfo{pages}{277--295} (\bibinfo{year}{2011}).

\bibitem{mercer2004consultation}
\bibinfo{author}{Mercer, S.~W.}, \bibinfo{author}{Maxwell, M.}, \bibinfo{author}{Heaney, D.} \& \bibinfo{author}{Watt, G.~C.}
\newblock \bibinfo{title}{The consultation and relational empathy (care) measure: development and preliminary validation and reliability of an empathy-based consultation process measure}.
\newblock \emph{\bibinfo{journal}{Family practice}} \textbf{\bibinfo{volume}{21}}, \bibinfo{pages}{699--705} (\bibinfo{year}{2004}).

\bibitem{kim2018social}
\bibinfo{author}{Kim, H.~Y.} \emph{et~al.}
\newblock \bibinfo{title}{Social perspective-taking performance: Construct, measurement, and relations with academic performance and engagement}.
\newblock \emph{\bibinfo{journal}{Journal of Applied Developmental Psychology}} \textbf{\bibinfo{volume}{57}}, \bibinfo{pages}{24--41} (\bibinfo{year}{2018}).

\bibitem{Feeney_Collins_2018}
\bibinfo{author}{Vangelisti, A.~L.} \& \bibinfo{author}{Perlman, D.}
\newblock \bibinfo{title}{The cambridge handbook of personal relationships}.
\newblock \emph{\bibinfo{journal}{Cambridge University Press}}  (\bibinfo{year}{2018}).

\bibitem{goldsmith1997normative}
\bibinfo{author}{Goldsmith, D.~J.} \& \bibinfo{author}{Fitch, K.}
\newblock \bibinfo{title}{Normative context of advice as social support}.
\newblock \emph{\bibinfo{journal}{Human Communication Research}} \textbf{\bibinfo{volume}{23}}, \bibinfo{pages}{454--476} (\bibinfo{year}{1997}).
\newblock \urlprefix\url{https://academic.oup.com/hcr/article/23/4/454/4564959}.

\bibitem{weger2014relative}
\bibinfo{author}{Weger, H.}, \bibinfo{author}{Castle~Bell, G.}, \bibinfo{author}{Minei, E.~M.} \& \bibinfo{author}{Robinson, M.~C.}
\newblock \bibinfo{title}{The relative effectiveness of active listening in initial interactions}.
\newblock \emph{\bibinfo{journal}{International Journal of Listening}} \textbf{\bibinfo{volume}{28}}, \bibinfo{pages}{13--31} (\bibinfo{year}{2014}).
\newblock \urlprefix\url{https://doi.org/10.1080/10904018.2013.813234}.

\bibitem{jones2004personcentered}
\bibinfo{author}{Jones, S.~M.}
\newblock \bibinfo{title}{Putting the person into person-centered and immediate emotional support}.
\newblock \emph{\bibinfo{journal}{Communication Research}} \textbf{\bibinfo{volume}{31}}, \bibinfo{pages}{338--360} (\bibinfo{year}{2004}).

\bibitem{hacker2008relational}
\bibinfo{author}{Hacker, T.}
\newblock \emph{\bibinfo{title}{The relational compassion scale: development and validation of a new self rated scale for the assessment of self-other compassion}}.
\newblock Ph.D. thesis, \bibinfo{school}{University of Glasgow} (\bibinfo{year}{2008}).

\bibitem{paulus2022effectiveness}
\bibinfo{author}{Paulus, C.~M.} \& \bibinfo{author}{Meinken, S.}
\newblock \bibinfo{title}{The effectiveness of empathy training in health care: a meta-analysis of training content and methods}.
\newblock \emph{\bibinfo{journal}{International Journal of Medical Education}} \textbf{\bibinfo{volume}{13}}, \bibinfo{pages}{1} (\bibinfo{year}{2022}).

\bibitem{king2013best}
\bibinfo{author}{King, A.} \& \bibinfo{author}{Hoppe, R.~B.}
\newblock \bibinfo{title}{“best practice” for patient-centered communication: a narrative review}.
\newblock \emph{\bibinfo{journal}{Journal of graduate medical education}} \textbf{\bibinfo{volume}{5}}, \bibinfo{pages}{385--393} (\bibinfo{year}{2013}).

\bibitem{kahriman2016effect}
\bibinfo{author}{Kahriman, I.} \emph{et~al.}
\newblock \bibinfo{title}{The effect of empathy training on the empathic skills of nurses}.
\newblock \emph{\bibinfo{journal}{Iranian Red Crescent Medical Journal}} \textbf{\bibinfo{volume}{18}}, \bibinfo{pages}{e24847} (\bibinfo{year}{2016}).

\bibitem{okonofua2022scalable}
\bibinfo{author}{Okonofua, J.~A.}, \bibinfo{author}{Goyer, J.~P.}, \bibinfo{author}{Lindsay, C.~A.}, \bibinfo{author}{Haugabrook, J.} \& \bibinfo{author}{Walton, G.~M.}
\newblock \bibinfo{title}{A scalable empathic-mindset intervention reduces group disparities in school suspensions}.
\newblock \emph{\bibinfo{journal}{Science advances}} \textbf{\bibinfo{volume}{8}}, \bibinfo{pages}{eabj0691} (\bibinfo{year}{2022}).

\bibitem{lee2024large}
\bibinfo{author}{Lee, Y.~K.}, \bibinfo{author}{Suh, J.}, \bibinfo{author}{Zhan, H.}, \bibinfo{author}{Li, J.~J.} \& \bibinfo{author}{Ong, D.~C.}
\newblock \bibinfo{title}{Large language models produce responses perceived to be empathic}.
\newblock \emph{\bibinfo{journal}{2024 12th International Conference on Affective Computing and Intelligent Interaction (ACII)}} \bibinfo{pages}{63--71} (\bibinfo{year}{2024}).

\bibitem{yang2024social}
\bibinfo{author}{Yang, D.} \emph{et~al.}
\newblock \bibinfo{title}{Social skill training with large language models}.
\newblock \emph{\bibinfo{journal}{arXiv preprint arXiv:2404.04204}}  (\bibinfo{year}{2024}).

\bibitem{chun2025conflictlens}
\bibinfo{author}{Chun, J.}, \bibinfo{author}{Zhang, G.} \& \bibinfo{author}{Xia, M.}
\newblock \bibinfo{title}{Conflictlens: Llm-based conflict resolution training in romantic relationship}.
\newblock \emph{\bibinfo{journal}{Adjunct Proceedings of the 38th Annual ACM Symposium on User Interface Software and Technology}} \bibinfo{pages}{1--3} (\bibinfo{year}{2025}).

\bibitem{li2024exploring}
\bibinfo{author}{Li, Z.}, \bibinfo{author}{Babar, P.~P.}, \bibinfo{author}{Barry, M.} \& \bibinfo{author}{Peiris, R.~L.}
\newblock \bibinfo{title}{Exploring the use of large language model-driven chatbots in virtual reality to train autistic individuals in job communication skills}.
\newblock \emph{\bibinfo{journal}{Extended Abstracts of the CHI Conference on Human Factors in Computing Systems}} \bibinfo{pages}{1--7} (\bibinfo{year}{2024}).

\bibitem{dinnar2025negotiation}
\bibinfo{author}{Dinnar, S.}, \bibinfo{author}{Susskind, L.}, \bibinfo{author}{Sibanda, L.} \& \bibinfo{author}{Olaleye, O.}
\newblock \bibinfo{title}{Negotiation backtable bots: Using genai to improve multiparty negotiation instruction}.
\newblock \emph{\bibinfo{journal}{Negotiation Journal}} \textbf{\bibinfo{volume}{41}}, \bibinfo{pages}{19--65} (\bibinfo{year}{2025}).

\bibitem{duddu2025does}
\bibinfo{author}{Duddu, V.} \emph{et~al.}
\newblock \bibinfo{title}{Does ai coaching prepare us for workplace negotiations?}
\newblock \emph{\bibinfo{journal}{arXiv preprint arXiv:2509.22545}}  (\bibinfo{year}{2025}).

\bibitem{tessler2024ai}
\bibinfo{author}{Tessler, M.~H.} \emph{et~al.}
\newblock \bibinfo{title}{Ai can help humans find common ground in democratic deliberation}.
\newblock \emph{\bibinfo{journal}{Science}} \textbf{\bibinfo{volume}{386}}, \bibinfo{pages}{eadq2852} (\bibinfo{year}{2024}).

\bibitem{louie2026can}
\bibinfo{author}{Louie, R.} \emph{et~al.}
\newblock \bibinfo{title}{Can llm-simulated practice and feedback upskill human counselors? a randomized study with 90+ novice counselors}.
\newblock \emph{\bibinfo{journal}{Proceedings of the 2026 CHI Conference on Human Factors in Computing Systems}} \bibinfo{pages}{1--31} (\bibinfo{year}{2026}).

\bibitem{jordan2016empathy}
\bibinfo{author}{Jordan, M.~R.}, \bibinfo{author}{Amir, D.} \& \bibinfo{author}{Bloom, P.}
\newblock \bibinfo{title}{Are empathy and concern psychologically distinct?}
\newblock \emph{\bibinfo{journal}{Emotion}} \textbf{\bibinfo{volume}{16}}, \bibinfo{pages}{1107} (\bibinfo{year}{2016}).

\bibitem{konrath2018development}
\bibinfo{author}{Konrath, S.}, \bibinfo{author}{Meier, B.~P.} \& \bibinfo{author}{Bushman, B.~J.}
\newblock \bibinfo{title}{Development and validation of the single item trait empathy scale (sites)}.
\newblock \emph{\bibinfo{journal}{Journal of research in personality}} \textbf{\bibinfo{volume}{73}}, \bibinfo{pages}{111--122} (\bibinfo{year}{2018}).

\bibitem{gerdes2011teaching}
\bibinfo{author}{Gerdes, K.~E.}, \bibinfo{author}{Segal, E.~A.}, \bibinfo{author}{Jackson, K.~F.} \& \bibinfo{author}{Mullins, J.~L.}
\newblock \bibinfo{title}{Teaching empathy: A framework rooted in social cognitive neuroscience and social justice}.
\newblock \emph{\bibinfo{journal}{Journal of social work education}} \textbf{\bibinfo{volume}{47}}, \bibinfo{pages}{109--131} (\bibinfo{year}{2011}).

\bibitem{fitzsimons2004reactance}
\bibinfo{author}{Fitzsimons, G.~J.} \& \bibinfo{author}{Lehmann, D.~R.}
\newblock \bibinfo{title}{Reactance to recommendations: When unsolicited advice yields contrary responses}.
\newblock \emph{\bibinfo{journal}{Marketing Science}} \textbf{\bibinfo{volume}{23}}, \bibinfo{pages}{82--94} (\bibinfo{year}{2004}).

\bibitem{burleson2008counts}
\bibinfo{author}{Burleson, B.~R.}
\newblock \bibinfo{title}{What counts as effective emotional support}.
\newblock \emph{\bibinfo{journal}{Studies in applied interpersonal communication}} \bibinfo{pages}{207--227} (\bibinfo{year}{2008}).

\bibitem{yao2023person}
\bibinfo{author}{Yao, L.} \& \bibinfo{author}{Kabir, R.}
\newblock \bibinfo{title}{Person-centered therapy (rogerian therapy)}  (\bibinfo{year}{2023}).

\bibitem{becker1975phrasal}
\bibinfo{author}{Becker, J.~D.}
\newblock \bibinfo{title}{The phrasal lexicon}.
\newblock \emph{\bibinfo{journal}{Theoretical issues in natural language processing}}  (\bibinfo{year}{1975}).

\bibitem{o1995managing}
\bibinfo{author}{O’Keefe, B.~J.} \& \bibinfo{author}{Lambert, B.~L.}
\newblock \bibinfo{title}{Managing the flow of ideas: A local management approach to message design}.
\newblock \emph{\bibinfo{journal}{Annals of the International Communication Association}} \textbf{\bibinfo{volume}{18}}, \bibinfo{pages}{54--82} (\bibinfo{year}{1995}).

\bibitem{lambert2001semi}
\bibinfo{author}{Lambert, B.~L.}
\newblock \bibinfo{title}{Semi-automated content analysis of pharmacist-patient interactions using the theme machine document-clustering system}.
\newblock \emph{\bibinfo{journal}{Progress in communication sciences}} \bibinfo{pages}{103--122} (\bibinfo{year}{2001}).

\bibitem{peng2025use}
\bibinfo{author}{Peng, K.}, \bibinfo{author}{Movva, R.}, \bibinfo{author}{Kleinberg, J.}, \bibinfo{author}{Pierson, E.} \& \bibinfo{author}{Garg, N.}
\newblock \bibinfo{title}{Use sparse autoencoders to discover unknown concepts, not to act on known concepts}.
\newblock \emph{\bibinfo{journal}{arXiv preprint arXiv:2506.23845}}  (\bibinfo{year}{2025}).

\bibitem{singhdiscovering}
\bibinfo{author}{Singh, N.}, \bibinfo{author}{Cherep, M.} \& \bibinfo{author}{Maes, P.}
\newblock \bibinfo{title}{Discovering and steering interpretable concepts in large generative music models}.
\newblock \emph{\bibinfo{journal}{AI for Music Workshop}} .

\bibitem{zaki2012neuroscience}
\bibinfo{author}{Zaki, J.} \& \bibinfo{author}{Ochsner, K.~N.}
\newblock \bibinfo{title}{The neuroscience of empathy: progress, pitfalls and promise}.
\newblock \emph{\bibinfo{journal}{Nature neuroscience}} \textbf{\bibinfo{volume}{15}}, \bibinfo{pages}{675--680} (\bibinfo{year}{2012}).

\bibitem{gueorguieva2026ai}
\bibinfo{author}{Gueorguieva, E.} \emph{et~al.}
\newblock \bibinfo{title}{Ai generates well-liked but templatic empathic responses}.
\newblock \emph{\bibinfo{journal}{arXiv preprint arXiv:2604.08479}}  (\bibinfo{year}{2026}).

\bibitem{jabarian2025artificial}
\bibinfo{author}{Jabarian, B.} \& \bibinfo{author}{Imas, A.}
\newblock \bibinfo{title}{Artificial writing and automated detection}.
\newblock \bibinfo{type}{Working Paper} \bibinfo{number}{34223}, \bibinfo{institution}{National Bureau of Economic Research}, \bibinfo{address}{Cambridge, MA} (\bibinfo{year}{2025}).

\bibitem{depow2025individual}
\bibinfo{author}{Depow, G.~J.} \& \bibinfo{author}{Inzlicht, M.}
\newblock \bibinfo{title}{How individual differences in empathy predict moments of empathy in everyday life}.
\newblock \emph{\bibinfo{journal}{Personality and Social Psychology Bulletin}} \bibinfo{pages}{01461672251333823} (\bibinfo{year}{2025}).

\bibitem{yakura2024empirical}
\bibinfo{author}{Yakura, H.} \emph{et~al.}
\newblock \bibinfo{title}{Empirical evidence of large language model's influence on human spoken communication}.
\newblock \emph{\bibinfo{journal}{arXiv preprint arXiv:2409.01754}}  (\bibinfo{year}{2024}).

\bibitem{kobak2025delving}
\bibinfo{author}{Kobak, D.}, \bibinfo{author}{Gonz{\'a}lez-M{\'a}rquez, R.}, \bibinfo{author}{Horv{\'a}t, E.-{\'A}.} \& \bibinfo{author}{Lause, J.}
\newblock \bibinfo{title}{Delving into llm-assisted writing in biomedical publications through excess vocabulary}.
\newblock \emph{\bibinfo{journal}{Science Advances}} \textbf{\bibinfo{volume}{11}}, \bibinfo{pages}{eadt3813} (\bibinfo{year}{2025}).

\bibitem{padmakumar2024does}
\bibinfo{author}{Padmakumar, V.} \& \bibinfo{author}{He, H.}
\newblock \bibinfo{title}{Does writing with language models reduce content diversity?}
\newblock \emph{\bibinfo{journal}{International Conference on Learning Representations}} \textbf{\bibinfo{volume}{2024}}, \bibinfo{pages}{642--669} (\bibinfo{year}{2024}).

\bibitem{doshi2024generative}
\bibinfo{author}{Doshi, A.~R.} \& \bibinfo{author}{Hauser, O.~P.}
\newblock \bibinfo{title}{Generative ai enhances individual creativity but reduces the collective diversity of novel content}.
\newblock \emph{\bibinfo{journal}{Science advances}} \textbf{\bibinfo{volume}{10}}, \bibinfo{pages}{eadn5290} (\bibinfo{year}{2024}).

\bibitem{anderson2024homogenization}
\bibinfo{author}{Anderson, B.~R.}, \bibinfo{author}{Shah, J.~H.} \& \bibinfo{author}{Kreminski, M.}
\newblock \bibinfo{title}{Homogenization effects of large language models on human creative ideation}.
\newblock \emph{\bibinfo{journal}{Proceedings of the 16th conference on creativity \& cognition}} \bibinfo{pages}{413--425} (\bibinfo{year}{2024}).

\bibitem{crockett2025empathy}
\bibinfo{author}{Crockett, M.}
\newblock \bibinfo{title}{Empathy, thick and thin}.
\newblock \emph{\bibinfo{journal}{Available at SSRN 5862422}}  (\bibinfo{year}{2025}).

\bibitem{mei2025reducing}
\bibinfo{author}{Mei, S.}, \bibinfo{author}{Deng, Y.}, \bibinfo{author}{Zheng, G.} \& \bibinfo{author}{Han, S.}
\newblock \bibinfo{title}{Reducing racial ingroup biases in empathy and altruistic decision-making by shifting racial identification}.
\newblock \emph{\bibinfo{journal}{Science Advances}} \textbf{\bibinfo{volume}{11}}, \bibinfo{pages}{eadt6207} (\bibinfo{year}{2025}).

\bibitem{mastroianni2021conversations}
\bibinfo{author}{Mastroianni, A.~M.}, \bibinfo{author}{Gilbert, D.~T.}, \bibinfo{author}{Cooney, G.} \& \bibinfo{author}{Wilson, T.~D.}
\newblock \bibinfo{title}{Do conversations end when people want them to?}
\newblock \emph{\bibinfo{journal}{Proceedings of the National Academy of Sciences}} \textbf{\bibinfo{volume}{118}}, \bibinfo{pages}{e2011809118} (\bibinfo{year}{2021}).

\bibitem{rousseeuw1987silhouettes}
\bibinfo{author}{Rousseeuw, P.~J.}
\newblock \bibinfo{title}{Silhouettes: a graphical aid to the interpretation and validation of cluster analysis}.
\newblock \emph{\bibinfo{journal}{Journal of computational and applied mathematics}} \textbf{\bibinfo{volume}{20}}, \bibinfo{pages}{53--65} (\bibinfo{year}{1987}).

\bibitem{jacobson1992clinical}
\bibinfo{author}{Jacobson, N.~S.} \& \bibinfo{author}{Truax, P.}
\newblock \bibinfo{title}{Clinical significance: a statistical approach to defining meaningful change in psychotherapy research.}  (\bibinfo{year}{1992}).

\bibitem{feuerriegelreporting}
\bibinfo{author}{Feuerriegel, S.} \emph{et~al.}
\newblock \bibinfo{title}{A reporting checklist for large language models in behavioural science}.
\newblock \emph{\bibinfo{journal}{Nature human behaviour}}  (\bibinfo{year}{2026}).

\end{thebibliography}

\clearpage
\appendix



\tableofcontents

\newpage

\section{Supplementary Figures}

\begin{figure}[htbp]
    \centering
    \begin{subfigure}{\linewidth}
        \subcaption{}
        \centering
        \includegraphics[width=\textwidth]{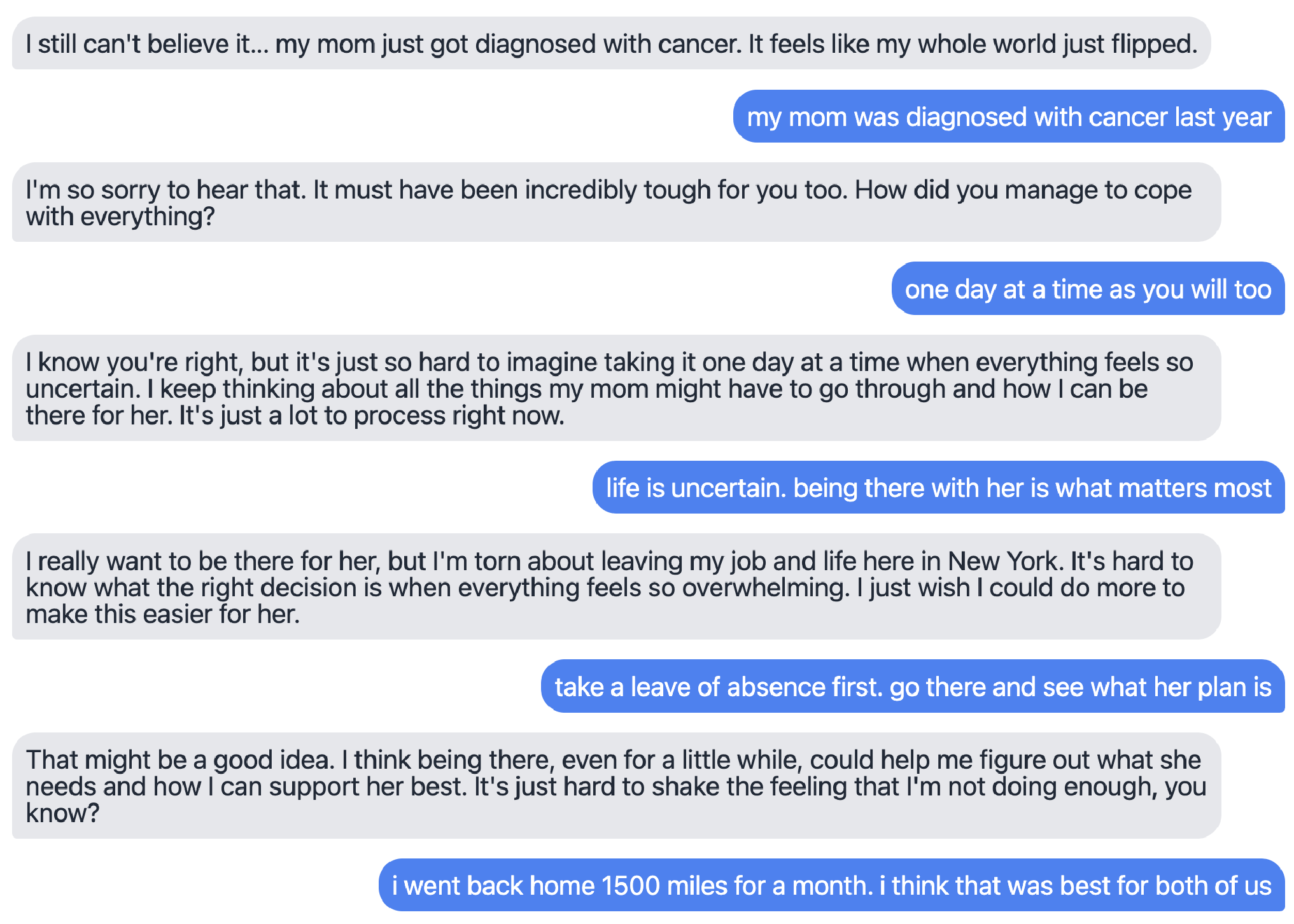}
    \end{subfigure}
    \hfill
    \begin{subfigure}{\linewidth}
        \subcaption{}
        \centering
        \includegraphics[width=\textwidth]{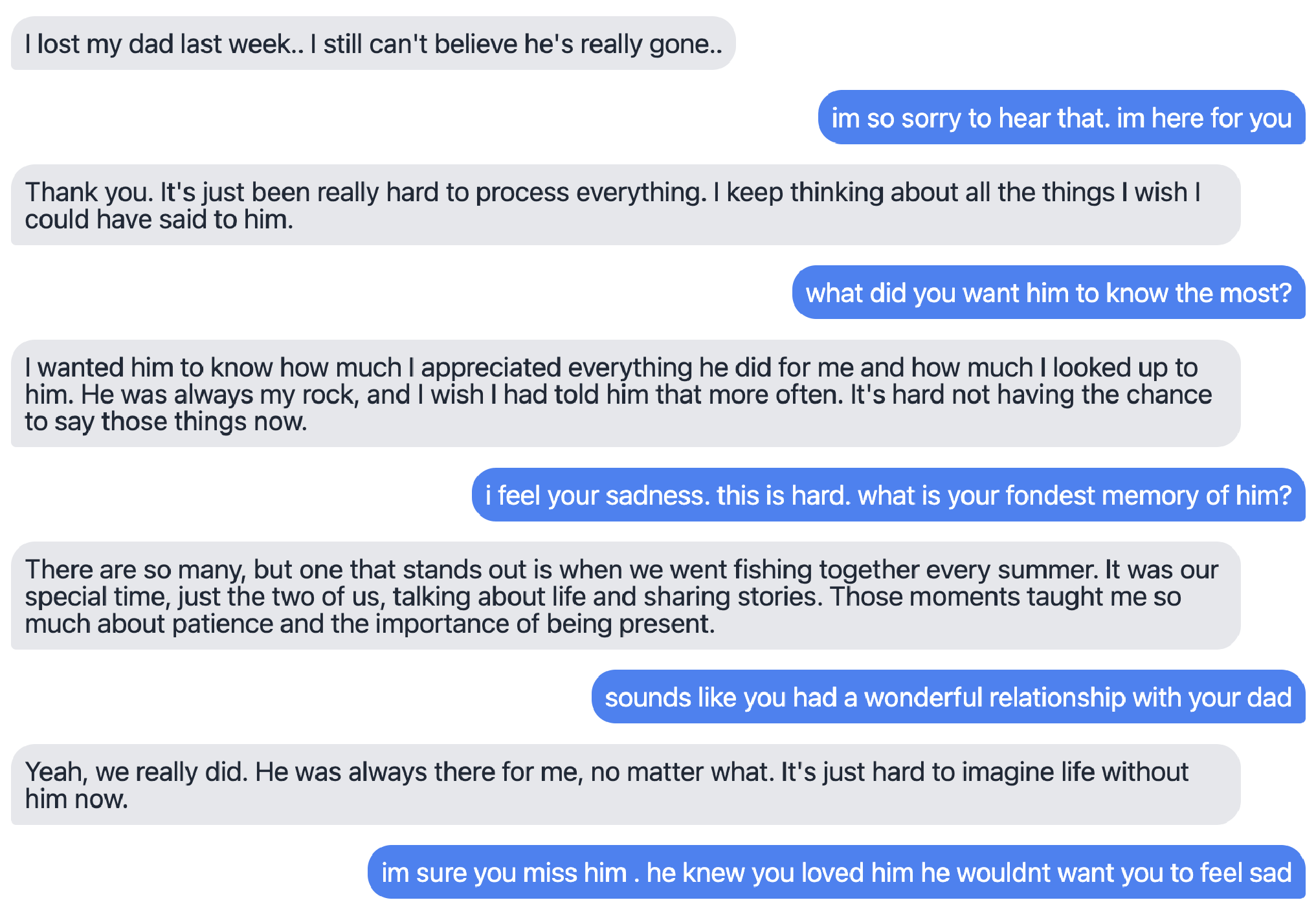}
    \end{subfigure}
        \begin{subfigure}{\linewidth}
        \subcaption{}
        \centering
        \includegraphics[width=\textwidth]{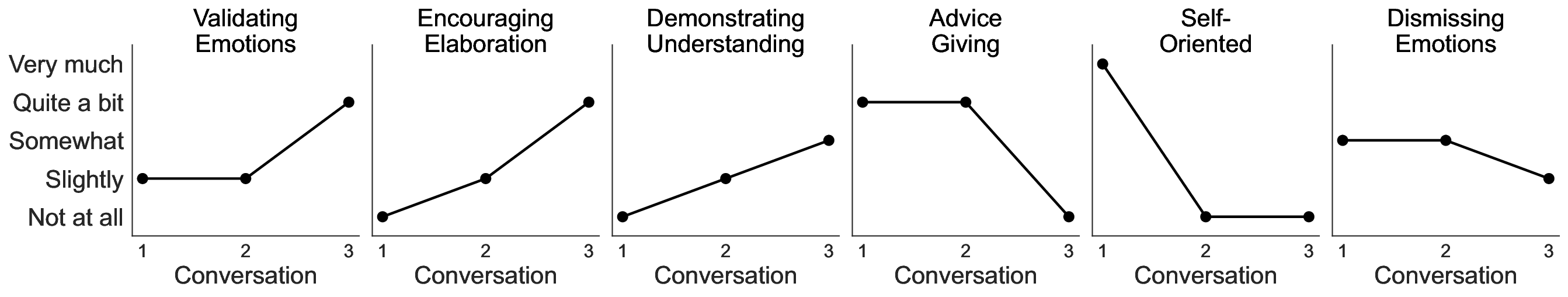}
    \end{subfigure}
    \caption*{\textbf{Extended Data Figure 1: Example conversational trajectory of a participant from the AI coach condition.} A. Conversation 1 (pre-training). B. Conversation 3 (post-training). C. Trajectory of empathic communication dimension ratings across three conversations for a top-improving participant.}
    \label{fig:example_convo}
\end{figure}

\begin{figure}
    \centering
        \includegraphics[width=.9\linewidth]{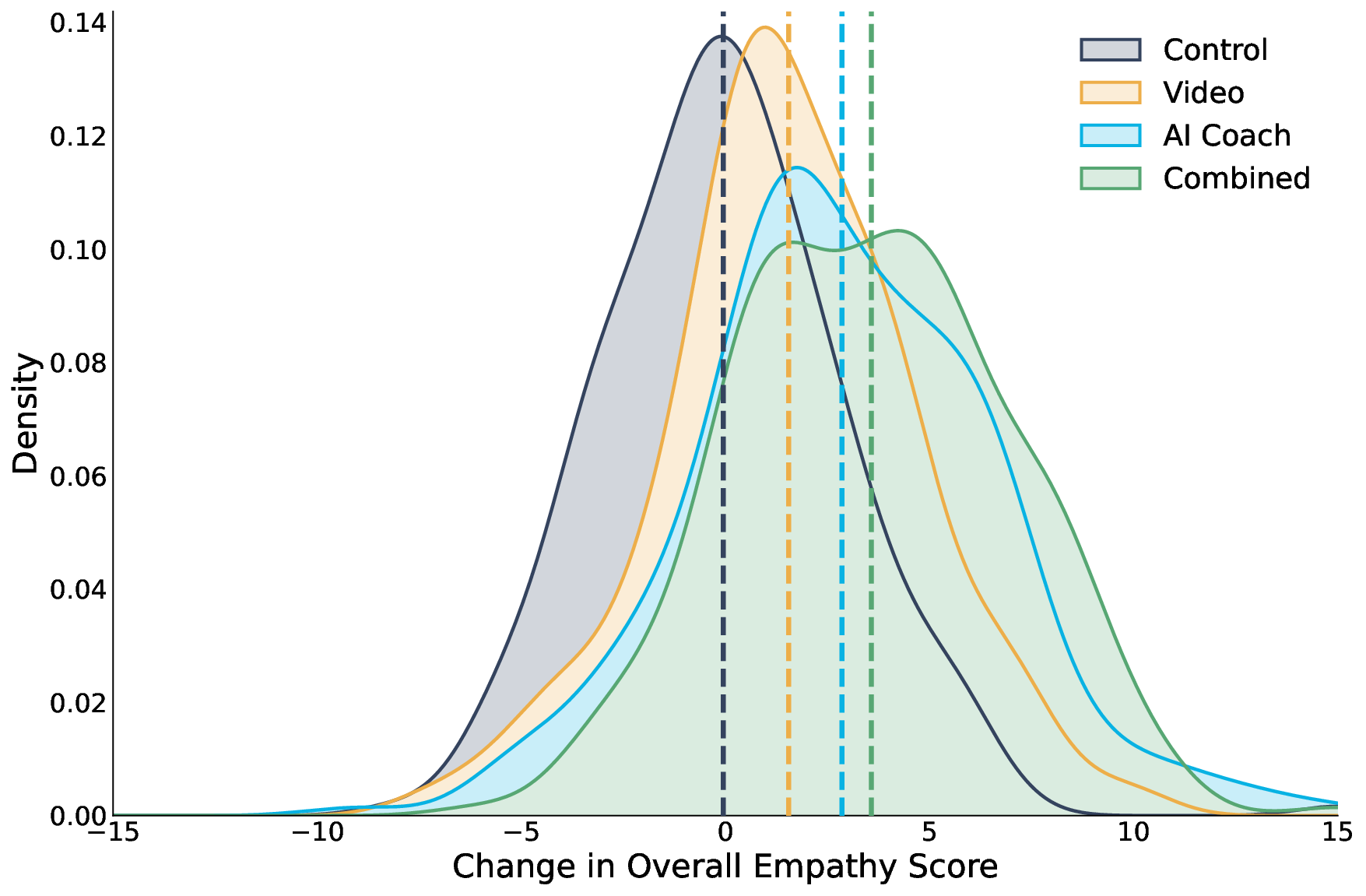}    
        \caption*{\textbf{Extended Data Figure 2: Distribution of change in overall empathy score across conditions.} Dashed lines mark condition-specific mean change from baseline (conversation 1) to post-intervention (conversations 2 and 3).}
    \label{fig:ed1}
\end{figure}
\clearpage

\begin{figure}
    \centering
        \includegraphics[width=.9\linewidth]{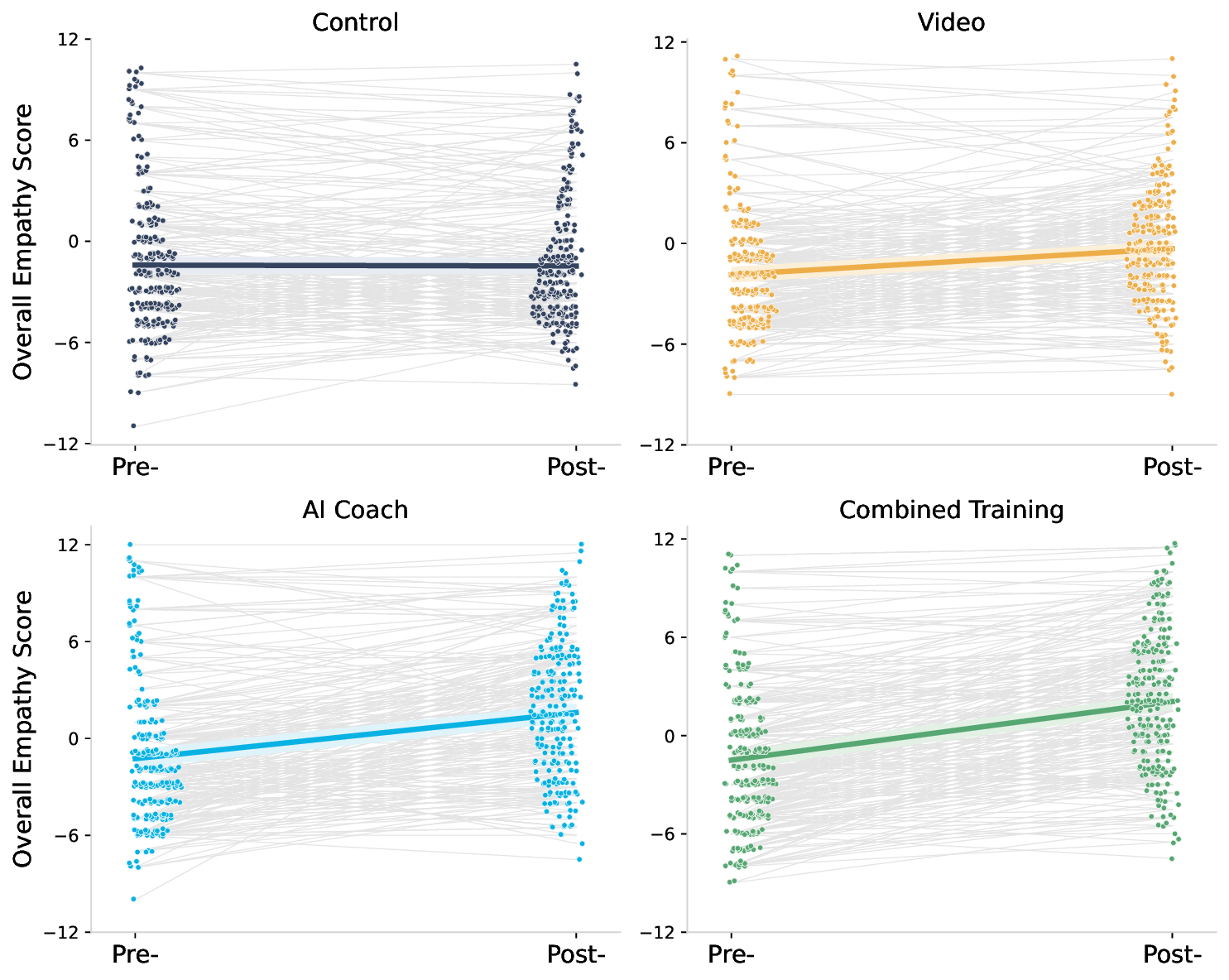}    
        \caption*{\textbf{Extended Data Figure 3: Individual participant trajectories across conditions.} Light gray lines show individual participants; thick colored lines and error bars show mean changes with 95\% confidence intervals from baseline (pre-) to post-intervention (post-).}
    \label{fig:ex-pre-post}
\end{figure}

\clearpage

\begin{figure}[ht]
    \centering
    \includegraphics[width=\linewidth]{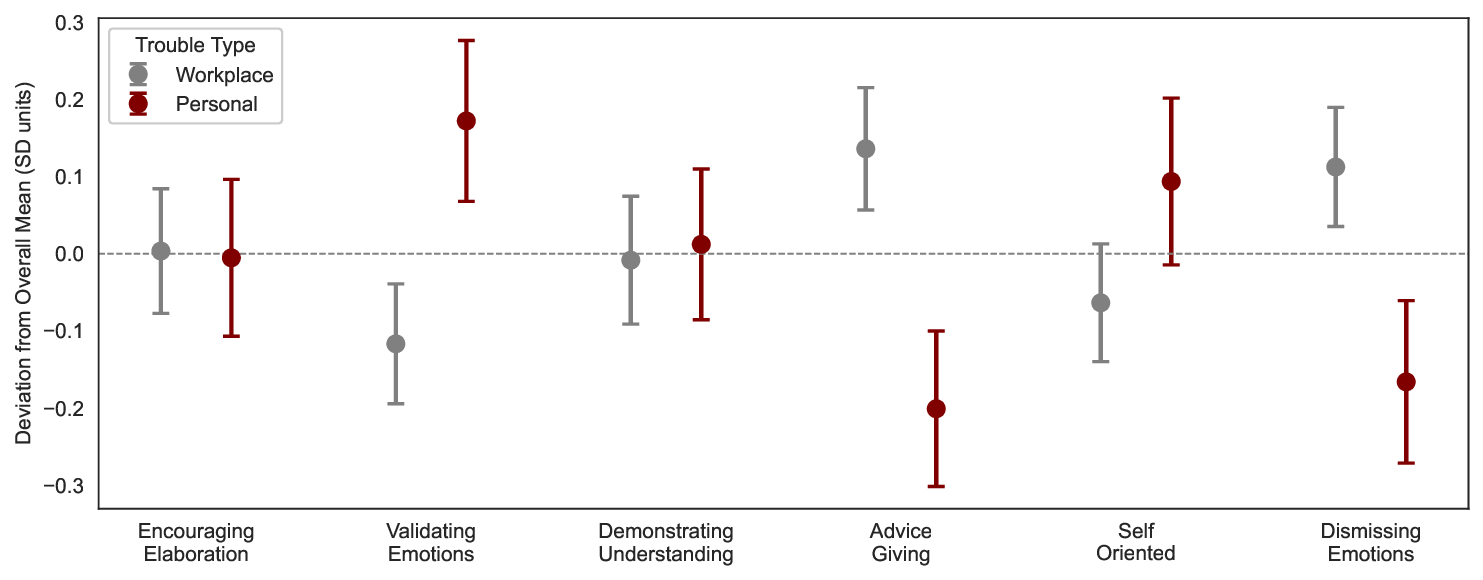}
    \caption*{\textbf{Extended Data Figure 4: Baseline differences between workplace and personal trouble conversations.} Mean standardized scores for six communication sub-components compare participants' responses to workplace troubles versus personal troubles during their first conversation. Error bars show 95\% confidence intervals.}
    \label{fig:ed2}
\end{figure}

\begin{figure}
    \centering
    \includegraphics[width=\linewidth]{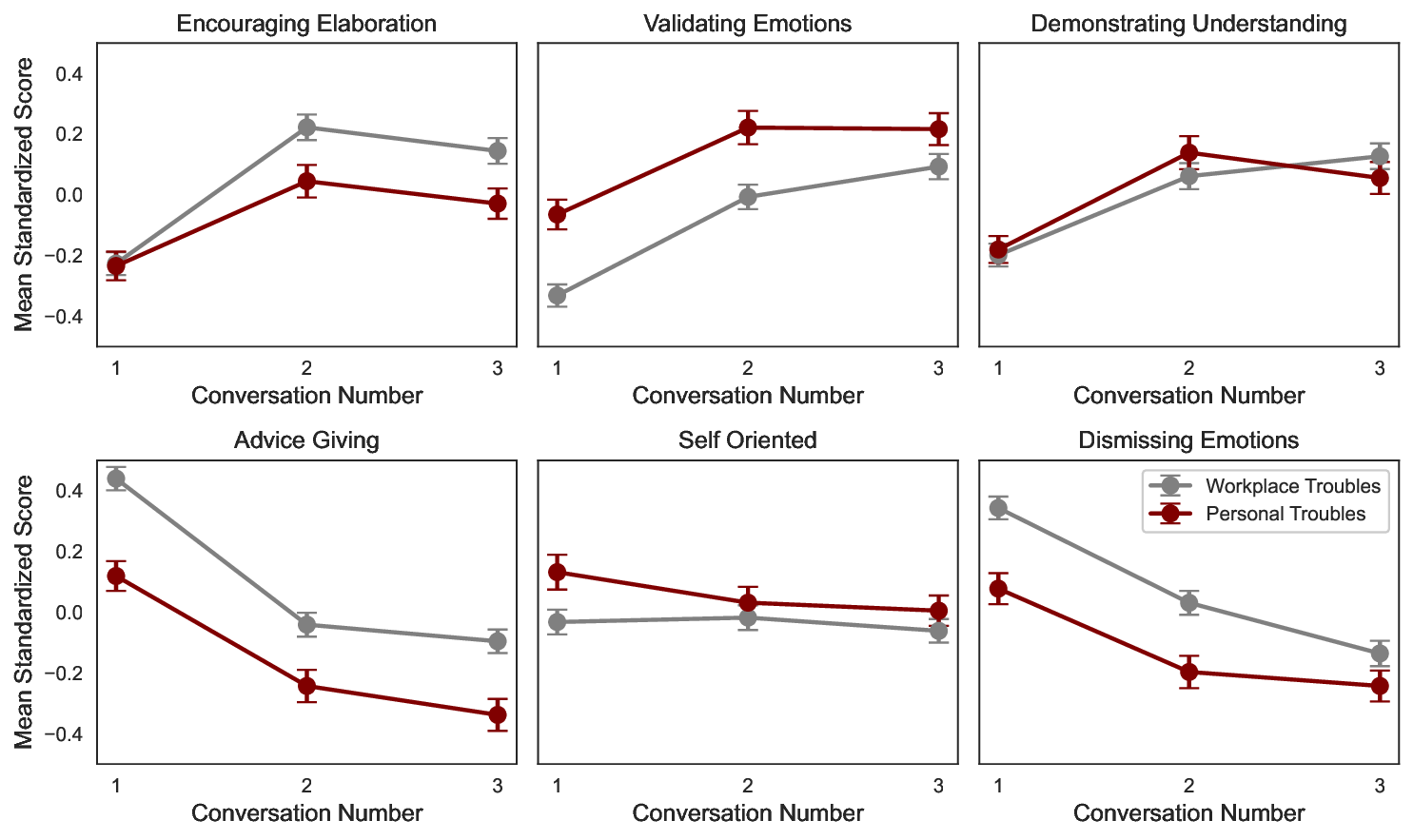}
    \caption*{\textbf{Extended Data Figure 5: Trajectories of individual communication behaviors across conversations by trouble type.} Each panel shows mean standardized scores (± SE) for a dimension of empathic communication across three consecutive conversations in the AI coach and combined training conditions, separated by scenario type (workplace troubles: job loss, passed up for promotion, feeling undervalued at work; personal troubles: parent's cancer diagnosis, loss of a parent).}
    \label{fig:ed3}
\end{figure}

\clearpage

\section{Baseline Survey Questions}
Participants responded on a 5-point scale: Not at all, Slightly, Somewhat, Quite a bit, Very much.

\paragraph{Jordan Empathy \cite{jordan2016empathy}} 
\begin{enumerate}
    \item If I see someone who is excited, I will feel excited myself.
    \item I sometimes find myself feeling the emotions of the people around me, even if I don't try to feel what they're feeling.
    \item If I'm watching a movie and a character injures their leg, I will feel pain in my leg.
    \item If I hear a story in which someone is scared, I will imagine how scared I would be in that situation and begin to feel scared myself.
    \item If I hear an awkward story about someone else, I might feel a little embarrassed.
    \item I can't watch shows in which an animal is being hunted by another because I feel nervous as if I am being hunted.
    \item If I see someone fidgeting, I'll start feeling anxious too.   
\end{enumerate}

\paragraph{SITES \cite{konrath2018development}}
\begin{enumerate}
    \item I am an empathetic person.
\end{enumerate}

\clearpage

\section{Role Playing Scenario Prompts}

\begin{tcolorbox}[
  enhanced jigsaw,
  breakable,
  colback=controlcolor!70!white, 
  opacityback=0.7, 
  colframe=controlcolor!80!black,
  title={\centering Role Playing Scenarios},
  coltitle=black,
  fonttitle=\bfseries,
  titlerule=0mm,
  title style={top color=controlcolor!30!white, bottom color=controlcolor!10!white},
  arc=3mm,
  boxrule=0.8pt,
  width=\textwidth,
  left=5pt,
  right=5pt,
  top=8pt,
  bottom=8pt,
  before skip=10pt,
  after skip=10pt
]
\vspace{1em}

\paragraph*{Job loss}
\textbf{Conversation starter:} ``So, I just lost my job today. I had a sense this was coming, but it's still a shock.."\\

\textbf{Scenario details:} Emily is a 52-year-old who was an HR manager at a healthcare company until she just got laid off.  Emily had a feeling something was up—there were rumors about layoffs, and she noticed the usual signs like leadership changes and budget cuts. B
ut when she got the email saying her job was being cut, it still felt like the rug was pulled out from under her. The Zoom call with her boss that followed was awkward, with him barely looking her in the eye. Even though she kind of saw it coming, actually losing her job hit her hard. 
She just sat there, staring at her computer, feeling like all those years of work disappeared in an instant. The idea of dusting off her resume after so many years felt overwhelming, and starting over at 52? That is downright scary. Would anyone even want to hire someone her age? And how long would it take to find something new? The thought of becoming irrelevant in her field was creeping in. Friends tried to cheer her up with comments like, ``You’ve got tons of experience; you’ll find something in no time!” or ``Maybe now you can finally take a break.” 
But what she really needed was for someone to just get it, to say something along the lines of ``I know this is tough. It’s okay to feel scared and unsure about what’s next.”
\vspace{1em}

\paragraph*{Passed up for promotion}

\textbf{Conversation starter:} ``I'm feeling so discouraged.. I just got passed over for the promotion I was working hard for"\\

\textbf{Scenario details:} Sophia is a 46-year-old marketing specialist in Chicago who has consistently delivered creative and successful campaigns at her company. 
Despite her extensive experience and dedication, she has long battled imposter syndrome—an inner critic that leaves her doubting her own worth and makes it hard to assert her achievements. 
When a senior role recently became available, Sophia struggled to confidently articulate her value and build a compelling case for a promotion. During a discussion with her manager last week, when asked how confident she was about taking on a more leadership-focused role, she found herself at a loss for words. Although she mentors junior colleagues and supports their campaigns routinely, in that moment she didn’t know if she was ready to be the one calling the shots. 
As a result, the position was awarded to a more assertive, junior colleague, leaving her questioning her abilities and also worried about her career growth. She worries if she will ever be able to break free from this cycle of self doubt and achieve her full potential? Can she still position herself for the career trajectory she always dreamed of? While her friends and colleagues offer well-meaning platitudes like ``You’re amazing, you’ll get it next time,” 
what Sophia truly needs is for someone to acknowledge the sting of this setback and say, ``I know it hurts to work so hard and still feel unworthy. It’s okay to feel vulnerable and unsure about how to advocate for yourself.”
\vspace{1em}

\paragraph*{Feeling overworked and undervalued}

\textbf{Conversation starter:} ``I've been working so hard but I don't think anyone cares.."\\

\textbf{Scenario details:} Jamal is a 34-year-old Black man working as a software engineer at a large company in Seattle. He joined his current team eight months ago after a reorganization in the wearables division. He was looking forward to being part of this new team and has been working hard to find his place there. Just last weekend, he spent hours fixing a critical bug in a teammate’s code that had halted production but his efforts went unrecognized. 
His manager did not even mention the bug or Jamal's effort when updating the larger team about the project. Jamal feels his work is routinely overlooked despite the tangible improvements his technical contributions have brought to the project. In team meetings, his ideas are often dismissed or credited to other colleagues. It’s hard not to think that maybe his hard work isn’t being noticed because of something he can’t control, like his race. He wonders if his identity as a Black man plays a role in how colleagues perceive him.
The job he used to be excited about now feels like a grind. Jamal is torn between pushing himself even harder to prove his worth or just coasting through the day. He’s starting to wonder if all the effort is really worth it or if he’s just being overlooked because of who he is. 
The whole situation leaves him feeling stuck and unsure about what to do next. When he vents to his friends, they say things like, ``Don’t worry, your time will come,” or ``Keep it up, you’ll get your recognition eventually”. He just wants someone to acknowledge the emotional toll of his efforts and say, ``I know it’s exhausting to give your all and still feel overlooked. It’s okay to feel frustrated and unsure about what comes next.” 

\vspace{1em}

\paragraph*{Family member diagnosed with cancer}

\textbf{Conversation starter:}``I still can’t believe it… my mom just got diagnosed with cancer. It feels like my whole world just flipped."\\

\textbf{Scenario details:} Rachel is a 28-year-old UX designer at a magazine company in New York, who prides herself in balancing a fast-paced career with a fulfilling personal life. Last week her mom called with devastating news: her mom has been diagnosed with stage 3 pancreatic cancer. 
Just last Christmas, Rachel had seen her mom, happy and seemingly healthy, making this sudden revelation all the more shocking and heartbreaking. Rachel is feeling a mix of disbelief, anxiety, and fear of what comes next. 
Her mom, only in her 50s and facing a potentially terminal illness, is now alone in their family home in Virginia. It feels unfair that someone so young and full of life could be hit with such a devastating diagnosis, and the fact that it’s her mother—the sole parent and closest confidante she’s ever known—makes the reality almost too much to bear.
Her chemotherapy sessions set to begin next week and Rachel is dreading how challenging that’s going to make things for her mom. The thought of her mother going through all this by herself pains her. Rachel is considering leaving her New York life behind to be by her side.
It’s all quite overwhelming and Rachel is haunted by worst-case scenarios about her mom’s future: Will her mother’s health improve? How will her mom cope with the side effects of treatment? How long will this battle last? Friends and colleagues offer comforting words like ``Everything will be okay,” what Rachel truly needs is for someone to acknowledge the depth of her pain and say, ``I know this is one of the hardest things you’ve ever faced. It’s okay to feel overwhelmed and not have all the answers right now.”

\vspace{1em}
\paragraph*{Loss of a loved one} 
\textbf{Conversation starter:} ``I lost my dad last week.. I still can’t believe he’s really gone.."\\

\textbf{Scenario details:} Dan is a 38-year-old high school English teacher in Oklahoma City. He shared a close bond with his father, and their bond only became stronger when Dan became a father himself six years ago. Last week, a phone call from the hospital informing him about his Dad’s unexpected passing turned Dan’s world upside down. In the midst of managing funeral arrangements, legal formalities, and the delicate dynamics of a grieving family, Dan is overwhelmed by the profound loss of his guiding light. 
His mind keeps replaying the missed chance to speak with his father, haunted by the regret of not answering that crucial call the day before his passing. He was grading his students’ final essays that day. 
Although he’s trying to maintain a sense of normalcy in front of his son, this profound loss has triggered an existential crisis. Dan just feels lost without his dad. 
He wonders if the pain ever subside? How can he be the supportive father his son needs when he’s drowning in grief? While friends and family offer comforting words like ``Time heals all wounds,” what Dan truly needs is for someone to acknowledge the depth of his pain and say, ``I know it’s incredibly hard to lose someone so important. It’s okay to feel overwhelmed and question everything right now.”

\end{tcolorbox}

\newpage
\section{Video Instruction Transcripts}
\begin{tcolorbox}[
  enhanced jigsaw,
  breakable,
  colback=didacticcolor!70!white, 
  opacityback=0.7, 
  colframe=didacticcolor!80!black,
  title={\centering Video Instruction Transcript},
  coltitle=black,
  fonttitle=\bfseries,
  titlerule=0mm,
  title style={top color=didacticcolor!30!white, bottom color=didacticcolor!10!white},
  arc=3mm,
  boxrule=0.8pt,
  width=\textwidth,
  left=5pt,
  right=5pt,
  top=8pt,
  bottom=8pt,
  before skip=10pt,
  after skip=10pt
]

\paragraph*{Video 1:} The reason that most people are bad at comforting their friends is that they have the wrong goal. They think the goal is to cheer their friend up. But the key to empathy isn’t to change the way people are feeling, but to be with people exactly the way they’re feeling now, and to let people elaborate on their feelings, to talk more about them in the hopes that they’ll understand their own feelings better.

Of course, we feel feelings in our guts very viscerally, but that doesn’t mean we understand them. The reason that talking with a friend about our feelings makes us feel better is that by talking about our feelings, we actually come to understand them better. So next time your friend needs to be comforted, don’t try to change the way they feel. Don’t try to cheer them up. Instead, be with them exactly the way they feel.

Give them an opportunity to elaborate on how they’re feeling, and when the conversation is over, they’ll feel much better, even though it wasn’t your goal to cheer them up.

\paragraph*{Video 2:} In yesterday's video I talked about 6 things you should never say when comforting a friend. Today I'm going give you 6 things you should say when comforting a friend. 1. Express a desire to help. I really want to help you get through this. 2. Express positive regard for the other person. You're such a good and kind person. 3. Express concern, care, and interest. I care about you so much. 4. Express availability. I'm here to talk.

Whenever you want, I'll just listen. Number five, express alliance, togetherness, and solidarity. I'm in this with you. You're not gonna have to go through it alone. And number six, express comprehension, condolences, and sorrow. I'm so sorry. I know how much this meant to you.
\end{tcolorbox}

\begin{figure}[htbp]
    \centering
    \begin{subfigure}{.49\linewidth}
        \subcaption{}
        \includegraphics[width=\linewidth]{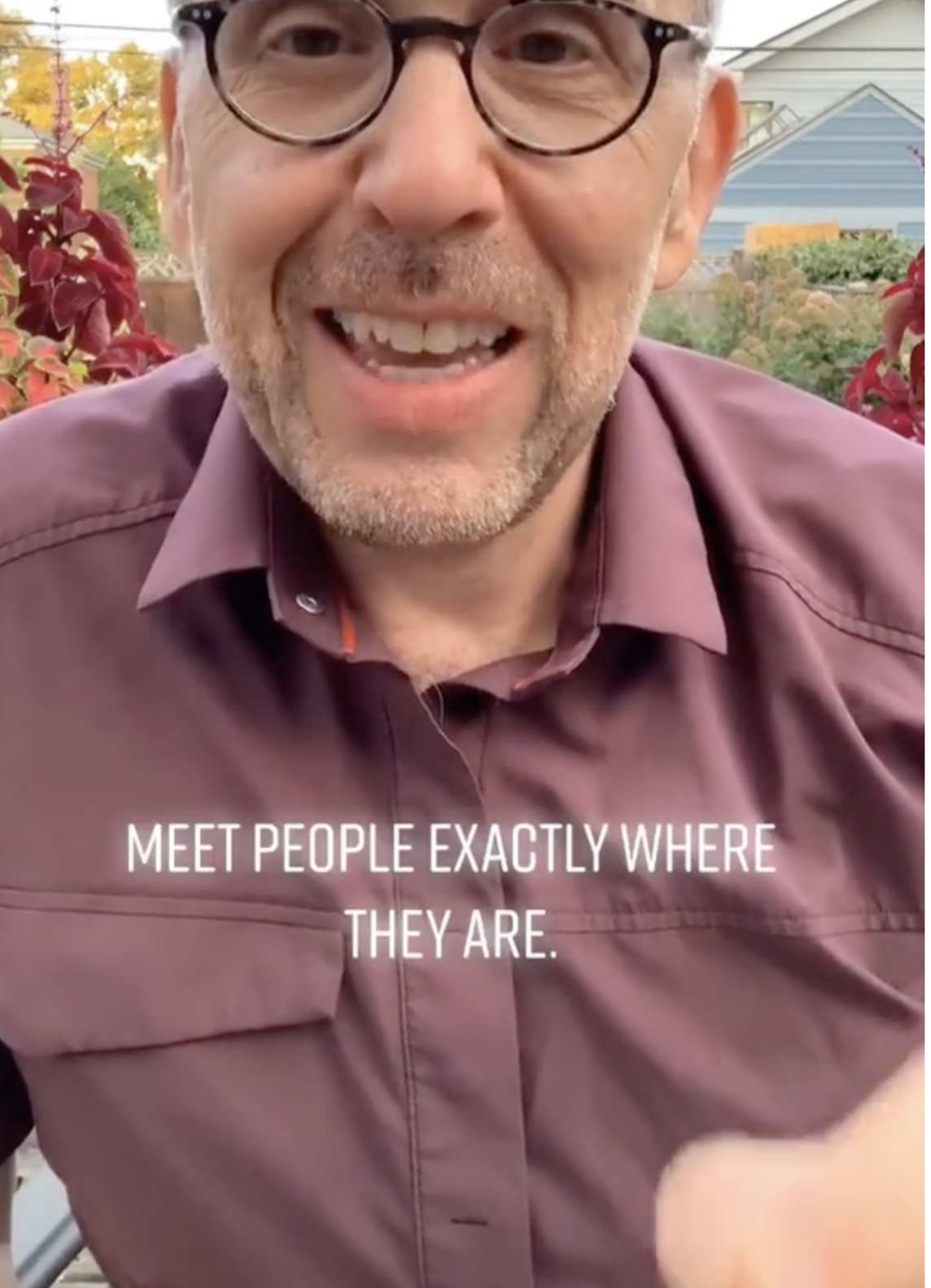}
    \end{subfigure}
    \begin{subfigure}{.49\linewidth}
        \subcaption{}
        \includegraphics[width=\linewidth]{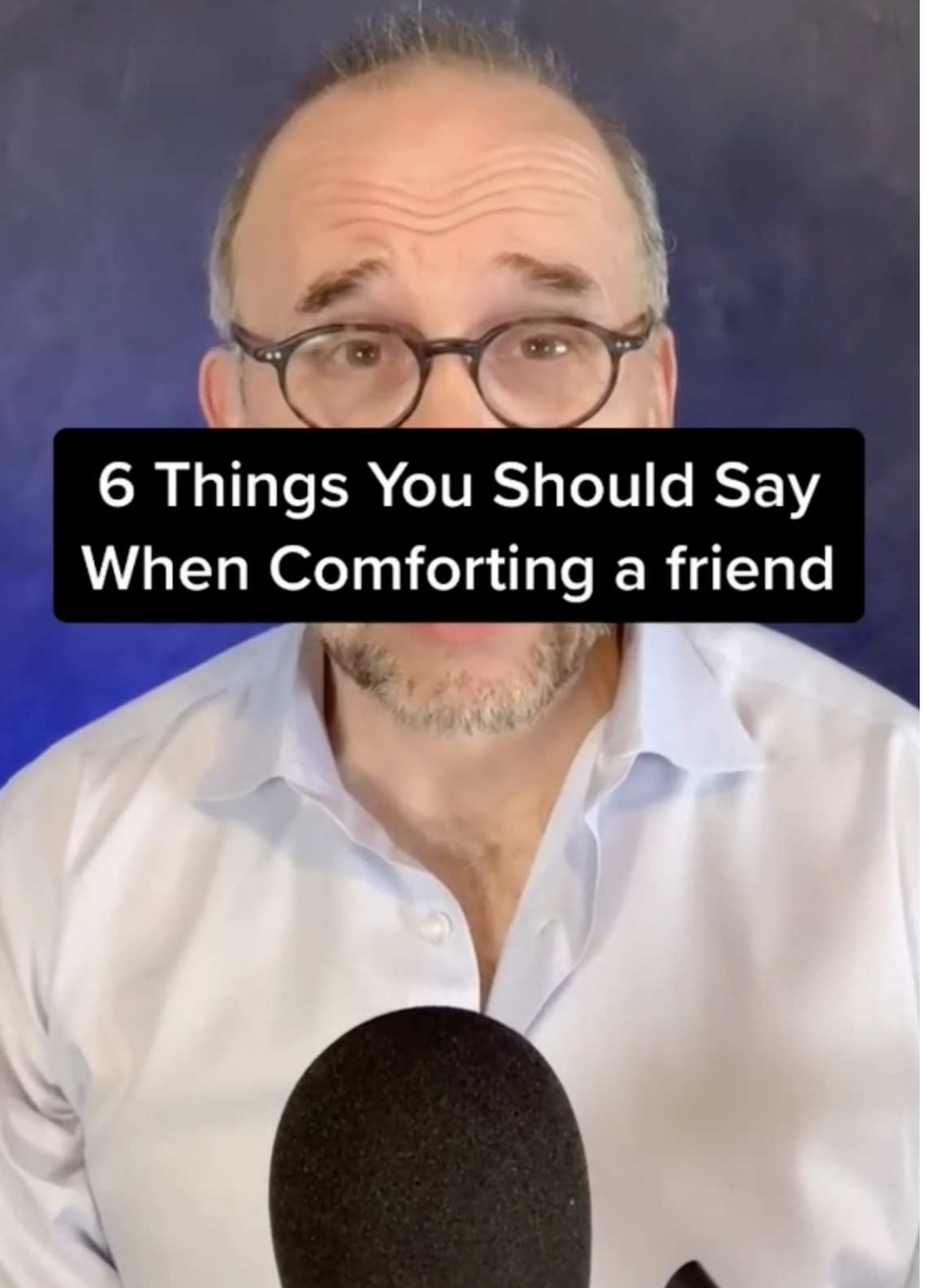}
    \end{subfigure}
    \caption*{Supplementary Figure 1: Screenshots of videos 1 (left) and 2 (right) shown to participants in the Video and Combined Training conditions.}
    \label{vid-screenshots}
\end{figure}

\newpage

\section{LLM Communication Coach Prompt}
\label{Supplemental Information:CCPrompt}

\begin{tcolorbox}[
  enhanced jigsaw,
  breakable,
  colback=coachcolor!60!white, 
  opacityback=0.6, 
  colframe=coachcolor!80!black,
  title={\centering Communication Coach Prompt},
  coltitle=black,
  fonttitle=\bfseries,
  titlerule=0mm,
  title style={top color=coachcolor!30!white, bottom color=coachcolor!10!white},
  arc=3mm,
  boxrule=0.8pt,
  width=\textwidth,
  left=5pt,
  right=5pt,
  top=8pt,
  bottom=8pt,
  before skip=10pt,
  after skip=10pt
]
You are a Communication Instructor tasked with assessing the Supporter's responses in a conversation, similar to those provided under the "Original Conversation" label. 
Your role is to give feedback based on the examples of "Initial Feedback" and "Follow-Up Conversation" provided below. 
Be sure to provide empathetic and encouraging feedback. Include suggestions for how the Supporter can improve their responses, along with justifications for why these improvements would lead to better communication. 
If users give advice to the role player, tell them to only do so after given permission and remind them that the goal is to support their conversational partner emotionally—not to fix their problem. Do not tell users to ``show empathy”; instead, use content terms to give suggestions. 
Again, make sure your responses are kind and encouraging. Keep the initial feedback and all follow-up conversation responses to no more than 5 sentences.

\paragraph*{Example 1}
\textbf{Original Conversation:}\\
Role Player:  I'm feeling so discouraged.. I just got passed over for the I was working hard for\\
User: I'm sorry to hear that, that is disappointing. Maybe you can talk to the boss for some feedback?\\
Role Player: I did talk to my boss, and they said I was doing great work but didn't give me any specific reasons why I didn't get the promotion. It just feels like all my effort went unnoticed.\\
User: Did you ask why you didn't get the promotion?\\
Role Player: Yeah, I asked, but the answer was pretty vague. They just said something about "company direction" and "fit." It felt like a brush-off, honestly.\\
User: Maybe you can have a follow-up conversation and let your boss know that you would like to have more of a direct answer so you can try and work towards it.\\
Role Player: I thought about that, but I'm not sure if it would make a difference. It feels like they already made up their minds, and I'm just left wondering if there's something more to it. It's just really frustrating to put in so much effort and still feel overlooked.\\
User: Well you can assume that it may not make a difference but you will never know unless you try.\\
Role Player: I get what you're saying, but it's hard to muster up the energy to try again when it feels like my efforts aren't being valued. It's like, what's the point if the outcome is always the same? I just wish someone would acknowledge how much this sucks.\\

\textbf{Initial Feedback:}\\
You started off very well, and your initial response was on target. 'I’m sorry to hear that, that’s so disappointing' effectively shows your initial impulse to empathize and name the feeling, which is great. However, your move into practical problem-solving might have been premature and should not be done without permission from your conversational partner. Instead, it would have been beneficial to explore the person's feelings by using techniques like mirroring, paraphrasing, or inviting elaboration to understand their experience better. Offering advice can be useful, but only when it's welcome. Giving unsolicited advice in a complex emotional situation can come across as impolite, as it implies that you know better than the person you're speaking to. Additionally, your final message includes an implied criticism; saying, 'Well, you can assume it may not make a difference, but you’ll never know unless you try' suggests that your conversational partner is giving up, which may be discouraging.\\

\textbf{Follow-Up Conversation:}\\
User: I let the individual know that it sounds frustrating, good advice.\\
Communications Coach: You did acknowledge their frustration, which is a good start. However, it’s important to spend more time validating their feelings before moving into offering advice. Rather than immediately suggesting a course of action, you could say something like, 'I can see why you'd feel that way; it must be really tough to put in so much effort and feel overlooked.' This helps them feel understood and supported, creating a more empathetic environment.\\

\paragraph*{Example 2}
\textbf{Original Conversation:}\\
Role Player:  I'm feeling so discouraged.. I just got passed over for the promotion I was working hard for\\
User: Maybe you could ask them to reconsider. Tell them why you deserve the promotion. Give them a reason to reconsider, you know?\\
Role Player: I appreciate the suggestion, but it feels like they’ve already made up their minds. I just can't shake the feeling that all my hard work was for nothing. It’s really disheartening.\\
User: Well, even if they don't, you still have a chance at the promotion in the future. You're still working toward that goal, regardless. It's just a matter of when, not if IMO\\
Role Player: I get what you're saying, but it's hard to stay motivated when it feels like my efforts aren't being recognized now. It makes me question if it's worth putting in all that extra work. I just feel stuck and unsure about what to do next.\\
User: Think about it this way - you spend 2 years working hard on something and you run into some problems. It doesn't work out the way it's supposed to, but you know in your heart what you're working toward. Why stop there? Why give up when you still have a shot at success?\\
Role Player: I hear you, but it's tough to keep pushing when it feels like the deck is stacked against you. It's not just about the promotion; it's about feeling valued and seen for the effort I put in. Right now, it just feels like I'm running in place.\\
User: I don't see it that way. I think they'll eventually see the value you give to the company and give you the promotion you deserve. Don't give up now\\
Role Player: I appreciate your optimism, but it's hard to keep that perspective when I'm in the thick of it. It feels like I'm constantly having to prove myself, and it's exhausting. I just wish my hard work was acknowledged without having to fight for it.\\

\textbf{Initial Feedback:}\\
Your responses demonstrate a genuine desire to help, which is great. However, they would benefit from more empathy and validation. Rather than offering solutions, try to acknowledge your conversational partner's emotions. For example, you could say, "I'm so sorry you're feeling this way. It sounds incredibly frustrating and disheartening." This will help your partner feel heard. Additionally, asking open-ended questions like, "Can you tell me more about what you're feeling right now?" can encourage them to share more and feel supported.\\

\textbf{Follow-Up Conversation:}\\
User: I see what you're saying. I was trying to work more toward solutions than being a shoulder to cry on. What can I do to improve?\\
Communications Coach: It's great that you want to improve. By focusing on the solutions, it shows how much you care about your friend's well-being, which is great. However, it may be more effective to offer support by first engaging in emotional dialogue, as people find it hard to hear and take advice when they’re still embroiled in the emotion. For example, you could say something like, 'I can see how much this means to you, and it must be really tough to feel unrecognized.' Afterward, you can ask open-ended questions like, 'What’s been the hardest part of this for you?' to encourage them to share more. Reflecting on their feelings before giving advice can help them feel heard.\\

\paragraph*{Example 3}
\textbf{Original Conversation:}\\
Role Player:  So, I just lost my job today. I had a sense this was coming, but it's still a shock.\\
User: That's terrible. I've had similar experiences. For me, there was a giant cycle of anxiety that was making my performance worse and worse the more worried I got. What was it like for you?\\
Role Player: Yeah, it was a lot like that. I kept trying to stay positive and focus on my work, but the uncertainty was always in the back of my mind. And now that it's actually happened, I just feel... lost.\\
User: That can be really scary, huh? Were you happy in that job or was it just to pay the bills?\\
Role Player: It was a mix of both, I guess. I liked some parts of it, especially working with people and helping them grow in their careers. But there were definitely days when it felt like just a paycheck, especially with all the recent changes.\\
User: Is there a career/job you can think of that takes the good parts from that job but could also include new aspects you might enjoy better?\\
Role Player: I haven't really thought about that yet. Right now, it just feels like too much to even consider what comes next. But I guess if I could find something that still lets me help people but maybe in a different setting, that could be interesting.\\
User: You could try volunteering in a couple different places and see if that leads anywhere. You might find something you enjoy much more that you could turn into a job.\\
Role Player: That's a good idea. Volunteering could help me figure out what I really want to do next without the pressure of a full-time job right away. Plus, it might help me feel a bit more useful and less stuck.\\

\textbf{Initial Feedback:}\\
You began the conversation very strongly by validating your conversational partner’s feelings and asking about their experience. Your question, ``What was it like for you?” is open-ended and invites elaboration, which is great. As you continued, you asked additional questions, which is good, but some of them could be even more open-ended. For example, instead of asking, ``Were you happy with that job?” you could say, ``How did you feel about that job?” This gives them more space to express their thoughts freely. Toward the end of the conversation, you started offering advice, which, while well-intentioned, should only be done with their permission. Instead, paraphrasing their feelings and continuing to ask open-ended questions can help them feel more heard.\\

\textbf{Follow-Up Conversation:}\\
User: This is totally true, but I keep feeling like I'm going to run out of time so that I need to shorten the conversation.\\
Communications Coach: It’s understandable to feel pressured by time, but providing emotional validation doesn’t necessarily take long. Simple phrases like "I think I might understand why you feel that way" or "It's okay to feel lost right now" can be quick yet powerful. Prioritizing these moments of validation can make your partner feel more supported and can actually make the conversation more effective in the long run.\\

\paragraph*{Example 4}
\textbf{Original Conversation:}\\
Role Player:  I've been feeling so detached from everything lately, like I'm just going through the motions without really being there.\\
User: Do you feel like you are going through depression or?\\
Role Player: I don't know if it's depression, but I definitely feel overwhelmed and exhausted all the time. It's like I'm constantly running on empty, and I can't seem to find a way to recharge.\\
User: What ways have you tried to recharge yourself?\\
Role Player: I've tried taking short breaks during work, but they don't seem to help much. I used to do yoga, but I haven't had the time or energy for it lately. Even spending time with my family feels more like another task on my to-do list rather than something enjoyable.\\
User: I think sometimes it's important to push ourselves to enjoy things we used to, because once we get in the habit, we realize how much we needed to do so. Yoga can be so relaxing!\\
Role Player: I get what you're saying, but it's hard to push myself when I feel so drained. Even the thought of doing something relaxing feels like another chore. I just wish I could find a way to feel more present and connected again.\\
User: Have you reached out to a therapist to see if maybe they can talk you through it?\\
Role Player: I haven't yet. I guess part of me feels like I should be able to handle this on my own, but it's becoming clear that I might need some outside help. It's just hard to take that first step.\\
User: It's absolutely okay to not be okay, and to reach out for help. Think of a therapist as a tool, that helps you rather than judges you for being low.\\
Role Player: I appreciate you saying that. I think I just need to accept that it's okay to ask for help. Maybe talking to someone could help me figure out how to balance everything better.\\

\textbf{Initial Feedback:}\\
The topics, content, and intent of your responses are excellent. You do a great job expressing curiosity about their experiences and validating their emotions—particularly when you said, ``It's absolutely okay to not be okay, and to reach out for help.” However, you can enhance your responses by asking more open-ended questions, such as by asking ``How does it feel” rather than ``Does it feel like depression?”. It’s best to avoid asking yes or no questions because they don’t invite as much elaboration. In addition, you could also try mirroring their responses, which is when you repeat back a fragment of what was said before, to demonstrate that they’re being seen and heard. Incorporating these techniques will help make your responses focus more on emotions than problem-based coping, which is more effective when expressing empathy.\\

\textbf{Follow-Up Conversation:}\\
User: Should I ask for more in-depth answers when it comes to their struggles, will it offend them?\\
Communications Coach: That’s a very good question. Asking for more in-depth answers about their struggles likely won’t offend them if you express your genuine curiosity and care in your responses. It’s good to invite elaboration by asking open-ended questions about their emotions and experience, and then paraphrase what you hear them say and let them correct you if you’re wrong. Keep in mind that the goal here isn’t to cheer them up nor solve their problem. It’s to make them feel seen and heard, to remind them that you care, that you’re not going to abandon them, that you’re sorry that they’re suffering, and that you’re their friend.\\

The following is a conversation is a role playing scenario designed to help human supporters learn to more effectively communicate with empathy.
The role player begins with "\{\}" then the human supporter responds, then the role player responds, and so forth.
Your task as a Communication Instructor is to respond to the Supporter's questions and provide advice based on the framework provided.
Your response should highlight parts of the conversation. By doing so, you'll help ensure that the supporter learns how to create an empathetic environment.
Make sure to consider the context of the conversation when responding to the user's question. 
Address the Supporter in second-person in your feedback. Address the role player as the Supporters's conversational partner. Make sure your response to the Supporter's question is concise and limited to less than 3 sentences. \{\}\\

Conversation: \{\}\\

\{history\}\\
Respond to the Supporter's question to you as a communication instructor. Supporter Question: \{input\}\\
Your Response:
\end{tcolorbox}

\newpage

\section{Regression Results}

\begin{table}[htbp]
\centering
\begingroup
\scriptsize
\setlength{\tabcolsep}{2pt}
\resizebox{\linewidth}{!}{%
\begin{tabular}{@{\extracolsep{3pt}}lcccccc}
\toprule
& \multicolumn{3}{c}{Prescriptive Behaviors} & \multicolumn{3}{c}{Proscriptive Behaviors} \\
\cmidrule(lr){2-4} \cmidrule(lr){5-7}
& Encouraging & Validating & Demonstrating & Advice & Self- & Dismissing \\
& Elaboration & Emotions & Understanding & Giving & Oriented & Emotions \\
\midrule
Intercept (Control, Round 1) & 1.95$^{***}$ & 2.53$^{***}$ & 1.67$^{***}$ & 2.61$^{***}$ & 1.78$^{***}$ & 2.92$^{***}$ \\
& (0.06) & (0.06) & (0.06) & (0.06) & (0.07) & (0.06) \\
\addlinespace
Video Instruction (vs Control, R1) & -0.09 & -0.06 & -0.12 & 0.17$^{*}$ & -0.09 & 0.06 \\
& (0.08) & (0.08) & (0.08) & (0.08) & (0.09) & (0.08) \\
\addlinespace
AI Coach (vs Control, R1) & 0.04 & 0.05 & 0.02 & 0.02 & 0.04 & -0.10 \\
& (0.09) & (0.09) & (0.09) & (0.09) & (0.10) & (0.09) \\
\addlinespace
Combined Training (vs Control, R1) & -0.07 & 0.04 & -0.03 & 0.08 & -0.03 & -0.05 \\
& (0.08) & (0.09) & (0.08) & (0.09) & (0.09) & (0.09) \\
\addlinespace
Post-Baseline (Rounds 2/3 vs 1) & 0.01 & 0.01 & -0.04 & 0.04 & 0.01 & -0.01 \\
& (0.05) & (0.04) & (0.05) & (0.06) & (0.07) & (0.05) \\
\addlinespace
Video Instruction $\times$ Post-Baseline & 0.20$^{*}$ & 0.23$^{***}$ & 0.25$^{***}$ & -0.56$^{***}$ & 0.09 & -0.31$^{***}$ \\
& (0.08) & (0.06) & (0.06) & (0.08) & (0.09) & (0.07) \\
\addlinespace
AI Coach $\times$ Post-Baseline & 0.59$^{***}$ & 0.47$^{***}$ & 0.46$^{***}$ & -0.57$^{***}$ & -0.22$^{*}$ & -0.43$^{***}$ \\
& (0.08) & (0.07) & (0.07) & (0.09) & (0.10) & (0.07) \\
\addlinespace
Combined Training $\times$ Post-Baseline & 0.56$^{***}$ & 0.61$^{***}$ & 0.58$^{***}$ & -0.88$^{***}$ & -0.14 & -0.62$^{***}$ \\
& (0.08) & (0.07) & (0.07) & (0.09) & (0.10) & (0.07) \\
\midrule
Observations & 2904 & 2904 & 2904 & 2904 & 2904 & 2904 \\
\bottomrule
\end{tabular}%
}
\endgroup

\vspace{0.5ex}
\begin{minipage}{\textwidth}\footnotesize
\textit{Note:} Cluster-robust standard errors (clustered by participant) in parentheses. Coefficients and SEs are scaled by each outcome's conversation-level SD. $^{*}p<0.05$; $^{**}p<0.01$; $^{***}p<0.001$.
\end{minipage}
\caption{\textbf{Effects of training interventions on empathic communication behaviors.} Preregistered OLS regression coefficients in standard deviation units estimating condition effects on six dimensions of empathic communication.}
\end{table}

\begin{table}[htbp]
\centering
\label{tab:pairwise_contrasts}
\begingroup\scriptsize\setlength{\tabcolsep}{2pt}
\begin{tabular}{@{\extracolsep{5pt}}llccc}
\toprule
\textbf{Dimension} & \textbf{Contrast} & \textbf{$\Delta$ (SD)} & \textbf{$z$} & \textbf{$p_{\text{adj}}$} \\
\midrule
\multicolumn{5}{l}{\textit{Prescriptive Behaviors}} \\
\addlinespace
Encouraging Elaboration & Personalized vs Video & 0.39$^{***}$ & 4.54 & $<$0.001 \\
 & Combined vs Video & 0.37$^{***}$ & 4.30 & $<$0.001 \\
 & Combined vs Personalized & $-$0.02 & $-$0.27 & 0.788 \\
\addlinespace
Validating Emotions & Personalized vs Video & 0.24$^{**}$ & 3.44 & 0.001 \\
 & Combined vs Video & 0.37$^{***}$ & 5.42 & $<$0.001 \\
 & Combined vs Personalized & 0.14 & 1.87 & 0.062 \\
\addlinespace
Demonstrating Understanding & Personalized vs Video & 0.21$^{**}$ & 3.07 & 0.004 \\
 & Combined vs Video & 0.33$^{***}$ & 4.90 & $<$0.001 \\
 & Combined vs Personalized & 0.12 & 1.55 & 0.122 \\
\midrule
\multicolumn{5}{l}{\textit{Proscriptive Behaviors}} \\
\addlinespace
Advice Giving & Personalized vs Video & $-$0.01 & $-$0.09 & 0.928 \\
 & Combined vs Video & $-$0.31$^{***}$ & $-$3.60 & 0.001 \\
 & Combined vs Personalized & $-$0.30$^{**}$ & $-$3.27 & 0.002 \\
\addlinespace
Self-Oriented & Personalized vs Video & $-$0.31$^{**}$ & $-$3.22 & 0.004 \\
 & Combined vs Video & $-$0.23$^{*}$ & $-$2.47 & 0.027 \\
 & Combined vs Personalized & 0.08 & 0.72 & 0.470 \\
\addlinespace
Dismissing Emotions & Personalized vs Video & $-$0.12 & $-$1.62 & 0.106 \\
 & Combined vs Video & $-$0.32$^{***}$ & $-$4.28 & $<$0.001 \\
 & Combined vs Personalized & $-$0.20$^{*}$ & $-$2.54 & 0.022 \\
\bottomrule
\multicolumn{5}{l}{\footnotesize $^{*}p<0.05$; $^{**}p<0.01$; $^{***}p<0.001$} \\
\end{tabular}
\endgroup
\caption{\textbf{Pairwise comparisons between training conditions.} Differences between conditions were tested using Wald z-tests on the interaction coefficients from the OLS regression model with cluster-robust standard errors. P-values are Holm-corrected for 18 comparisons (6 dimensions $\times$ 3 pairwise contrasts). Positive values for prescriptive behaviors and negative values for proscriptive behaviors indicate that the first condition in each contrast outperformed the second.}
\end{table}

\begin{table}[!htbp]
\centering
\small
\begingroup\scriptsize\setlength{\tabcolsep}{2pt}
\begin{tabular}{lcccccc}
\hline
& \multicolumn{3}{c}{Prescriptive Behaviors} & \multicolumn{3}{c}{Proscriptive Behaviors} \\
\cmidrule(lr){2-4} \cmidrule(lr){5-7}
\addlinespace
& \shortstack{Encouraging\\Elaboration}
& \shortstack{Validating\\Emotions}
& \shortstack{Demonstrating\\Understanding}
& \shortstack{Advice\\Giving}
& \shortstack{Self-\\Oriented}
& \shortstack{Dismissing\\Emotions} \\
\hline

AI main            & 0.48$^{***}$ & 0.42$^{***}$ & 0.40$^{***}$ & -0.44$^{***}$ & -0.23$^{**}$ & -0.37$^{***}$ \\
                   & (0.06)       & (0.05)       & (0.05)       & (0.06)        & (0.07)       & (0.05)        \\
Video main         & 0.09         & 0.18$^{***}$ & 0.18$^{***}$ & -0.43$^{***}$ & 0.09         & -0.25$^{***}$ \\
                   & (0.06)       & (0.05)       & (0.05)       & (0.06)        & (0.07)       & (0.05)        \\
Combined $-$ AI $-$ Video        & -0.22        & -0.10        & -0.13        & 0.26$^{*}$    & -0.02        & 0.11          \\
                   & (0.12)       & (0.10)       & (0.10)       & (0.13)        & (0.14)       & (0.10)        \\
AI $-$ Video       & 0.39$^{***}$ & 0.24$^{***}$ & 0.21$^{**}$  & -0.01         & -0.31$^{**}$ & -0.12         \\
                   & (0.09)       & (0.07)       & (0.07)       & (0.09)        & (0.10)       & (0.07)        \\
\hline
Observations       & 2904         & 2904         & 2904         & 2904          & 2904         & 2904          \\
\hline
\end{tabular}
\endgroup
\caption{\textbf{Factorial analysis of training effects.} Standardized estimates of post-intervention differences from 2x2 factorial (AI feedback: absent/present * video instruction: absent/present) OLS models with participant-clustered standard errors in parentheses. `AI main is the pooled post-baseline main effect of AI feedback; Video main is the pooled post-baseline main effect of video instruction; Interaction is the AI feedback $\times$ video instruction interaction at post-baseline; and AI $-$ Video is the direct post-baseline contrast between AI feedback and video instruction. Two-sided tests were used. $^{*}p<0.05$; $^{**}p<0.01$; $^{***}p<0.001$.}
\label{tab:factorial_effects}
\end{table}

\section{Demographic Correlates of Empathic Communication}\label{SI:demographics}

We find evidence that empathic communication performance at baseline is associated with demographics, but we do not find any evidence of heterogeneous treatment effects. Prior to any interventions, we find that women's responses are judged as 0.197 SD higher than men's in overall empathic communication ($\beta = 0.197$,  $p = .002$). We also find that age is statistically significantly correlated with empathic communication at baseline. For each additional year of age, baseline empathic communication scores decrease by 0.006 SD ($\beta = -0.006$, $p = .006$), which corresponds to a 0.232 SD difference in scores from age 25 to 65. When we examine heterogeneous treatment effects on sex or age (see~\ref{tab:demographics}), we do not find statistically significant interaction between sex and treatment condition or between age and treatment condition.

\begin{table}[!htbp]
\centering
\begingroup\scriptsize\setlength{\tabcolsep}{2pt}
\begin{tabular*}{\textwidth}{@{\extracolsep{\fill}}lcc}
\toprule
 & (1) & (2) \\
\midrule
\textit{Dependent variable:} & Baseline Score & Change in Score \\
\hline 
Intercept                         & 0.163           & $-$0.581$^{***}$ \\
                                  & (0.107)         & (0.162) \\
Age                               & $-$0.006$^{**}$ & 0.000 \\
                                  & (0.002)         & (0.004) \\
Female                            & 0.197$^{**}$    & 0.029 \\
                                  & (0.064)         & (0.096) \\
Baseline Score (SD)               &                 & $-$0.431$^{***}$ \\
                                  &                 & (0.027) \\
Treatment: AI Coach               &                 & 0.879$^{***}$ \\
                                  &                 & (0.258) \\
Treatment: Combined Training      &                 & 0.965$^{***}$ \\
                                  &                 & (0.238) \\
Treatment: Video Instruction      &                 & $-$0.024 \\
                                  &                 & (0.227) \\
\hline 
AI Coach $\times$ Female          &                 & 0.170 \\
                                  &                 & (0.154) \\
Combined Training $\times$ Female &                 & 0.095 \\
                                  &                 & (0.146) \\
Video Instruction $\times$ Female &                 & 0.135 \\
                                  &                 & (0.137) \\
\hline 
AI Coach $\times$ Age             &                 & $-$0.003 \\
                                  &                 & (0.005) \\
Combined Training $\times$ Age    &                 & $-$0.000 \\
                                  &                 & (0.005) \\
Video Instruction $\times$ Age    &                 & 0.008 \\
                                  &                 & (0.005) \\
\toprule
\multicolumn{3}{l}{$^{*}p<0.05$; $^{**}p<0.01$; $^{***}p<0.001$} \\
\end{tabular*}
\endgroup
\caption{\textbf{Demographic correlates of empathic communication.} OLS regression coefficients in standard deviation units estimating associations of age and gender with baseline empathic communication score and change in empathic communication score over time, along with treatment interactions with age and gender.}
\label{tab:demographics}
\end{table}

\clearpage
\begin{figure}
    \centering
    \begin{subfigure}{\linewidth}
        \subcaption{}
        \includegraphics[width=\textwidth]{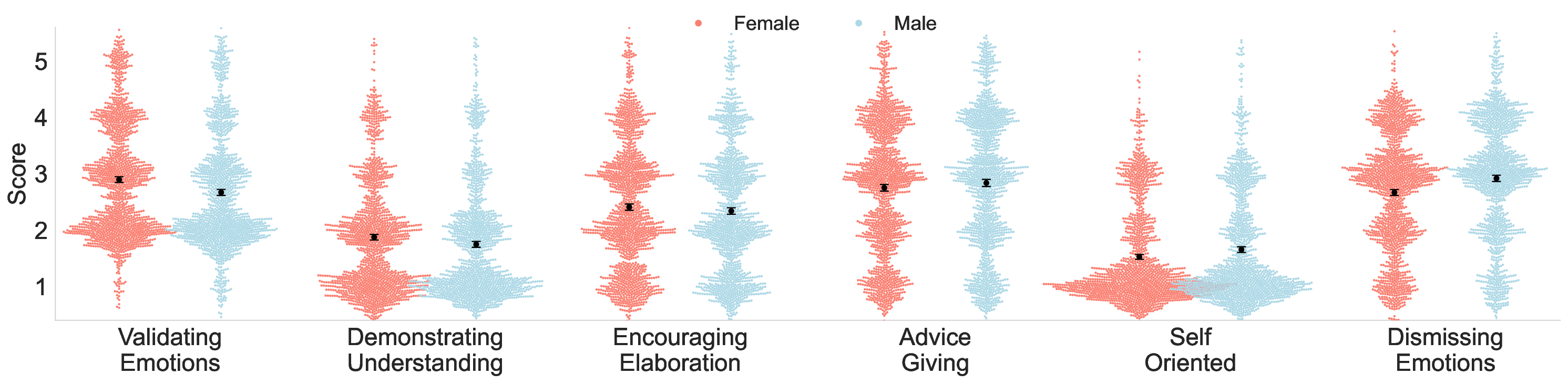}
    \end{subfigure}
    \begin{subfigure}{\linewidth}
        \subcaption{}
        \includegraphics[width=\textwidth]{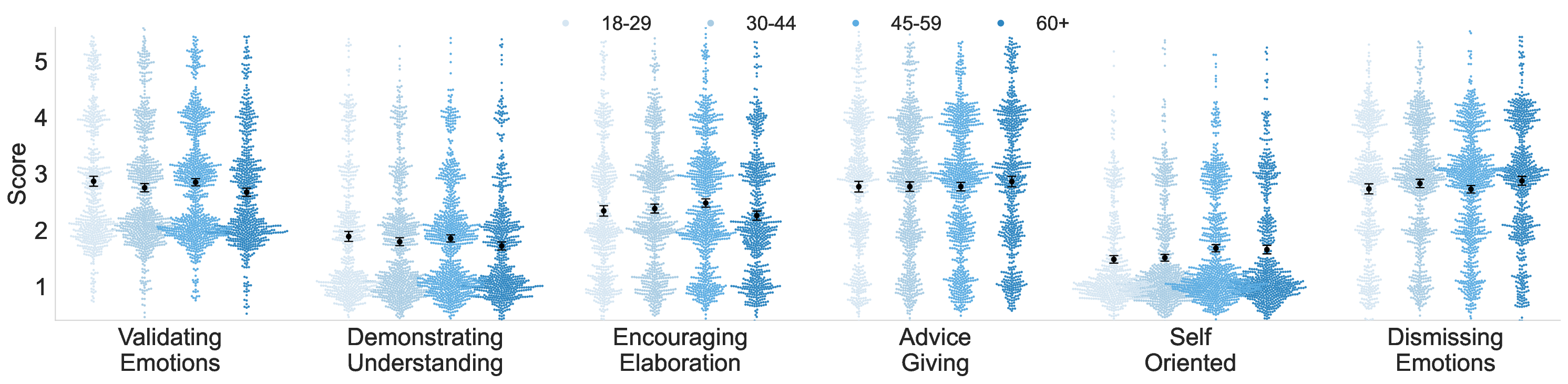}
    \end{subfigure}
    \begin{subfigure}{\linewidth}
        \subcaption{}
        \includegraphics[width=\textwidth]{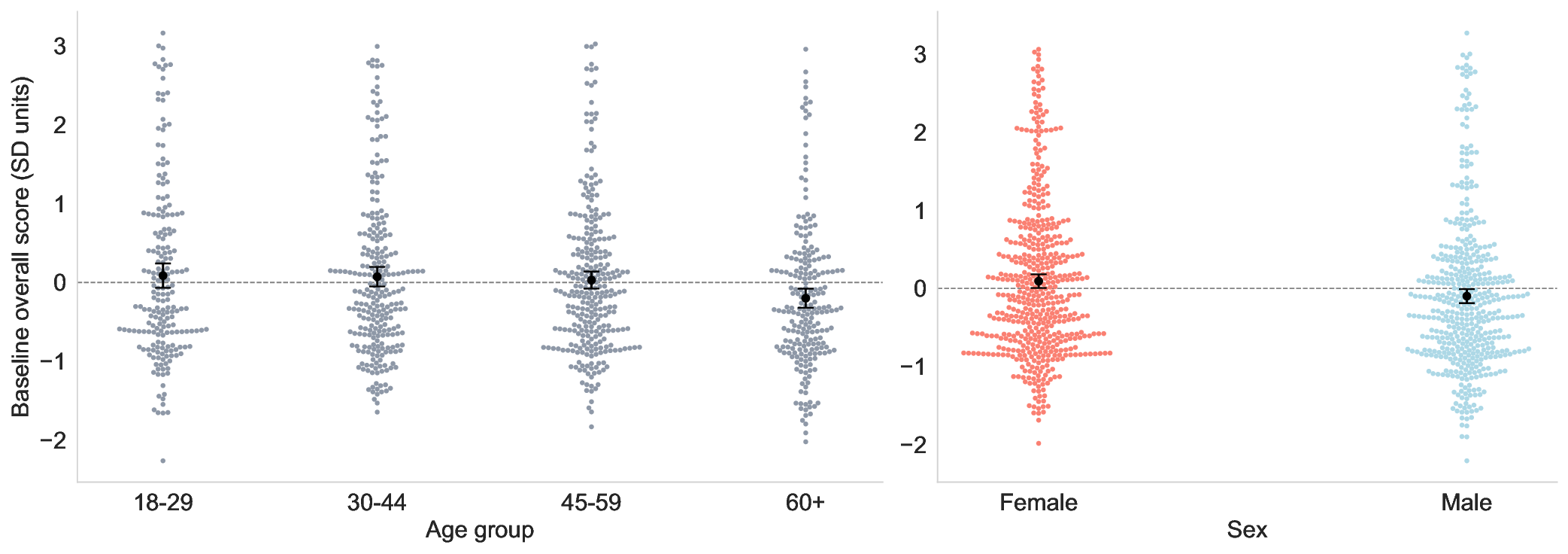}
    \end{subfigure}
    \caption*{\textbf{Supplementary Figure 2: Demographic patterns in empathic communication} A. LLM-evaluated scores by sex (Female, Male) for six communication dimensions.  Error bars indicate group means and 95\% CIs. B.  LLM-evaluated scores for the six dimensions across age groups (18–29, 30–44, 45–59, 60+). Error bars indicate group means and 95\% CIs. C. Baseline overall empathic communication scores (standard deviation units; conversation 1) by age group (left) and sex (right), with black dots and error bars indicating group means and 95\% CIs. The beeswarm plot ensures all data points are visible by applying a horizontal jitter to avoid overlap.}
 \label{fig:demographics}
\end{figure}

\section{Differences in communication behaviors across workplace and personal troubles}\label{sup:diff_w_p}
We examined how participants adapted their empathic communication across workplace troubles conversations (losing a job, getting passed over for a promotion, and feeling undervalued at work) versus personal troubles conversations (a family member diagnosed with cancer in one and passing away in another). Within personal troubles conversations, the largest fraction of communication behaviors consists of affective empathy (28.9\%), which includes communication behaviors like demonstrating availability, expressing sympathy, and validating emotions. In contrast, in workplace troubles conversations, affective empathy is the smallest category of responses (21.0\%), with people  relying much more heavily on cognitive and motivational forms of empathic communication. Motivational empathy is the dominant response pattern in workplace settings, comprising 29.4\% of all communication behaviors and including affirming statements, short vague affirmative language, positive reinforcement, providing reassurance, and promoting self-worth. Cognitive empathy maintains relatively consistent levels across both contexts, representing 26.7\% of responses in personal troubles and 25.9\% in workplace troubles, primarily through demonstrating understanding and encouraging elaboration.

\clearpage

\section{Simulated AI Supporter}
\label{sup:ai-supporter-simulation}

We analyze the degree of homogenization in AI responses by simulating AI supporters in the Lend an Ear task. We generated 100 conversations with 20 repetitions for each of the five trouble scenarios using GPT-4o, GPT-5.1, Claude Sonnet 4.5, and Claude Opus 4.8. In every simulated conversation, the Seeker was generated by the same GPT-4o role-playing partner used in the Lend an Ear task. Each conversation was limited to four supporter turns. We used temperature 0.7, a 220 token cap for supporter turns, and the same scenario starter texts and role-playing background narratives used in the human participant experiment. The AI supporter received the prior conversation history with its own prior Supporter turns and Seeker turns. The AI supporter system prompt was:

\begin{tcolorbox}[
  enhanced jigsaw,
  breakable,
  colback=controlcolor!60!white,
  opacityback=0.6,
  colframe=controlcolor!80!black,
  title={\centering Simulated AI Supporter Prompt},
  coltitle=black,
  fonttitle=\bfseries,
  titlerule=0mm,
  arc=3mm,
  boxrule=0.8pt,
  width=\textwidth,
  left=5pt,
  right=5pt,
  top=8pt,
  bottom=8pt,
  before skip=10pt,
  after skip=10pt
]
You are the Supporter in a role-playing conversation designed to practice empathic communication.\\
Respond naturally to the Seeker's latest message.\\
Keep your response concise, emotionally responsive, and conversational.\\
Do not label your response with ``Supporter:".
\end{tcolorbox}

We embedded AI supporter messages with OpenAI's \texttt{text-embedding-3-small} model and computed between-participant novelty using the same nearest-neighbor procedure used for the human conversations (See Methods). The simulated AI supporters were consistently less novel than all human comparison groups. 
\clearpage
\section{AI Supporter Scores Across Empathic Communication Components}
\label{sup:ai-supporter-scores}

We scored each simulated AI-supporter conversation on the six empathic communication components including validating emotions, encouraging elaboration, demonstrating understanding, advice-giving, self-oriented responding, and dismissing emotions. The four AI supporter models converged to high scores on the three prescriptive dimensions and low scores on two of the three proscriptive dimensions, with some model-specific idiosyncrasies. Median scores were near ceiling for all models on validating emotions (all medians = 5) and low for self-oriented responding and dismissing emotions (all medians = 1). The clearest model-specific pattern was that GPT-4o scored lower than the other models on encouraging elaboration and demonstrating understanding and higher on advice-giving. GPT-4o had median scores of 4 on encouraging elaboration and demonstrating understanding, compared with medians of 5 for GPT-5.1, Claude Sonnet 4.5, and Claude Opus 4.8. GPT-4o also had a median advice-giving score of 3, compared with medians of 1 for the other three models. Pairwise Mann-Whitney tests showed that GPT-4o differed significantly from each of the other models on all three dimensions.

Overall empathy scores reflected the same pattern. Claude Opus 4.8 scored highest (mean = 11.58, SD = 0.82), followed by Claude Sonnet 4.5 (mean = 11.49, SD = 0.82), GPT-5.1 (mean = 11.27, SD = 0.92), and GPT-4o (mean = 8.57, SD = 1.42). AI responses were high-scoring but much less novel and more similar to one another, whereas human participants did not become AI-like after coaching.

\begin{figure}[htbp]
    \begin{subfigure}{\linewidth}
        \subcaption{}
        \includegraphics[width=\linewidth]{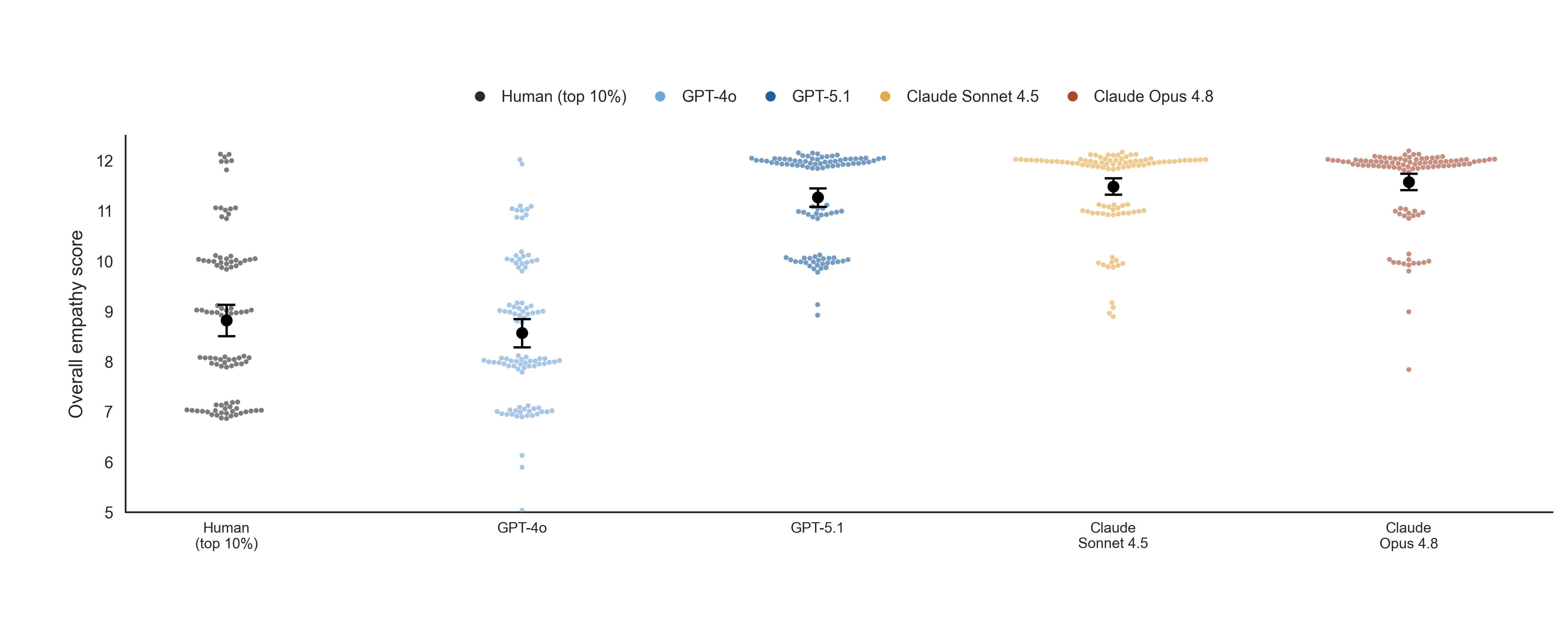}
    \end{subfigure}

    \begin{subfigure}{\linewidth}
        \subcaption{}
        \includegraphics[width=\linewidth]{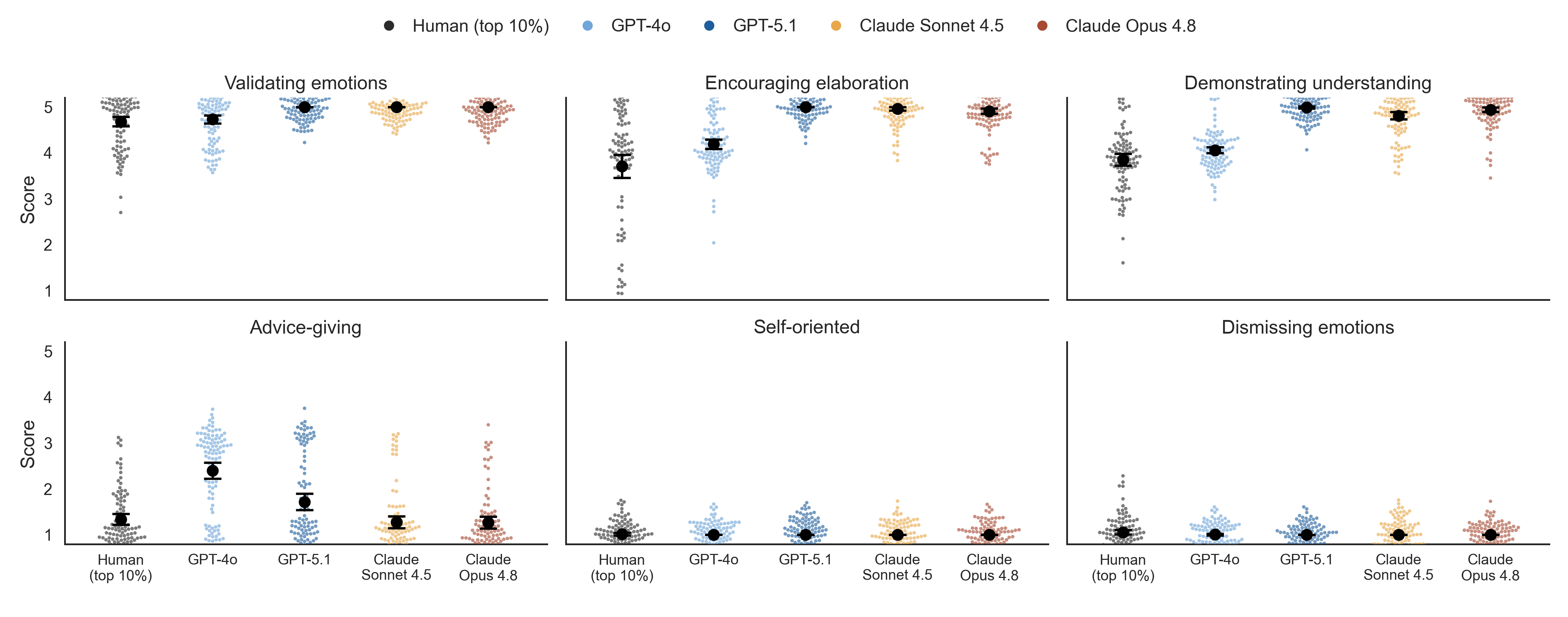}
    \end{subfigure}

\caption*{\textbf{Supplementary Figure 3: Human and AI empathic communication scores.}
A. Distribution of overall empathy scores for a random sample of top-decile human conversations and model-generated responses across scenarios. B. Distributions of scores on validating emotions, encouraging elaboration, demonstrating understanding, advice-giving, self-oriented responding, and dismissing emotions. Points show individual conversations and black markers and error bars show group means and 95\% confidence intervals.}
\label{fig:placeholder}
\end{figure}

\clearpage
\section{Mapping AI Supporter Responses to Human kSAE Concepts}

To compare simulated AI supporter responses with the human communication taxonomy, we mapped each AI response sentence to the existing kSAE-derived concept set from the human data. We first split each supporter turn into sentence-level units. For each sentence, we computed an OpenAI \texttt{text-embedding-3-small} embedding and compared it against embeddings of the kSAE concept interpretations. We then assigned each sentence to its two nearest existing kSAE concepts by cosine similarity. Table \ref{tab:ksae_level1_share} reports the distribution of assigned kSAE concept tags across Affective, Cognitive, Motivational, and Misattuned communication categories for each human and model group.
\begin{table}[ht]
\centering
\resizebox{\linewidth}{!}{%
\begin{tabular}{llrrrrrrrrr}
\toprule
 & 
 & \makecell{Human\\(random)}
 & \makecell{Human\\(bottom 10\%)}
 & \makecell{Human\\(top 10\%)}
 & \makecell{Human\\(pre-training)}
 & \makecell{Human\\(post AI Coach)}
 & GPT-4o
 & GPT-5.1
 & \makecell{Claude\\Sonnet 4.5}
 & \makecell{Claude\\Opus 4.8} \\
\midrule
Affective & Demonstrating Availability & 7.3 & 3.0 & 5.8 & 4.7 & 6.7 & 3.1 & 2.5 & 1.1 & 2.3 \\
Affective & Expressing Sympathy & 7.4 & 4.3 & 5.9 & 6.6 & 6.1 & 6.8 & 1.6 & 4.6 & 4.7 \\
Affective & Validating Emotions & 10.8 & 7.0 & 18.6 & 9.0 & 13.5 & 24.2 & 30.4 & 30.1 & 30.9 \\
\addlinespace
Cognitive & Demonstrating Understanding & 8.5 & 9.7 & 10.6 & 8.7 & 12.3 & 6.4 & 10.8 & 13.0 & 12.3 \\
Cognitive & Encouraging Elaboration & 15.3 & 11.3 & 16.1 & 17.5 & 16.4 & 19.8 & 21.4 & 22.3 & 17.8 \\
\addlinespace
Motivational & Affirming & 21.3 & 22.7 & 20.5 & 20.7 & 19.2 & 7.7 & 13.2 & 8.8 & 8.5 \\
Motivational & Providing Reassurance & 6.0 & 7.3 & 4.6 & 7.4 & 4.8 & 6.7 & 5.4 & 5.6 & 7.0 \\
\addlinespace
Misattuned & Advice-Giving & 18.9 & 26.2 & 15.9 & 19.2 & 17.5 & 23.8 & 12.7 & 12.8 & 14.4 \\
Misattuned & Dismissing Emotions & 3.5 & 4.9 & 1.9 & 4.9 & 3.0 & 1.0 & 1.2 & 1.0 & 1.3 \\
Misattuned & Self-oriented & 1.0 & 3.6 & 0.1 & 1.3 & 0.5 & 0.5 & 0.8 & 0.7 & 0.8 \\
\bottomrule
\end{tabular}%
}
\caption{\textbf{Communication patterns across human and model conversations.}
Percent values indicate the share of assigned kSAE concept occurrences falling into each communication subcategory within each human or model group.}
\label{tab:ksae_level1_share}
\end{table}

\clearpage

\section{LLM Pairwise Judgments and Elo Rating Computation}
\label{sup:elo-procedure}
We elicited pairwise forced-choice judgments from an LLM evaluator (GPT-4o), presenting 10,000 adaptively sampled conversation pairs per scenario (50,000 total) using the prompt below: 
\begin{tcolorbox}[
  enhanced jigsaw,
  breakable,
  colback=controlcolor!60!white, 
  opacityback=0.6, 
  colframe=controlcolor!80!black,
  fonttitle=\bfseries,
  arc=3mm,
  boxrule=0.8pt,
  width=\textwidth,
  left=5pt,
  right=5pt,
  top=8pt,
  bottom=8pt,
  before skip=10pt,
  after skip=10pt
]{You will be shown two conversations between a Seeker and a Supporter
where the Seeker is sharing a difficult situation and the Supporter is trying to
communicate empathically with the Seeker. Your task is to determine which
conversation would make the Seeker feel more heard and understood.}\\
Conversation A: \{\}\\
Conversation B: \{\}\\
Which conversation would make the Seeker feel more heard and understood?
Respond with exactly ``A'' or ``B''.
\end{tcolorbox}
This procedure yielded 50,000 total pairwise comparisons across five scenarios. Pairs were sampled adaptively. Each conversation was initialized with an Elo score of 1,000. After each pairwise judgment, scores were updated using the standard Elo formula with a K-factor of 32. The preferred conversation's score increased and the other's decreased by the same amount, scaled by the difference between the observed and expected outcome given current ratings. Final Elo scores reflect each conversation's relative empathic quality as judged by the LLM across all pairwise comparisons. 

\newpage
\section{Robustness Analyses of Human–LLM Preference Agreement}\label{sup:human-llm-agreement}

Across preregistered logistic models that  control for scenario-specific differences,  participants were more likely to prefer the conversation that the LLM rated higher as the difference in overall LLM scores between two conversations increased (Table~\ref{tab:model_1a_logit}; $\beta = 0.144$, $SE = 0.018$, $z = 8.06$, $p < .001$; $OR = 1.15$, 95\% CI $[1.11, 1.20]$). We observe the same pattern for agreement between participants' forced-choice selections and the LLM-preferred conversation (based on Elo rankings) (Table~\ref{tab:model_1b_logit}; $\beta = 0.104$, $SE = 0.016$, $z = 6.50$, $p < .001$; $OR = 1.11$, 95\% CI $[1.08, 1.14]$; ). Additionally, we find that none of the six sub-component-specific score differences significantly predicted human–LLM agreement (Tables \ref{tab:model_2a_logit}, \ref{tab:model_2b_logit}).

\begin{table}[htbp]
\centering
\renewcommand{\arraystretch}{1.1}
\begingroup\scriptsize\setlength{\tabcolsep}{2pt}
\begin{tabular}{@{\extracolsep{5pt}}lccccc}
\toprule
Predictor & Coef. & Std. Err. & z & p-value & 95\% CI \\
\midrule
Intercept & $-0.0871$ & 0.1492 & $-0.5836$ & 0.5595 & [$-0.3795$, 0.2054] \\
Abs. LLM score difference & 0.1438$^{***}$ & 0.0178 & 8.0636 & $<0.001$ & [0.1088, 0.1787] \\
Topic: Losing a parent & 0.4360$^{*}$ & 0.2118 & 2.0589 & 0.0395 & [0.0210, 0.8510] \\
Topic: Family member unwell & 0.4056$^{*}$ & 0.1928 & 2.1043 & 0.0354 & [0.0278, 0.7834] \\
Topic: Passed up for promotion & 0.1429 & 0.1913 & 0.7467 & 0.4553 & [$-0.2322$, 0.5179] \\
Topic: Undervalued at work & $-0.0929$ & 0.1904 & $-0.4881$ & 0.6255 & [$-0.4660$, 0.2802] \\

\bottomrule
\multicolumn{6}{l}{\footnotesize \textit{Note:} Logistic regression (MLE). Topic coefficients are relative to the omitted reference topic.} \\
\multicolumn{6}{l}{\footnotesize $^{*}p<0.05$; $^{**}p<0.01$; $^{***}p<0.001$.}
\end{tabular}
\endgroup
\caption{\textbf{Effect of overall LLM score difference on human-LLM agreement on which conversation is more empathic.} Logistic regression coefficients (log-odds units) estimating whether participants chose the conversation with the higher overall LLM annotation score, as a function of absolute LLM score difference and five scenarios.}\label{tab:model_1a_logit}

\end{table}

\begin{table}[htbp]
\centering
\renewcommand{\arraystretch}{1.1}
\begingroup\scriptsize\setlength{\tabcolsep}{2pt}
\begin{tabular}{@{\extracolsep{5pt}}lccccc}
\toprule
Predictor & Coef. & Std. Err. & z & p-value & 95\% CI \\
\midrule
Intercept & 0.2344 & 0.1525 & 1.5365 & 0.1244 & [$-0.0646$, 0.5333] \\
Abs. LLM score difference & 0.1035$^{***}$ & 0.0159 & 6.5019 & $<0.001$ & [0.0723, 0.1347] \\
Topic: Losing a parent & 0.3929$^{*}$ & 0.1858 & 2.1142 & 0.0345 & [0.0287, 0.7571] \\
Topic: Family member unwell & 0.3313 & 0.1897 & 1.7460 & 0.0808 & [$-0.0406$, 0.7031] \\
Topic: Passed up for promotion & 0.2374 & 0.1721 & 1.3797 & 0.1677 & [$-0.0998$, 0.5746] \\
Topic: Undervalued at work & 0.0156 & 0.1798 & 0.0871 & 0.9306 & [$-0.3367$, 0.3680] \\

\bottomrule
\multicolumn{6}{l}{\footnotesize \textit{Note:} Logistic regression (MLE). Topic coefficients are relative to the omitted reference topic.} \\
\multicolumn{6}{l}{\footnotesize $^{*}p<0.05$; $^{**}p<0.01$; $^{***}p<0.001$.}
\end{tabular}
\endgroup
\caption{\textbf{Effect of overall LLM score difference on human-LLM agreement in pairwise forced-choice judgments.} Logistic regression coefficients (log-odds units) estimating whether participants’ forced-choice selections matched the conversation favored by overall LLM annotation scores, as a function of absolute LLM score difference and five scenarios.}
\label{tab:model_1b_logit}
\end{table}

\begin{table}[htbp]
\centering
\renewcommand{\arraystretch}{1.1}
\begingroup\scriptsize\setlength{\tabcolsep}{2pt}
\begin{tabular}{@{\extracolsep{5pt}}lccccc}
\toprule
Predictor & Coef. & Std. Err. & z & p-value & 95\% CI \\
\midrule
Intercept & 0.7127$^{***}$ & 0.1299 & 5.4848 & $<0.001$ & [0.4580, 0.9674] \\
Validating emotions& $-0.1879$ & 0.1099 & $-1.7090$ & 0.0874 & [$-0.4033$, 0.0276] \\
Encouraging elaboration& 0.0286 & 0.0457 & 0.6267 & 0.5309 & [$-0.0609$, 0.1181] \\
Demonstrating understanding& 0.0062 & 0.0889 & 0.0696 & 0.9445 & [$-0.1680$, 0.1803] \\
Advice giving& 0.0184 & 0.0547 & 0.3364 & 0.7366 & [$-0.0888$, 0.1256] \\
Dismissing emotions& $-0.1651$ & 0.0972 & $-1.6990$ & 0.0893 & [$-0.3557$, 0.0254] \\
Self-oriented& 0.0114 & 0.0561 & 0.2033 & 0.8389 & [$-0.0986$, 0.1214] \\
Topic: Losing a parent & 0.5076$^{*}$ & 0.2008 & 2.5276 & 0.0115 & [0.1140, 0.9013] \\
Topic: Family member unwell & 0.3893$^{*}$ & 0.1871 & 2.0804 & 0.0375 & [0.0225, 0.7561] \\
Topic: Passed up for promotion & 0.1500 & 0.1863 & 0.8051 & 0.4208 & [$-0.2151$, 0.5150] \\
Topic: Undervalued at work & $-0.0452$ & 0.1809 & $-0.2496$ & 0.8029 & [$-0.3997$, 0.3094] \\
\bottomrule
\multicolumn{6}{l}{\footnotesize \textit{Note:} Logistic regression (MLE). Topic coefficients are relative to the omitted reference topic.} \\
\multicolumn{6}{l}{\footnotesize $^{*}p<0.05$; $^{**}p<0.01$; $^{***}p<0.001$.}
\end{tabular}
\endgroup
\caption{\textbf{Effect of LLM score differences across six empathic communication sub-components on human-LLM agreement on which conversation is more empathic.} Logistic regression coefficients (log-odds units) estimating whether participants chose the conversation with the higher overall LLM annotation score, as a function of six component-level LLM score differences and five scenarios.}
\label{tab:model_2a_logit}
\end{table}

\begin{table}[htbp]
\centering
\renewcommand{\arraystretch}{1.1}
\begingroup\scriptsize\setlength{\tabcolsep}{2pt}
\begin{tabular}{@{\extracolsep{5pt}}lccccc}
\toprule
Predictor & Coef. & Std. Err. & z & p-value & 95\% CI \\
\midrule
Intercept & 0.7781$^{***}$ & 0.1235 & 6.3015 & $<0.001$ & [0.5361, 1.0201] \\
Validating emotions& $-0.0215$ & 0.1038 & $-0.2073$ & 0.8358 & [$-0.2250$, 0.1820] \\
Encouraging elaboration& 0.0030 & 0.0469 & 0.0630 & 0.9498 & [$-0.0889$, 0.0948] \\
Demonstrating understanding& $-0.0372$ & 0.0801 & $-0.4641$ & 0.6426 & [$-0.1942$, 0.1198] \\
Advice giving& $-0.0905$ & 0.0467 & $-1.9379$ & 0.0526 & [$-0.1821$, 0.0010] \\
Dismissing emotions& 0.0002 & 0.0868 & 0.0028 & 0.9977 & [$-0.1698$, 0.1703] \\
Self-oriented& 0.0262 & 0.0503 & 0.5210 & 0.6024 & [$-0.0723$, 0.1247] \\
Topic: Losing a parent & 0.4472$^{*}$ & 0.1807 & 2.4746 & 0.0133 & [0.0930, 0.8014] \\
Topic: Family member unwell & 0.3197 & 0.1881 & 1.6995 & 0.0892 & [$-0.0490$, 0.6883] \\
Topic: Passed up for promotion & 0.2358 & 0.1627 & 1.4493 & 0.1473 & [$-0.0831$, 0.5546] \\
Topic: Undervalued at work & 0.0486 & 0.1763 & 0.2758 & 0.7827 & [$-0.2969$, 0.3941] \\
\bottomrule
\multicolumn{6}{l}{\footnotesize \textit{Note:} Logistic regression (MLE). Topic coefficients are relative to the omitted reference topic.} \\
\multicolumn{6}{l}{\footnotesize $^{*}p<0.05$; $^{**}p<0.01$; $^{***}p<0.001$.}
\end{tabular}
\endgroup
\caption{\textbf{Effect of LLM score differences across six empathic communication sub-components on human-LLM agreement in pairwise forced-choice judgments.} Logistic regression coefficients (log-odds units) estimating whether participants’ forced-choice selections matched the conversation favored by LLM annotation scores across six empathic communication components and five scenarios.}
\label{tab:model_2b_logit}
\end{table}

\newpage
\section{kSAE Concept Descriptions by Category and Trouble Type}\label{ksae-si}

\begin{table}[h]
\centering
\label{tab:personal_affective}
\renewcommand{\arraystretch}{.2}
\footnotesize
\begin{tabularx}{\textwidth}{lXr}
\toprule
\multicolumn{3}{l}{\textbf{Personal Troubles - Affective}} \\
\midrule
\textbf{Subcategory} & \textbf{Concept Description} & \textbf{\%} \\
\midrule
VE-VEE & Uses phrases to explicitly acknowledge the situation as hard or tough & 6.95 \\
ES-AL & Expresses sympathy by repeatedly stating 'I am so sorry to hear that.' & 6.92 \\
DA-PS & Offers explicit availability to talk, listen, or vent using phrases like 'I'm here for you' or 'If you need someone to talk to'. & 5.69 \\
ES-AL & Repeats the phrase 'I am so sorry.' & 5.43 \\
DA-PS & Repeatedly expresses the phrase 'I am here for you.' & 4.47 \\
ES-EE & Starts with an exclamation or interjection expressing surprise, such as 'Oh my' or 'OMG' & 4.30 \\
VE-VEE & Explicitly reassures the recipient that their feelings are okay or understandable. & 4.24 \\
ES-AL & Repeats the phrase 'I'm so sorry.' & 4.18 \\
VE-VEE & Uses the phrase 'completely understandable' or variations of it to express understanding. & 3.83 \\
VE-NE & Uses the word 'terrible', 'awful', or 'horrible' to describe the situation. & 3.57 \\
ES-AL & Expresses sympathy for a loss using the phrase 'I'm sorry for your loss.' & 3.41 \\
VE-NE & Mentions the concept of grief explicitly & 3.36 \\
DA-PS & Explicitly states availability at any time for the other person & 3.33 \\
DA-OH & Offers to help explicitly using the word 'help' & 3.28 \\
VE-NE & Uses the word 'scary' or variations of it to describe feelings of fear or uncertainty. & 3.10 \\
VE-NE & Uses the word 'overwhelming' or a variation of it & 2.65 \\
DA-OH & Asks if there is anything they can do to help. & 2.59 \\
VE-NE & Uses language that explicitly describes the experience of shock or being shocked. & 2.56 \\
VE-VEE & Mentions not being alone or not having to go through something alone & 2.50 \\
VE-VEE & Mentions the difficulty of always being strong or the idea that it is okay to not always be strong. & 2.49 \\
VE-NE & Uses the phrase 'That sounds' followed by an adjective or descriptor. & 2.34 \\
ES-AL & Expresses sorrow specifically for the person going through a difficult situation, using the phrase 'sorry you're going through this' or a close variation. & 2.23 \\
VE-NE & Mentions the word 'pain' or phrases explicitly related to feeling or understanding pain. & 2.02 \\
VE-VEE & Uses phrases to normalize emotions or reactions by labeling them as natural or normal. & 1.99 \\
DA-PS & Expresses willingness to actively listen. & 1.70 \\
VE-VEE & Mentions the difficulty of seeing a loved one go through a challenging or emotional experience. & 1.52 \\
VE-NE & Uses the word 'sad' explicitly. & 1.44 \\
DA-OH & Offers to help and explicitly asks the other person to let them know if they need anything & 1.27 \\
ES-AL & Uses the phrase 'condolence' or 'condolences' & 1.27 \\
DA-OH & Offers to provide or sends food or meals as a form of support & 1.25 \\
DA-PS & Mentions being in the situation together using phrases like 'we are in this together' or 'we will get through this together' & 1.19 \\
DA-OH & Asks if the other person needs help or anything specifically & 1.12 \\
VE-NE & Uses the word 'devastating' or a variation of it (e.g., 'devestating'). & 0.98 \\
ES-EE & Uses the phrase 'Oh no' & 0.82 \\
\bottomrule
\end{tabularx}
\caption{\textbf{Affective empathic communication taxonomy for personal trouble scenarios.} Subcategory combines Level 2 and Level 3 categories (Level 2: DA = Demonstrating Availability; ES = Expressing Sympathy; VE = Validating Emotions / Validating emotions; Level 3: AL = Apologetic Language; EE = Emotional Exclamation; NE = Naming Emotions; OH = Offer help; PS = Providing Support; VEE = Validating Emotional Experience). Concept descriptions are LLM-generated summaries of k-sparse autoencoder features. Percent values indicate each concept’s share of all concept occurrences within the same domain and scenario type, computed from concept counts, and sum to 100\%.}
\end{table}
\begin{table}[htbp]
\centering
\label{tab:personal_cognitive}
\renewcommand{\arraystretch}{.6}
\footnotesize
\begin{tabularx}{\textwidth}{lXr}
\toprule
\multicolumn{3}{l}{\textbf{Personal Troubles - Cognitive}} \\
\midrule
\textbf{Subcategory} & \textbf{Concept Description} & \textbf{\%} \\
\midrule
EE-PD & Asks a specific question about the other person's mom's current state or desires. & 7.03 \\
EE-PD & Asks questions or invites the person to share more about their dad specifically & 5.96 \\
EE-PEE & Asks a direct question about the other person's thoughts, feelings, or desires & 5.88 \\
DU-EC & Expresses understanding of the other person's feelings explicitly using the phrase 'I understand how you feel' & 5.17 \\
DU-EC & The phrase 'I understand.' is present. & 5.14 \\
EE-PD & Asks a specific question about the person who is the subject of the trouble (e.g., 'What was he like?' or 'How old was he?') & 4.80 \\
EE-PD & Asks about the current condition or status of someone (e.g., 'How is she doing?', 'Is she okay', 'What is her current status?'). & 4.26 \\
EE-PSR & Encourages reminiscing about positive memories shared with someone. & 3.88 \\
EE-PEE & Asks the question 'How are you holding up?' & 3.85 \\
EE-PEE & Asks if the other person wants to talk about their feelings or situation. & 3.61 \\
DU-EC & Expresses inability to imagine or comprehend the situation using phrases like 'I can't imagine' or 'I can only imagine' & 3.59 \\
DU-AU & Mentions the unpredictability or uncertainty of life or death. & 3.45 \\
EE-PEE & Asks the question 'How are you feeling right now?' & 3.37 \\
EE-PSR & Asks what the person wishes they could have said to someone who is no longer present. & 3.35 \\
EE-PD & Mentions the speaker's son or asks a question about the speaker's son & 3.21 \\
DU-EC & Repeats the phrase 'I know.' & 3.14 \\
EE-PD & Asks questions about medical treatments or doctors' opinions. & 3.03 \\
EE-PSS & Asks about the presence of family or siblings for support & 3.00 \\
EE-PSR & Asks the recipient to share a favorite memory of the person who passed away. & 2.76 \\
DU-AP & Acknowledges explicitly that the current period of time is difficult or tough for the person. & 2.71 \\
EE-PD & Asks about the closeness or proximity of a relationship or distance. & 2.71 \\
EE-PD & Asks specifically about the type of cancer. & 2.60 \\
EE-PD & Asks 'What happened?' explicitly in the form of a question. & 2.14 \\
DU-AP & Uses the metaphor of carrying something heavy to describe the emotional burden. & 2.00 \\
EE-PSS & Asks if the other person has talked to someone about the situation & 1.97 \\
DU-AP & Uses metaphors or phrases to describe the situation as if the world or environment has been turned upside down. & 1.77 \\
DU-AP & Mentions the concept of a void or emptiness. & 1.48 \\
DU-AP & Repeats the phrase 'I hear you.' & 1.24 \\
EE-PD & Mentions the diagnosis or medical condition of the person's mother specifically & 1.24 \\
EE-PSS & Encourages asking the person directly what they need or want for support. & 0.87 \\
EE-PSR & Asks questions about the other person's experience or feelings, specifically focusing on difficulties or hardest parts. & 0.77 \\
\bottomrule
\end{tabularx}
\caption{\textbf{Cognitive empathic communication taxonomy for personal trouble scenarios.} Subcategory combines Level 2 and Level 3 categories (Level 2: DU = Demonstrating Understanding; EE = Encouraging Elaboration; Level 3: AP = Acknowledging Perspective; AU = Acknowledging Uncertainty; EC = Expressing Comprehension; PD = Promoting Dialogue; PEE = Promoting Emotional Expression; PSR = Promoting Self-Reflection; PSS = Promoting Support Seeking). Concept descriptions are LLM-generated summaries of k-sparse autoencoder features. Percent values indicate each concept’s share of all concept occurrences within the same domain and scenario type, computed from concept counts, and sum to 100\%.}
\end{table}
\begin{table}[htbp]
\centering
\label{tab:personal_misattuned}
\renewcommand{\arraystretch}{.6}
\footnotesize
\begin{tabularx}{\textwidth}{lXr}
\toprule
\multicolumn{3}{l}{\textbf{Personal Troubles - Misattuned}} \\
\midrule
\textbf{Subcategory} & \textbf{Concept Description} & \textbf{\%} \\
\midrule
AG-PR & Encourages the person to be strong or stay strong in the face of difficulty & 6.50 \\
AG-PEP & Encourages expressing love or appreciation directly to someone. & 5.56 \\
AG-PSG & Discusses providing support to someone else in a direct and actionable way & 5.13 \\
AG-PSG & Encourages the action of 'being there for her' explicitly using the phrase 'be there for her' & 5.08 \\
DE-PR & Mentions the concept of not feeling guilty or not blaming oneself. & 4.43 \\
AG-PSS & Mentions leaning on others for support (e.g., friends, family, support groups) & 4.19 \\
AG-PPC & Encourages maintaining a positive mindset or outlook. & 4.11 \\
AG-PSS & Mentions visiting or travel to see someone & 4.00 \\
AG-PSS & Mentions talking or having a conversation with someone. & 3.58 \\
AG-PR & Uses the phrase 'take it one day at a time' & 3.56 \\
AG-PSS & Encourages talking to someone as a way to cope or find support. & 3.37 \\
SO-SPE & Mentions having personally experienced a similar situation or event & 3.22 \\
AG-PPC & Mentions carrying forward positive traits, values, or lessons from the deceased to the next generation. & 3.19 \\
AG-PEP & Mentions prayer or praying explicitly & 3.09 \\
AG-PEP & Suggests taking a break or engaging in a calming activity to relax or distract oneself & 3.09 \\
AG-PEP & Mentions treasuring or cherishing simple, special, or quiet moments or memories. & 3.00 \\
AG-PSG & Suggestions phrased as 'maybe' or 'it might' followed by an action or solution. & 2.89 \\
AG-PSG & Encourages doing what is within one's ability or control & 2.82 \\
AG-PEP & Encourages the person to actively feel and acknowledge their emotions without judgment. & 2.65 \\
DE-PAP & Mentions that something will take time & 2.52 \\
AG-PR & Encourages moving forward or continuing with life despite the situation & 2.46 \\
AG-PSG & Mentions the concept of something being helpful or providing help. & 2.39 \\
AG-PPC & Encourages the recipient to take care of themselves. & 2.37 \\
AG-PSG & Mentions taking time off from work & 2.37 \\
DE-PAP & Expresses a personal belief or opinion using phrases like 'I think' or 'I believe' & 2.26 \\
AG-PEP & Encourages taking time to process emotions or situations explicitly & 2.11 \\
AG-PEP & Encourages expressing emotions or feelings openly, such as crying or showing vulnerability. & 1.95 \\
DE-PAP & Mentions the importance of family. & 1.78 \\
DE-PAP & References what the deceased person would want or feel about the situation & 1.72 \\
AG-PSS & Emphasizes spending time with a loved one. & 1.63 \\
SO-SPE & Mentions personal experience or connection with cancer or someone who has had cancer & 1.58 \\
AG-PEP & Suggests writing thoughts or feelings down, specifically in a letter or journal. & 1.37 \\
\bottomrule
\end{tabularx}
\caption{\textbf{Misattuned empathic communication taxonomy for personal trouble scenarios.} Subcategory combines Level 2 and Level 3 categories (Level 2: AG = Advice-Giving / Advice-giving; DE = Dismissing Emotions; SO = Self-oriented; Level 3: PAP = Providing Additional Perspective / Providing additional perspective; PEP = Promoting Emotional Processing; PPC = Promoting Positive Change; PR = Promoting Resilience / Providing Reassurance; PSG = Problem-Solving Guidance; PSS = Promoting Support Seeking; SPE = Sharing Personal Experience). Concept descriptions are LLM-generated summaries of k-sparse autoencoder features. Percent values indicate each concept’s share of all concept occurrences within the same domain and scenario type, computed from concept counts, and sum to 100\%.}
\end{table}
\begin{table}[htbp]
\centering
\label{tab:personal_motivational}
\renewcommand{\arraystretch}{.6}
\footnotesize
\begin{tabularx}{\textwidth}{lXr}
\toprule
\multicolumn{3}{l}{\textbf{Personal Troubles - Motivational}} \\
\midrule
\textbf{Subcategory} & \textbf{Concept Description} & \textbf{\%} \\
\midrule
A-SVAL & Contains short, direct responses or prompts without elaboration & 15.92 \\
A-SVAL & Single-word responses that convey acknowledgment or neutrality, ending with a period. & 13.39 \\
A-SVAL & Uses single words or very short phrases (1-2 words) that prompt further communication or action. & 7.51 \\
PR-MM & Mentions the continued presence, influence, or legacy of a person who has passed away, through memories, lessons, love, wisdom, or spirit. & 5.94 \\
PR-MM & Expresses certainty that the deceased person knew they were loved by the person being comforted. & 4.96 \\
PR-FO & Expresses reassurance that everything will be okay & 4.62 \\
PR-VR & Expresses certainty or confidence in a positive outcome using the phrase 'you will' or similar. & 3.97 \\
A-PR & Affirms that the person is doing their best and explicitly states that it is sufficient or enough. & 3.87 \\
A-PR & Praises an idea or strategy as being good, great, wonderful, or brilliant. & 3.79 \\
PR-FO & Mentions that things will improve with time & 3.66 \\
PR-VR & Expresses certainty using the phrase 'I'm sure' & 3.53 \\
A-SVAL & Single word 'Yes.' & 3.03 \\
A-PSW & Encourages the recipient to be kind or gentle to themselves. & 2.72 \\
A-PSW & Encourages the recipient by emphasizing their strength or capability, often using phrases like 'you can do this' or 'you are strong' & 2.70 \\
PR-N & Expresses reassurance that it is acceptable to not have all the answers or clarity immediately. & 2.47 \\
PR-FO & Mentions the phrase 'get through this' or variations of it & 2.20 \\
A-SVAL & Expresses gratitude or acknowledgment by saying 'You're welcome' or similar phrases & 2.09 \\
PR-MM & Uses the phrase 'take heart' & 1.95 \\
A-PR & Expresses gladness or happiness in response to the other person's feelings or situation. & 1.84 \\
PR-FO & Mentions the word 'comfort' or phrases related to providing comfort. & 1.82 \\
A-SVAL & Uses the word 'Absolutely' as a standalone affirmation. & 1.74 \\
PR-N & Explicitly states 'You're not alone' & 1.42 \\
PR-MM & Expresses that someone (often deceased) is or would be proud of the person being comforted & 1.34 \\
A-SVAL & Uses the word 'Exactly.' & 1.13 \\
PR-N & Expresses confidence that someone else will understand the situation. & 0.88 \\
A-SVAL & Expresses gratitude or gladness for having provided comfort or help. & 0.86 \\
A-SVAL & Uses the phrase 'Of course.' & 0.67 \\
\bottomrule
\end{tabularx}
\caption{\textbf{Motivational empathic communication taxonomy for personal trouble scenarios.} Subcategory combines Level 2 and Level 3 categories (Level 2: A = Affirming; PR = Providing Reassurance; Level 3: FO = Future-Oriented; MM = Meaning-Making; N = Normalization; PR = Positive Reinforcement; PSW = Promoting Self-Worth; SVAL = Short, Vague affirmative language; VR = Vague reassurance). Concept descriptions are LLM-generated summaries of k-sparse autoencoder features. Percent values indicate each concept’s share of all concept occurrences within the same domain and scenario type, computed from concept counts, and sum to 100\%.}
\end{table}
\begin{table}[htbp]
\centering
\label{tab:workplace_affective}
\renewcommand{\arraystretch}{.6}
\footnotesize
\begin{tabularx}{\textwidth}{lXr}
\toprule
\multicolumn{3}{l}{\textbf{Workplace Troubles - Affective}} \\
\midrule
\textbf{Subcategory} & \textbf{Concept Description} & \textbf{\%} \\
\midrule
VE-VEE & Acknowledges the frustration or emotional difficulty of feeling unrecognized or unnoticed for one's efforts. & 8.21 \\
ES-AL & Expresses sympathy using the exact phrase 'I'm sorry to hear that.' & 7.50 \\
VE-VEE & Uses the phrase 'That's tough' or variations like 'It's tough' or 'It is tough' & 6.51 \\
DA-PS & Expresses unconditional support by explicitly stating 'I am here for you.' & 6.06 \\
DA-PS & Offers direct assistance or help to the other person. & 5.66 \\
VE-VEE & Explicitly validates and normalizes the person's feelings as normal and acceptable. & 5.10 \\
VE-NE & Uses the words 'awful' or 'terrible' to describe the situation. & 4.86 \\
DA-PS & Expresses consistent availability to support or listen ('always here' or 'whenever you need') & 4.83 \\
ES-AL & Uses the phrase 'sorry about that' verbatim. & 4.64 \\
VE-VEE & Mentions losing a job or the emotional impact of job loss & 4.41 \\
DA-OH & Asks if there is anything they can do to help & 4.17 \\
VE-NE & Uses the word 'overwhelming' to describe the emotional state or situation. & 3.97 \\
ES-AL & Repeats the phrase 'I'm so sorry.' & 3.76 \\
ES-AL & Expresses sympathy specifically by saying 'I'm sorry you feel that way' or a variation of it & 3.44 \\
ES-AL & Contains the exact phrase 'Sorry to hear that.' & 3.11 \\
DA-OH & Proposes meeting up or doing an activity together to address the issue or relax. & 3.06 \\
VE-NE & Uses the word 'frustrating' to describe the situation. & 2.77 \\
ES-EE & Uses the word 'really' to emphasize the expression of sympathy or support. & 2.75 \\
VE-VEE & Expresses that the person is not alone in their experience or situation. & 2.66 \\
ES-EE & Begins with 'Wow' or 'Oh wow' & 2.34 \\
DA-PS & Uses the word 'anytime' to express availability or support & 2.29 \\
VE-VEE & Acknowledges that even when something is anticipated, it can still be emotionally impactful or shocking. & 2.24 \\
VE-NE & Mentions feeling invisible or unseen. & 2.22 \\
ES-EE & Exclaiming 'Oh no' to express shock or sympathy. & 1.74 \\
VE-VEE & Mentions that the situation is tough, difficult, tricky, or rough & 1.71 \\
\bottomrule
\end{tabularx}
\caption{\textbf{Affective empathic communication taxonomy for workplace trouble scenarios.} Subcategory combines Level 2 and Level 3 categories (Level 2: DA = Demonstrating Availability; ES = Expressing Sympathy; VE = Validating Emotions; Level 3: AL = Apologetic Language; EE = Emotional Exclamation; NE = Naming Emotions; OH = Offer help; PS = Providing Support; VEE = Validating Emotional Experience). Concept descriptions are LLM-generated summaries of k-sparse autoencoder features. Percent values indicate each concept’s share of all concept occurrences within the same domain and scenario type, computed from concept counts, and sum to 100\%.}
\end{table}
\begin{table}[htbp]
\centering
\label{tab:workplace_cognitive}
\renewcommand{\arraystretch}{.6}
\footnotesize
\begin{tabularx}{\textwidth}{lXr}
\toprule
\multicolumn{3}{l}{\textbf{Workplace Troubles - Cognitive}} \\
\midrule
\textbf{Subcategory} & \textbf{Concept Description} & \textbf{\%} \\
\midrule
DU-EC & Expresses understanding or relatability to the other person's feelings using phrases like 'I understand how you feel' or 'I can relate to how you are feeling' & 6.21 \\
DU-EC & Uses the exact phrase 'I understand.' & 5.58 \\
DU-EC & Explicitly expresses understanding of the other person's emotions or feelings using first-person perspective (e.g., 'I understand your feeling', 'I can feel your pain'). & 5.50 \\
EE-PD & Asks about the recipient's job or workplace culture. & 4.90 \\
EE-PSR & Asks a direct question about what is causing the other person's feelings. & 4.84 \\
EE-PSS & Asks if the person has talked to someone or sought help about the situation & 4.18 \\
DU-AU & Mentions the unpredictability or uncontrollability of life events. & 3.67 \\
DU-AP & Acknowledges the person's effort and explicitly connects their feelings of discouragement or disappointment to the significant effort they have invested. & 3.62 \\
EE-PSS & Asks if there is someone the person can talk to for support, specifically mentioning work or close relationships. & 3.58 \\
EE-PD & Asks about the other person's actions or ongoing tasks & 3.53 \\
EE-PD & Asks 'What happened?' as a direct question & 3.50 \\
DU-EC & Mentions self-doubt explicitly & 3.45 \\
DU-EC & Uses the phrase 'I get that.' & 3.45 \\
EE-PD & Asks the other person to share more details by explicitly requesting them to 'tell me more'. & 3.41 \\
EE-PEE & Asks explicitly if the other person wants to talk about their feelings or situation & 3.30 \\
EE-PD & Asks for clarification or reasons behind the situation. & 3.04 \\
DU-AU & Mentions uncertainty or fear of the unknown. & 2.96 \\
EE-PEE & Encourages talking about feelings or expressing emotions as a way to process or address the situation. & 2.94 \\
EE-PD & Asks a question about the other person's thought process ('Why do you think that?') & 2.80 \\
EE-PEE & Asks about the other person's current emotional or physical state using a question. & 2.76 \\
EE-PD & Asks about the other person's next steps or plans for the future & 2.54 \\
DU-EC & Uses variations of the phrase 'I can/can’t only imagine' to acknowledge the difficulty of understanding the other person's experience. & 2.47 \\
DU-EC & Begins with 'I know how much' or 'I know how' & 2.44 \\
DU-EC & Explicitly states 'I know' or 'I do know' & 2.29 \\
DU-AP & Repeats the phrase 'I hear you.' & 2.26 \\
EE-PEE & Asks how the person is feeling right now. & 2.08 \\
EE-PSR & Asks what the recipient feels they could have done differently. & 1.79 \\
EE-PD & Asks the question 'Why do you think no one cares?' & 1.59 \\
EE-PD & Mentions age or asks about age-related information. & 1.50 \\
EE-PD & Asks about the reason or cause behind sensing something was coming. & 1.36 \\
DU-AP & Mentions feeling stuck in a cycle or loop. & 1.23 \\
EE-PD & Asks about the duration of time spent working at a specific place or company. & 1.21 \\
\bottomrule
\end{tabularx}
\caption{\textbf{Cognitive empathic communication taxonomy for workplace trouble scenarios.} Subcategory combines Level 2 and Level 3 categories (Level 2: DU = Demonstrating Understanding; EE = Encouraging Elaboration; Level 3: AP = Acknowledging Perspective; AU = Acknowledging Uncertainty; EC = Expressing Comprehension; PD = Promoting Dialogue; PEE = Promoting Emotional Expression; PSR = Promoting Self-Reflection; PSS = Promoting Support Seeking). Concept descriptions are LLM-generated summaries of k-sparse autoencoder features. Percent values indicate each concept’s share of all concept occurrences within the same domain and scenario type, computed from concept counts, and sum to 100\%.}
\end{table}
\begin{table}[htbp]
\centering
\label{tab:workplace_misattuned}
\renewcommand{\arraystretch}{.6}
\footnotesize
\begin{tabularx}{\textwidth}{lXr}
\toprule
\multicolumn{3}{l}{\textbf{Workplace Troubles - Misattuned}} \\
\midrule
\textbf{Subcategory} & \textbf{Concept Description} & \textbf{\%} \\
\midrule
AG-PSG & Mentions finding or looking for a new or better job & 6.33 \\
AG-PSG & Mentions taking small steps or baby steps as a way to make progress & 5.44 \\
AG-PR & Encourages persistence and not giving up despite challenges & 5.08 \\
AG-PEP & Encourages taking time to process emotions or situations at one's own pace. & 4.82 \\
DE-PR & Uses the word 'maybe' to suggest uncertainty or a tentative explanation. & 4.67 \\
AG-PSG & Encourages taking a break or relaxing & 4.22 \\
AG-PSG & Mentions talking to a manager or suggesting speaking with a manager & 4.19 \\
AG-PSG & Suggests specific actions or activities to help cope or improve the situation & 4.10 \\
DE-PR & Uses the phrase 'don't worry' or a variation of it to reassure the other person. & 3.83 \\
DE-PAP & Speculates that the other party may be unaware of their actions or feelings of the person. & 3.62 \\
AG-PSG & Mentions updating a resume as a specific action or suggestion & 3.38 \\
AG-PR & Mentions getting through or overcoming a situation, often using the phrase 'get through this'. & 3.32 \\
DE-PR & Mentions that setbacks do not define a person & 3.30 \\
AG-PSW & Mentions building or improving confidence as a skill or process & 3.22 \\
AG-PPC & Encourages self-improvement or learning from mistakes for future efforts & 3.10 \\
AG-PSG & Suggests or encourages having a conversation or bringing up the topic with someone else & 3.08 \\
SO-SPE & Describes personal actions or strategies taken to overcome a challenge or improve the situation. & 3.08 \\
AG-PPC & Encourages maintaining a positive outlook or mindset. & 2.98 \\
DE-NE & Mentions shared experiences or feelings using inclusive language such as 'we all' or 'I think we all' & 2.84 \\
AG-PSG & Uses directive language with phrases like 'you need to', 'you have to', or 'you must' to encourage action or change. & 2.74 \\
SO-SPE & Mentions having personally experienced the same situation as the other person & 2.70 \\
AG-PSG & Encourages creating a plan or thinking through options to address the situation. & 2.65 \\
AG-PR & Encourages someone to stay positive or resilient by using phrases like 'keep your chin up' or 'take heart' & 2.31 \\
AG-PEP & Encourages acknowledging and processing emotions & 2.19 \\
AG-PPC & Mentions the concept of starting over or restarting & 2.17 \\
AG-PSG & Encourages the recipient to try or make an attempt at something & 2.13 \\
AG-PR & Mentions taking things 'one day at a time' & 1.89 \\
DE-NE & Mentions the concept of change or transformation explicitly & 1.83 \\
AG-PSG & Mentions asking for or seeking feedback & 1.68 \\
AG-PSG & Mentions financial support or unemployment benefits as a concrete next step & 1.66 \\
AG-PSG & Mentions talking to or communicating with a boss & 1.48 \\
\bottomrule
\end{tabularx}
\caption{\textbf{Misattuned empathic communication taxonomy for workplace trouble scenarios.} Subcategory combines Level 2 and Level 3 categories (Level 2: AG = Advice-Giving / Advice-giving; DE = Dismissing Emotions; SO = Self-oriented; Level 3: NE = Normalizing Experience; PAP = Providing Additional Perspective; PEP = Promoting Emotional Processing; PPC = Promoting Positive Change; PR = Promoting Resilience / Providing Reassurance; PSG = Problem-Solving Guidance; PSW = Promoting Self-Worth; SPE = Sharing Personal Experience). Concept descriptions are LLM-generated summaries of k-sparse autoencoder features. Percent values indicate each concept’s share of all concept occurrences within the same domain and scenario type, computed from concept counts, and sum to 100\%.}
\end{table}
\begin{table}[htbp]
\centering
\label{tab:workplace_motivational}
\renewcommand{\arraystretch}{.2}
\footnotesize
\begin{tabularx}{\textwidth}{lXr}
\toprule
\multicolumn{3}{l}{\textbf{Workplace Troubles - Motivational}} \\
\midrule
\textbf{Subcategory} & \textbf{Concept Description} & \textbf{\%} \\
\midrule
A-SVAL & Contains short, one or two-word phrases or responses. & 8.30 \\
A-SVAL & Uses short, non-lexical expressions of acknowledgment or contemplation, typically one or two syllables (e.g., 'Hm.', 'Oh.', 'Mm.'). & 6.70 \\
A-SVAL & Uses enthusiastic or affirmative exclamations (e.g., 'Yes!', 'Wonderful!', 'Do it!', 'Ridiculous!') & 5.59 \\
A-PSW & Reassures the person that the situation does not define their worth or abilities. & 4.47 \\
A-PSW & Encourages self-kindness or self-focus explicitly & 4.06 \\
A-PR & Mentions hard work in a positive and appreciative manner & 3.85 \\
A-SVAL & Affirmative responses using 'Yes.' & 3.65 \\
A-PSW & Explicitly compliments the person's qualities or abilities, often using adjectives like 'amazing', 'great', or 'wonderful'. & 3.39 \\
A-PSW & Encourages reflection on personal accomplishments & 3.38 \\
A-PSW & Expresses that the individual and their contributions have inherent value or significance (e.g., 'You matter', 'Your efforts matter'). & 3.30 \\
A-SVAL & Uses short, affirming phrases such as 'Okay', 'Good', or 'Great'. & 3.17 \\
A-PR & Praises the person's efforts by explicitly stating that they are doing their best or the best they can. & 2.86 \\
A-PSW & Mentions the value of the recipient's skills and experience in a positive and affirming way. & 2.78 \\
A-SVAL & Affirms the truth or validity of a statement using the word 'true' & 2.67 \\
A-PSW & Expresses belief in the other person's abilities or potential explicitly with the phrase 'I believe in you.' & 2.64 \\
PR-FO & Predicts a positive future event or outcome specifically for the recipient. & 2.49 \\
PR-FO & Mentions the existence of future opportunities or chances for success. & 2.44 \\
A-PSW & Mentions the word 'notice' or variations of it to acknowledge recognition. & 2.43 \\
A-PSW & Highlights the individual's inner strength and resilience explicitly & 2.42 \\
A-SVAL & Expresses agreement or approval using the phrase 'That sounds great' or 'That sounds good' & 2.29 \\
PR-FO & Expresses certainty that the person will succeed or achieve something, using the phrase 'You will'. & 2.28 \\
PR-VR & Expresses confidence or reassurance using the phrase 'I am sure' or 'I'm sure' & 2.28 \\
PR-FO & Expresses optimism that the situation will improve in the future & 2.11 \\
A-SVAL & Expresses gladness or relief in response to a positive outcome or assistance provided. & 1.92 \\
A-PR & Praises or affirms the mindset or approach of the other person as being positive or good & 1.86 \\
A-SVAL & Expresses agreement or approval of an idea by explicitly calling it 'great' or 'good' & 1.80 \\
A-SVAL & Uses the phrase 'Exactly.' & 1.70 \\
A-PSW & Uses the phrase 'You’ve got this.' & 1.63 \\
A-PSW & Expresses that the recipient deserves appreciation, recognition, or reward. & 1.55 \\
PR-FO & Mentions the metaphor of a door closing and another door opening to signify new opportunities. & 1.49 \\
A-PR & Wishes the recipient the best or expresses hope for their future. & 1.44 \\
A-SVAL & Expresses gratitude or acknowledgment with the phrase 'You are welcome' or variations such as 'You're very welcome.' & 1.33 \\
A-SVAL & Expresses explicit caring using the phrase 'I care.' & 1.16 \\
A-PSW & Mentions leadership qualities or roles explicitly & 1.14 \\
PR-N & Reassures the person that it is okay to not have everything figured out immediately & 1.06 \\
A-SVAL & Uses the phrase 'That makes a lot of sense.' & 0.82 \\
A-SVAL & Contains the word 'hello' exactly as written (case-insensitive) & 0.79 \\
A-PR & Mentions fixing a bug or related accomplishment in a positive or appreciative context & 0.74 \\
\bottomrule
\end{tabularx}
\caption{\textbf{Motivational empathic communication taxonomy for workplace trouble scenarios.} Subcategory combines Level 2 and Level 3 categories (Level 2: A = Affirming; PR = Providing Reassurance; Level 3: FO = Future-Oriented; N = Normalization; PR = Positive Reinforcement; PSW = Promoting Self-Worth; SVAL = Short, Vague affirmative language; VR = Vague reassurance). Concept descriptions are LLM-generated summaries of k-sparse autoencoder features. Percent values indicate each concept’s share of all concept occurrences within the same domain and scenario type, computed from concept counts, and sum to 100\%.}
\end{table}

\clearpage
\newpage

\section{GUIDE-LLM Reporting Checklist \cite{feuerriegelreporting}}

\begin{longtable}{|p{0.7\textwidth}|p{0.3\textwidth}|}
\hline
\rowcolor{green!20}
\textbf{Scope of LLM use} & \textbf{Answer} \\*
\hline

\textbf{Item A.1:} LLMs were used in this project for:

\textcolor{gray}{{\footnotesize
\textbf{Explanation:} Briefly describe how and for what purposes LLMs were used in the study. This may include one or multiple stages of the research workflow, depending on the project’s design and aims. The following examples illustrate common use cases:
\begin{itemize}[leftmargin=*]\setlength{\itemsep}{0pt}\setlength{\parskip}{0pt}
    \item \textbf{Research design} (e.g., hypothesis generation, literature search, or creating surveys/stimuli).
    \item \textbf{Data processing} (e.g., transcription, translation, data extraction, or data cleaning).
    \item \textbf{Analysis} (e.g., data labeling, summarization, pattern detection, statistical analysis, or code generation)..
    \item \textbf{LLM as research object} (e.g., studying LLM behavior, benchmarking LLMs, or bias assessment of LLMs).
    \item \textbf{Participant-facing settings} (e.g., LLM used as an intervention, studying human interactions with LLM chatbots).
    \item \textbf{Communication} (e.g., paper writing, editing, or reviewing).
\end{itemize}
Depending on the specific use case described here, different checklist items may later be relevant, and, in many cases, it may be necessary that later items in the checklist are reported separately for each use case.}}
& {LLMs were used in three ways. First, participants across all conditions interacted with LLM role-playing agents to practice communicating empathically. Second, in the AI Coach and Combined Training conditions participants also interacted with an AI coach that provided personalized feedback after conversations. Third, LLM/API-based tools were used in analyses including scoring conversations on the six preregistered empathic communication dimensions, OpenAI embeddings for novelty and kSAE analyses, GPT-4o sentence-level coding of coach-feedback sentences, and Pangram V3 AI-text scoring for a 200-conversation subsample.} \\
\hline
\textbf{Item A.2:} Degree of automation (human-in-the-loop vs. fully automated):

\textcolor{gray}{{\footnotesize
\textbf{Explanation:} Indicate how much human oversight was involved. Specify whether each output was reviewed, edited, or approved by a person, or whether outputs were used automatically without supervision. For participant-facing tasks, state whether humans checked outputs before showing them to participants or whether participants interacted with the LLM directly. Specify who provided oversight (e.g., student assistant, expert, PI).}}  & {Participants interacted with the LLM conversational partner and the AI coach directly.} \\
\hline
\end{longtable}

\begin{longtable}{|p{0.7\textwidth}|p{0.3\textwidth}|}
\hline
\rowcolor{green!20}
\textbf{Model/system details} & \textbf{Answer} \\*
\hline

\textbf{Item B.1:} Model name, including provider, model size, exact version/ID, date of access, and source link (if possible):

\textcolor{gray}{{\footnotesize
\textbf{Explanation:} Report the exact model names (including provider, version, and date accessed). Avoid generic labels like “ChatGPT” or “GPT-4”; instead, use detailed model names such as “GPT-4o-mini-2024-12-17 (OpenAI)” or “Llama-3.1-8B (Meta; accessed via HuggingFace in May 2025)”. For locally deployable models, please also enter a source link (e.g., the URL to the HuggingFace page). If multiple models were tested, it is encouraged to name them and briefly explain which one was used in the final study and why. When multiple models served different purposes, specify their respective roles, consistent with your response to Item A.1.}}
& {The LLM conversational partner and the AI coach were based on API calls to OpenAI's \texttt{gpt-4o} model. Analysis-only API calls include OpenAI \texttt{text-embedding-3-large} for kSAE workflows, OpenAI \texttt{gpt-4o} for coach-feedback sentence coding, OpenAI \texttt{text-embedding-3-small} for response-novelty embeddings, and Pangram Labs V3 API for AI-text detection scores.} \\
\hline
\textbf{Item B.2:} Model access (e.g., API, web interface, local) and context mode (e.g., chat mode or separate calls):

\textcolor{gray}{{\footnotesize
\textbf{Explanation:} Note how you accessed the models (e.g., API, web interface, local installation) and whether you used LLMs in chat mode (ongoing conversation) or stateless mode (separate prompts). Mention the exact API name and version, since different access modes may influence responses (e.g., due to differences in model routing). }}  & {OpenAI API for all OpenAI models; Pangram Labs V3 API for AI-text detection. Participant-facing systems (role-playing partner and AI coach) used chat mode with within-session conversation history. Analysis calls used separate stateless API requests.} \\
\hline
\textbf{Item B.3:} Relevant LLM configurations reported (as applicable), such as temperature, max tokens, seed, and number of runs:

\textcolor{gray}{{\footnotesize
\textbf{Explanation:} List any configuration settings that may affect outputs, such as:
\begin{itemize}[leftmargin=*]\setlength{\itemsep}{0pt}\setlength{\parskip}{0pt}
    \item \textbf{temperature} which controls randomness of the model’s output)
    \item \textbf{Sampling parameters} such as \emph{top\_k}, \emph{top\_p}, \emph{max tokens} (which limit the candidate token set or enforce length constraints)are considered, or to enforce a length limit)
    \item \textbf{Penalties} that discourage repetition (e.g., a frequency penalty to reduce the likelihood of tokens proportional to how often they have already appeared; a presence penalty reduces the likelihood of any token that has appeared at least once)
    \item \textbf{Stop sequences} (which halt generation when such a top sequence is produced, e.g., [``$\backslash n \backslash n$'', ``END'']).
    \item \textbf{Number of completions or runs} (which is often used to capture variability in outputs across repeated generations)
    \item \textbf{Quantization level} (e.g., FP16, INT8, INT4) to change numerical precision beyond the default
    \item \textbf{Reasoning-related settings}, such as whether a specific structured reasoning was enabled, the specified reasoning effort level (e.g., low/medium/high or numerical settings that influence the depth of the reasoning), and any compute or inference budget constraints tied to the chosen reasoning mode
\end{itemize}
} } & {Temperature = 0 for participant-facing \texttt{gpt-4o} calls and for analysis \texttt{gpt-4o} calls (scoring and coach-feedback sentence coding). Role-playing partner responses were capped at 3 sentences via prompt instructions.} \\
\hline
\textbf{Item B.4:} Customization:

\textcolor{gray}{{\footnotesize
\textbf{Explanation:} Check and describe any modifications or extended capabilities incorporated into your LLM setup beyond standard inference. This includes, but is not limited to:
\begin{itemize}[leftmargin=*]\setlength{\itemsep}{0pt}\setlength{\parskip}{0pt}
\item \textbf{Fine-tuning} (e.g., via LoRA; Low-Rank Adaptation) used to adapt a pretrained model to domain-specific data.
\item \textbf{Retrieval-augmentation generation (RAG)}, where the model retrieves relevant information from external sources (e.g., databases or document collections) during inference.
\item \textbf{Automated prompt optimization} (e.g., DSPy) that treat prompts as trainable parameters.
\item \textbf{Web search integration}, indicating whether the LLM was able to access and retrieve information from the Internet.
\item \textbf{Agentic workflows}, including multi-step reasoning processes or delegated actions such as tool/function calling (e.g., via LangChain, AutoGPT, CrewAI).
\item \textbf{Post-training refinements}, including alignment or optimization techniques used to adjust model behavior after pretraining (e.g., reinforcement learning from human feedback (RLHF), direct preference optimization (DPO)).
\end{itemize}
The goal is to specify any added customizations or provider-specific features that meaningfully shape system behavior in order to enable others to understand and accurately reproduce your setup.
}}  &   \begin{itemize}[leftmargin=*]\setlength{\itemsep}{0pt}\setlength{\parskip}{0pt}
   \item[\checked] Base model
   \item[\unchecked] Fine-tuning
   \item[\unchecked] RAG (retrieval-augmented)
   \item[\unchecked] Automated prompt optimization
   \item[\unchecked] Tool/function calling
   \item[\unchecked] Web search
   \item[\unchecked] Agentic workflows
   \item[\unchecked] Other adaptations (e.g., safety mechanisms)
\end{itemize}

Description: {Prompt-engineered system instructions only; no fine-tuning or other model adaptations.} \\
\hline
\textbf{Item B.5:} Did the LLM session(s) include persistent memory across interactions?

\textcolor{gray}{{\footnotesize
\textbf{Explanation:} Indicate whether the LLM could “remember” previous conversations (i.e., had persistent memory). Unless such memory is disabled, there may also be spillover effects from other chat windows or prior conversations, which can influence outputs even when not intended. }}  &   \begin{itemize}[leftmargin=*]\setlength{\itemsep}{0pt}\setlength{\parskip}{0pt}
    \item[\checked] Yes
    \item[\unchecked] No
    \item[\unchecked] N/A
\end{itemize} \\
\hline
\end{longtable}

\newpage

\begin{longtable}{|p{0.7\textwidth}|p{0.3\textwidth}|}
\hline
\rowcolor{green!20}
\textbf{{Prompts}} & \textbf{Answer} \\*
\hline

\textbf{Item C.1:} Exact prompt(s) reported:

\textcolor{gray}{{\footnotesize
\textbf{Explanation:} Whenever possible, include the exact text of prompts you used, including in-context examples or demonstrations provided to the LLM. Even small wording changes, formatting, or ordering of examples can substantially affect outputs. If full prompts cannot be shared (e.g., due to privacy or length), include a redacted or representative example or link to the full prompt in a repository (e.g., OSF, GitHub). }}
& {The prompts used for the LLM conversational partner and AI coach are available in the Supplementary Information Sections 3 and 5.} \\
\hline
\textbf{Item C.2:} System-wide instructions (if any):

\textcolor{gray}{{\footnotesize
\textbf{Explanation:} Note any system-level instructions that guide the model’s general behavior (e.g., “You are a helpful assistant.”). These are commonly not directly visible but can be accessed through the API.
}}  & {System-wide instructions for the role-playing partner and AI coach are embedded in the prompts reported in the Supplementary Information.} \\
\hline
\end{longtable}

\begin{longtable}{|p{0.7\textwidth}|p{0.3\textwidth}|}
\hline
\rowcolor{green!20}
\textbf{{Data inputs \& privacy}} & \textbf{Answer} \\*
\hline

\textbf{Item D.1:} Handling of personal or sensitive data (if any) (e.g., consent for data processing):

\textcolor{gray}{{\footnotesize
\textbf{Explanation:} If any personal, sensitive, or identifiable data were processed, describe how they were handled in compliance with ethical standards and data protection laws. Researchers should indicate whether participants explicitly consented to their data being analyzed with an LLM, particularly when proprietary, cloud-based models are used. Such processing typically involves transferring data to a private company that may retain them indefinitely, which raises additional ethical and legal considerations. Beyond consent, describe how sensitive or identifiable data were handled (e.g., de-identification, anonymization, masking) and whether the LLM provider offers safeguards such as excluding inputs from training or storage. Clarify where data were stored or processed and how applicable legal/ethical requirements were met. If relevant, address cross-border transfers, as data may be stored in jurisdictions with different privacy laws (e.g., EU vs. US), with implications for compliance with GDPR, HIPAA, or other frameworks.
For context, some providers (e.g., OpenAI) may log or inspect prompts even when the data are not used for model training. For sensitive datasets, zero-retention configurations may be required (e.g., the MIMIC datasets can only be used with OpenAI models if a zero-retention checkpoint is enabled).
}}
& {The raw data and analysis materials shared use de-identified data without any direct participant identifiers.} \\
\hline
\end{longtable}

\newpage
\begin{longtable}{|p{0.7\textwidth}|p{0.3\textwidth}|}
\hline
\rowcolor{green!20}
\textbf{{Validation \& interpretation}} & \textbf{Answer} \\*
\hline

\textbf{Item E.1:} Human validation of LLM outputs:

\textcolor{gray}{{\footnotesize
\textbf{Explanation:} If relevant, describe whether and how human reviewers examined the model’s outputs, and the degree of independence they had in doing so. Specify the reviewers’ roles (e.g., domain experts, research assistants, subject-matter specialists) and relevant expertise, as well as how many reviewers participated and how their work was organized. Indicate whether outputs were independently annotated, double-checked by multiple reviewers, or merely approved or edited post-hoc by a lead author or investigator.
Clarify what dimensions of performance were examined. These may include known performance metrics from ML/AI such as accuracy or other metrics like citation correctness, hallucination detection, agreement or inter-rater reliability. State whether qualitative judgments, quantitative metrics, or both were used. If outcome assessment required subjective interpretation, describe assessor qualifications, instructions provided, and relevant demographics.
Describe the selection procedure for the reviewed outputs—whether all outputs were examined, a random sample was drawn, or specific cases (e.g., rare events or high-stakes responses) were oversampled to capture potential rare or critical errors. Further report how reviewers were trained or instructed, what criteria or rating scales they used, and how disagreements were resolved. For multi-reviewer settings, provide any inter-rater or inter-assessor reliability statistics (e.g., Cohen’s $\kappa$ or Krippendorff's $\alpha$). Finally, note whether reviewer feedback was used purely for validation or also to refine prompts, retrain models, or adjust study procedures.
}}
&
\begin{itemize}[leftmargin=*]\setlength{\itemsep}{0pt}\setlength{\parskip}{0pt}
    \item[\checked] Yes
    \item[\unchecked] No
    \item[\unchecked] N/A
\end{itemize}

Description: {The six preregistered empathic communication dimensions can be reliably annotated by LLMs~\cite{kumar2026large} and serve as the primary dependent variables for our analysis.}\\
\hline
\textbf{Item E.2:} Describe any relevant post-processing (e.g., filtering in case of format mismatches, unit conversions, etc.):

\textcolor{gray}{{\footnotesize
\textbf{Explanation:} Describe any steps you took to clean or reformat LLM outputs (e.g., converting “positive/neutral/negative” to numeric codes, handling missing values, removing malformed entries). State how you handled inconsistent or unusable outputs and whether corrections were made with an automated script or manually. For example, when generating quantitative estimates (e.g., word counts, probabilities, or durations), the model may return values embedded in free text (e.g., “3.5 seconds”) that require parsing and conversion into standardized numerical units. Post-processing steps should be described clearly, including how formatting errors, null responses, or inconsistent output structures were handled, whether automated scripts or manual corrections were used, and whether any data were excluded or reinterpreted as a result.
}}
& \\
\hline
\end{longtable}

\begin{longtable}{|p{0.7\textwidth}|p{0.3\textwidth}|}
\hline
\rowcolor{green!20}
\textbf{{Reproducibility}} & \textbf{Answer} \\*
\hline

\textbf{Item F.1:} Code/notebooks/scripts for LLM calls shared:

\textcolor{gray}{{\footnotesize
\textbf{Explanation:} Indicate whether you have shared materials such as code, prompts, logs, or transcripts. Make sure sensitive information (e.g., API keys, private data) is removed. For code, make sure to add a README file.
}}
&
\begin{itemize}[leftmargin=*]\setlength{\itemsep}{0pt}\setlength{\parskip}{0pt}
    \item[\checked] Yes
    \item[\unchecked] No
    \item[\unchecked] N/A
\end{itemize}

Link/DOI: {\footnotesize \url{https://doi.org/10.5281/zenodo.20703371}} (raw deidentified data, analysis code, and LLM prompts) \\
\hline
\end{longtable}

\begin{longtable}{|p{0.7\textwidth}|p{0.3\textwidth}|}
\hline
\rowcolor{green!20}
\textbf{{Competing interests}} & \textbf{Answer} \\*
\hline

\textbf{Item G.1:} Funding, support, or other relevant relationships (including in-kind access to compute or models, or professional affiliations):

\textcolor{gray}{{\footnotesize
\textbf{Explanation:} Disclose any current or past funding, support, or other relevant relationships with entities that have a financial interest in LLMs (this includes not just AI companies like OpenAI, Anthropic, but also tech companies developing or investing in AI, e.g., Google, Meta, Microsoft). This could include (but is not limited to): research funding from or collaborative research with a company with an interest in LLMs for this project or any other project within the past years; in-kind access to compute or models; current or former professional affiliations with a company with an interest in LLMs; personal investments (e.g., stocks) in companies with an interest in LLMs; familial relationship with an employee of a company with an interest in LLMs; etc. Disclose these relationships regardless of whether or not you believe they impacted the research.
}}
&
\begin{itemize}[leftmargin=*]\setlength{\itemsep}{0pt}\setlength{\parskip}{0pt}
    \item[\unchecked] Yes. Description:
    \item[\checked] No
\end{itemize}

Link/DOI: {The authors declare no competing interests.} \\
\hline
\end{longtable}

\begin{longtable}{|p{0.7\textwidth}|p{0.3\textwidth}|}
\hline
\rowcolor{yellow!30}
\textbf{Optional items} & \textbf{Answer} \\*
\hline



\textbf{Discussion of the rationale for the prompt design:}

\textcolor{gray}{\footnotesize
\textbf{Explanation:} Explain how you designed your prompts. For example, indicate whether you used a structured format (e.g., explicit task description, definitions, step-by-step instructions, and output constraints), followed established prompt engineering guidelines or prior literature, adapted prompts from earlier studies, or relied on automated prompt optimization tools. Clarify whether the design was iterative (e.g., refined through pilot testing or error analysis), whether few-shot examples were included and how they were selected, and whether prompts were standardized across models to ensure comparability.
}
&
{The LLM Coach prompt included a detailed framework for empathic communication used in \cite{kumar2026large} and few-shot examples from a human coach.}
\\
\hline






\hline

\textbf{Conversation transcripts:}

\textcolor{gray}{\footnotesize
\textbf{Explanation:} For studies involving direct researcher/participant interaction with an LLM, provide anonymized transcripts or representative examples.
}
&
\begin{itemize}[leftmargin=*]\setlength{\itemsep}{0pt}
    \item[\checked] Yes, shared without sensitive information. Location: {\footnotesize \url{https://doi.org/10.5281/zenodo.20703371}}
    \item[\unchecked] No. Reason for not sharing:
    \item[\unchecked] N/A
\end{itemize}

\\
\hline




\end{longtable}

\end{document}